\documentclass[10pt,journal,compsoc]{IEEEtran}

\ifCLASSOPTIONcompsoc
  \usepackage[nocompress]{cite}
\else
  \usepackage{cite}
\fi

\ifCLASSINFOpdf
  \usepackage[pdftex]{graphicx}
  \graphicspath{{../pdf/}{../jpeg/}}
  \DeclareGraphicsExtensions{.pdf,.jpeg,.png}
\else
  \usepackage[dvips]{graphicx}
  \graphicspath{{../eps/}}
  \DeclareGraphicsExtensions{.eps}
\fi

\usepackage{amsmath}
\usepackage{algorithmic}
\usepackage{array}

\ifCLASSOPTIONcompsoc
  \usepackage[caption=false,font=footnotesize,labelfont=sf,textfont=sf]{subfig}
\else
  \usepackage[caption=false,font=footnotesize]{subfig}
\fi

\usepackage{stfloats}
\ifCLASSOPTIONcaptionsoff
  \usepackage[nomarkers]{endfloat}
 \let\MYoriglatexcaption\caption
 \renewcommand{\caption}[2][\relax]{\MYoriglatexcaption[#2]{#2}}
\fi

\let\MYorigsubfloat\subfloat
\renewcommand{\subfloat}[2][\relax]{\MYorigsubfloat[]{#2}}

\usepackage{url}
\usepackage{color}

\usepackage{amsmath,amsfonts,bm}

\def\eqref#1{equation~\ref{#1}}

\def\1{\bm{1}}

\def\rvx{{\mathbf{x}}}

\def\rvz{{\mathbf{z}}}

\def\vf{{\bm{f}}}

\def\vx{{\bm{x}}}

\def\mF{{\bm{F}}}

\def\mX{{\bm{X}}}

\DeclareMathAlphabet{\mathsfit}{\encodingdefault}{\sfdefault}{m}{sl}
\SetMathAlphabet{\mathsfit}{bold}{\encodingdefault}{\sfdefault}{bx}{n}

\usepackage{lettrine}

\usepackage{booktabs}
\usepackage{multirow}
\usepackage{diagbox}

\usepackage[dvipsnames]{xcolor}

\usepackage{pifont}
\newcommand*\colourcheck[1]{%
  \expandafter\newcommand\csname #1check\endcsname{\textcolor{#1}{\ding{52}}}%
}
\colourcheck{blue}
\colourcheck{green}
\colourcheck{red}

\usepackage{tikz}
\newcommand*\circled[1]{\tikz[baseline=(char.base)]{
            \node[shape=circle,draw,inner sep=0.5pt] (char) {#1};}}

\usepackage{color}

\usepackage{colortbl}
\usepackage{comment}

\usepackage{enumitem}

\usepackage{hyperref}

\usepackage{siunitx}
\newcommand{\etal}{\textit{et al.~}}

\newcommand{\brown}[1]{\textcolor{black}{#1}}

\newcommand{\MYhref}[3][blue]{\href{#2}{\color{#1}{#3}}}

\newcommand{\newtext}[1]{{\textcolor{black}{#1}}}
\newcommand{\emptext}[1]{{\textcolor{black}{#1}}}

\newcommand{\topone}[1]{{\textcolor{WildStrawberry}{#1}}}
\newcommand{\toptwo}[1]{{\textcolor{RoyalBlue}{#1}}}
\newcommand{\topthree}[1]{{\textcolor{black}{#1}}}
\newcommand{\placeholder}[1]{{\textcolor{black}{#1}}}

\begin{document}
\title{StudioGAN: A Taxonomy and Benchmark \\ of GANs for Image Synthesis}
\author{Minguk Kang,
        Joonghyuk Shin,
        Jaesik Park,
\IEEEcompsocitemizethanks{\IEEEcompsocthanksitem Minguk Kang is with POSTECH, Pohang, Republic of Korea, 37673. Joonghyuk Shin, and Jaesik Park are with Seoul National University, Seoul, Republic of Korea, 08826
}
}

\IEEEtitleabstractindextext{%
\begin{abstract}
Generative Adversarial Network~(GAN) is one of the state-of-the-art generative models for realistic image synthesis. While training and evaluating GAN becomes increasingly important, the current GAN research ecosystem does not provide reliable benchmarks for which the evaluation is conducted consistently and fairly. Furthermore, because there are few validated GAN implementations, researchers devote considerable time to reproducing baselines. We study the taxonomy of GAN approaches and present a new open-source library named StudioGAN. StudioGAN supports 7 GAN architectures, 9 conditioning methods, 4 adversarial losses, 12 regularization modules, 3 differentiable augmentations, 7 evaluation metrics, and 5 evaluation backbones. 
With our training and evaluation protocol, we present a large-scale benchmark using various datasets (CIFAR10, ImageNet, AFHQv2, FFHQ, and Baby/Papa/Granpa-ImageNet) and 3 different evaluation backbones (InceptionV3, SwAV, and Swin Transformer). Unlike other benchmarks used in the GAN community, we train representative GANs, including BigGAN and StyleGAN series in a unified training pipeline and quantify generation performance with 7 evaluation metrics. The benchmark evaluates other cutting-edge generative models~(\textit{e.g.}, StyleGAN-XL, ADM, MaskGIT, and RQ-Transformer). StudioGAN provides GAN implementations, training, and evaluation scripts with the pre-trained weights. StudioGAN is available at~\MYhref[magenta]{https://github.com/POSTECH-CVLab/PyTorch-StudioGAN}{https://github.com/POSTECH-CVLab/PyTorch-StudioGAN}.

\end{abstract}

}
\maketitle
\IEEEdisplaynontitleabstractindextext

\IEEEpeerreviewmaketitle

\section{Introduction}
\label{sec:introduction}
\lettrine{G}{enerative} Adversarial Network (GAN)~\cite{Goodfellow2014GenerativeAN} is a well-known paradigm for approximating a real data distribution through an adversarial process. A GAN pipeline comprises a generator and a discriminator network. The discriminator tries to classify whether given samples are from the generator or the real data. On the other hand, the generator strives to generate realistic examples to induce the discriminator to misjudge the generated samples as real. By repeating the adversarial process, it is demonstrated that the generator can approximate the real data distribution~\cite{Goodfellow2014GenerativeAN}.

Since Radford~\etal~\cite{Radford2016UnsupervisedRL} showed that deep convolutional GAN (DCGAN) can generate high-quality natural images, GAN has been actively adopted for a variety of computer vision applications, including natural image synthesis~\cite{Radford2016UnsupervisedRL, nowozin2016f, Metz2017UnrolledGA, Zhao2016EnergybasedGA, lin2017softmax, Brock2019LargeSG, karras2020analyzing, karras2021alias, kang2021rebooting, kumari2022ensembling}, image-to-image translation~\cite{zhu2017unpaired, Wang2018HighResolutionIS, Choi_2018_CVPR, Liu_2019_ICCV, park2020cut, Richardson_2021_CVPR}, super-resolution imaging~\cite{Ledig_2017_CVPR, Wang2018ESRGANES, yuan2018unsupervised, guan2019srdgan}, and generative neural radiance field~\cite{Schwarz2020NEURIPS, piGAN2021, Niemeyer2021GIRAFFERS, Gu2021StyleNeRFAS, Zhou2021CIPS3DA3}.  Along with the increasing demand for developing a high-performance generative model, evaluating the quality of generated images has become an imperative procedure for GAN development. Inception score (IS)~\cite{Salimans2016ImprovedTF}, and Fr\'echet Inception Distance (FID)~\cite{Heusel2017GANsTB} are the representative metrics for the generative model evaluation, and follow-up studies adopt the two metrics for the comparison.

While researchers put much effort into properly evaluating GAN, there are still inappropriate training configurations, and evaluation procedures are inconsistent between proposed approaches. For instance, some software libraries show unexpected behavior in data processing procedures. A quantization error occurs when converting training images to an hdf5 format file for fast data I/O. Similarly, buggy image resizing libraries can cause aliasing artifacts in images, affecting the quality of generated images~\cite{parmar2021cleanfid, karras2021alias}. \newtext{The lack of elaborately designed training and evaluation protocols followed by reliable benchmark, impede the community's advancement in developing better generative models.}

Generative model evaluation is mainly concerned with IS~\cite{Salimans2016ImprovedTF} and FID~\cite{Heusel2017GANsTB} where a pre-trained InceptionV3~\cite{Szegedy2016RethinkingTI} is used to extract the features of images. As stated by Kynk{\"a}{\"a}nniemi~\etal~\cite{Kynkaanniemi2022TheRO}, FID can be hacked if information leakage from the InceptionV3 network takes place. \newtext{This implies the importance of comprehensive benchmark with various metrics and evaluation backbones, such as  IS, FID, Precision \& Recall~\cite{Kynknniemi2019ImprovedPA}, and Density \& Coverage~\cite{ferjad2020icml}, evaluated using different backbones, in order to thoroughly examine various aspects of models.}

In this work, we introduce a new software library, named \textbf{StudioGAN} which has implementations of 30 representative GANs, covering from DCGAN~\cite{Radford2016UnsupervisedRL} to StyleGAN3~\cite{karras2021alias} reproducing reported results in most cases. StudioGAN provides 7 GAN architectures, 9 conditioning methods, 4 adversarial losses, 12 regularization methods, and 7 evaluation metrics. StudioGAN allows training and evaluating representative GANs in the same environment, indeed with proper training and evaluation protocols for a fair comparison. 
Moreover, StudioGAN allows the change of evaluation backbone from InceptionV3 to modern networks, including SwAV~\cite{Caron2020UnsupervisedLO, Morozov2021OnSI}, DINO~\cite{caron2021emerging}, and Swin-T~\cite{liu2021swin}. As a result, we present an unparalleled amount of benchmark results instead of referring to the results from existing papers.

\newtext{Note that there are several papers~\cite{Zhang2019ConsistencyRF, gu2021vector, chang2022maskgit, lee2022autoregressive, rombach2022high} that provide benchmark tables of generative models. In these papers, authors contain the evaluation results from previous papers without being fully aware that the original papers often utilize different training and evaluation protocols. Correspondingly, these practices still result in unfairness due to the disparities in data preprocessing, the employed machine learning library, and some unintentional mistakes by researchers. Recent research by Parmar~\etal~\cite{parmar2021cleanfid} established protocols to train and evaluate GANs in a more controlled setup. While their primary goal is to establish a protocol for fairer comparison between generative models, the benchmark they present is limited to models with open-source implementations and includes a re-evaluation of pre-trained models. This motivates us to establish an extensive benchmark that is controlled from the scratch-training to fair and clear evaluation. By providing identical training and evaluating conditions, we aim to provide a playground where researchers can easily implement their new ideas, followed by an extensive, fairly computed benchmark.}

\brown{While the primary focus of our paper is to provide open-source implementations of GANs with an extensive benchmark, we also include a comparison between GANs and other generative models, such as diffusion models~\cite{ho2020denoising, song2021scorebased, vahdat2021score, dhariwal2021diffusion, dockhorn2022score} and auto-regressive models~\cite{esser2021taming, chang2022maskgit, lee2022autoregressive} in Section~\ref{evaluation_other_generative_models}. GANs noticeably differ from diffusion and auto-regressive models regarding the number of feed-forward processes for generating a single sample. Therefore, we present the inference speed and the number of model parameters to help analyze the strengths and weaknesses of each model. For diffusion models or auto-regressive models, we use official implementations and apply the same evaluation protocol used for GAN evaluation.}

In our extensive evaluation of GANs, we observe interesting findings on BigGAN, StyleGAN2, and StyleGAN3 characteristics. For example, we experimentally identify that StyleGAN2 tends to generate more diverse but lower quality images than BigGAN. However, we also witness that training StyleGAN2 and StyleGAN3 is more challenging than BigGAN if the targeted distribution has large inter-class variations. \brown{Furthermore, we discuss the potential hazards that can arise during a comparison with other models and add the future research direction at the end of the paper.} To the best of our knowledge, this is the first attempt to provide a fairly established benchmark for generative modeling on such a substantial scale.

In summary, this paper has the following contributions.
\begin{itemize}
    \item We present a taxonomy of GAN approaches based on five criteria (Sec.~\ref{sec:gan_taxonomy}).
    \item We pinpoint inappropriate practices that can cause poor generation results, reproducibility issues, and unfair evaluation (Sec.~\ref{sec:issues_in_training}). 
    \item We introduce StudioGAN, a software library that consists of 7 GAN architectures, 9 conditioning methods, 4 adversarial losses, 12 regularization modules, 3 differentiable augmentations, 7 evaluation metrics, and 5 evaluation backbones (Sec.~\ref{sec:studiogan}). Through carefully assembling provided modules, researchers can design diverse and reproducible GANs, including 30 existing GANs. 
    \item We provide large-scale evaluation benchmark for generative model development on standard and novel datasets (Sec.~\ref{sec:benchmark}). Some interesting findings from modern generative models are presented (Sec.~A4~and~\ref{evaluation_other_generative_models}).
\end{itemize}
\vspace{-2mm}
\section{Preliminary}
\label{sec:generatve_adversarial_network}
\noindent \textbf{Generative Adversarial Network.}
Goodfellow~\etal~\cite{Goodfellow2014GenerativeAN} proposed a framework for generative modeling based on a two-player minimax game. Generative Adversarial Network~(GAN) aims to estimate the distribution of an arbitrary data in high-dimensional space and consists of a generator and a discriminator network adversarial to each other. 

Specifically, the \emph{generator} $G: \mathcal{Z}\longrightarrow\mathcal{X}$ strives to map a latent variable $\rvz \sim p(\rvz)$ into a sample $G(\rvz)$ as if the sample is drawn from the real data distribution $p(\rvx)$. The \emph{discriminator} $D:\mathcal{X}\longrightarrow[0,1]$, on the other hand, tries to discriminate whether the given sample comes from the real data distribution or the generator, becoming the adversary to the generator. Mathematically, the generative and discriminative processes can be expressed as follows:
\begin{align*}
    \min_{G}\max_{D}~\mathbb{E}_{\rvx \sim {p(\rvx)}}[\log(D(\rvx))] + \mathbb{E}_{\rvz \sim {p(\rvz)}}[\log(1 - D(G(\rvz)))].
    \label{eq:eq1}\tag{1}
\end{align*}
\vspace{-2mm}

\noindent The adversarial training between the two networks converges to Nash equilibrium under the two assumptions: (1) the two models have enough capacity, and (2) the discriminator is allowed to reach its optimum against the generator at each step. Under these conditions, implicit distribution of the generator $p_{G(\rvz)}(\rvx)$ converges to the true data distribution $p(\rvx)$, and generator becomes an efficient sampler for the true data distribution. 

\vspace{2mm}
\noindent \textbf{Evaluating GANs.}
To quantify the generated image's quality without laborious human intervention, Inception Score~(\textbf{IS})~\cite{Salimans2016ImprovedTF} and Fr\'echet Inception Distance~(\textbf{FID})~\cite{Heusel2017GANsTB} were introduced and have been broadly adopted. Those metrics are computed on feature space, so all images involved in computing the metrics are mapped to the feature space using a pre-trained InceptionV3~\cite{Szegedy2016RethinkingTI} network. 
IS can be written as follows. 
\begin{align*}
    \text{IS}(\mX_{t}) &= \text{Score}(\mX_{t}, F = \text{InceptionV3}) \\
    &= \exp\bigg(\frac{1}{M}\sum_{i=1}^{M}\text{D}_{\text{KL}}\Big(p(y|\vx_{i})|\hat{p}(y)\Big)\bigg),
    \label{eq:IS}\tag{2}
\end{align*}
\vspace{-3mm}

\noindent where $\mX_{t} = \{\vx_{1},..., \vx_{M}\}$ is the image samples we target to evaluate, the posterior $p(y|\vx_{i})$ is estimated by $F$ (InceptionV3), $\text{D}_{\text{KL}}$ is Kullback–Leibler divergence, and $\hat{p}(y)$ is an empirical class distribution: $\frac{1}{M}\sum_{i=1}^{M}p(y|\vx_{i})$. FID is defined as follows.
\begin{align*}
    \text{FID}(\mX_{s}, \mX_{t}) &= \text{FD}(\mX_{s}, \mX_{t}, F = \text{InceptionV3}) \\
    &= \lVert \mu_{s} - \mu_{t} \rVert_{2}^{2} + \text{Tr}\Big(\Sigma_{s} + \Sigma_{t} -2(\Sigma_{s}\Sigma_{t})^{\frac{1}{2}}\Big),
    \label{eq:FID}\tag{3}
\end{align*}
\vspace{-3mm}

\noindent where $\mu$ and $\Sigma$ are the mean vector and covariance matrix of the features from $F$, and the subscripts $s$ and $t$ indicate the source and target, respectively.

\section{Taxonomy of GAN approaches}
\label{sec:gan_taxonomy}
\begin{figure*}[!ht]
    \centering
    \includegraphics[width=0.95\linewidth]{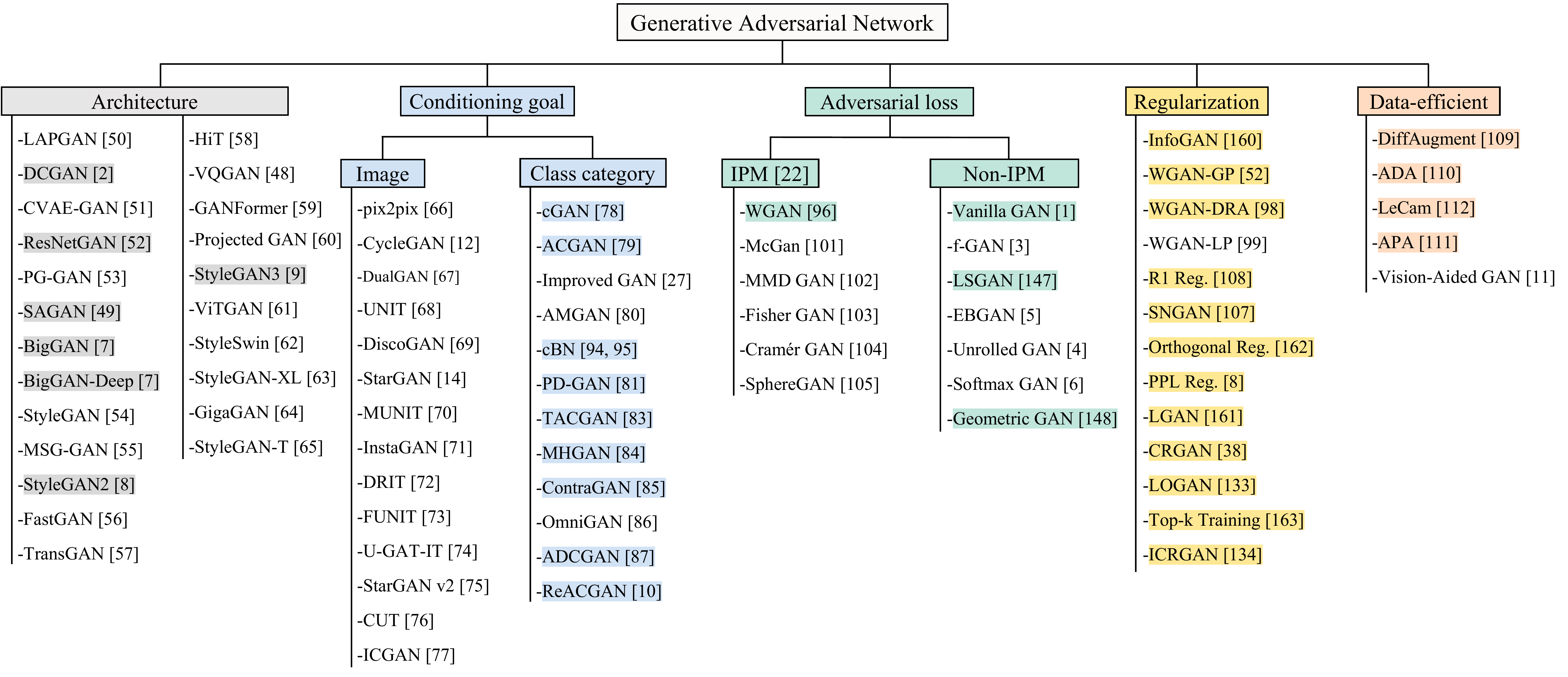}
    \vspace{-3mm}
    \caption{Taxonomy of GAN. The taxonomy comprises five divisions: architecture, conditioning goal, adversarial loss, regularization, and data-efficient training techniques. The colored elements indicate the models implemented in our proposed StudioGAN platform. StudioGAN provides 7 GAN architectures, 9 conditioning methods, 4 adversarial losses, 12 regularizations, and 5 data-efficient training methods. We will evaluate multiple GANs using 7 evaluation metrics (IS~\cite{Salimans2016ImprovedTF}, FID~\cite{Heusel2017GANsTB}, Precision~\cite{Kynknniemi2019ImprovedPA}, Recall~\cite{Kynknniemi2019ImprovedPA}, Density~\cite{ferjad2020icml}, Coverage~\cite{ferjad2020icml}, and Intra-class FID~\cite{Zhang2019SelfAttentionGA}) in section~\ref{sec:benchmark}.}
    \label{fig:taxonomy}
    \vspace{-2mm}
\end{figure*}

Since the invention of the GAN framework, enormous efforts have been made to utilize GANs for realistic data generation. We present a GAN taxonomy based on five divisions: architecture, conditioning goal, adversarial loss, regularization, and data-efficient training. Figure~\ref{fig:taxonomy} shows the taxonomy of representative approaches.

\subsection{GAN architecture}
\label{gan_architecture}
When GAN was first introduced to the field of machine learning, training GAN was known to be extremely difficult. Dynamics between the generator and the discriminator often undergo a mode-collapse problem for unknown reasons, resulting in meaningless outputs. The first prominent results in GAN development came from adopting a deep convolutional neural network as a backbone architecture~(DCGAN)~\cite{Radford2016UnsupervisedRL} instead of multi-layer perceptron~\cite{Goodfellow2014GenerativeAN}. Although the quality is still far from being realistic, DCGAN is at least capable of producing recognizable images and significantly stabilizes the dynamics between the generator and discriminator.
As deep learning and computing acceleration technologies advance, GAN benefits from larger and more complex network architectures~\cite{denton2015deep, bao2017cvae, Gulrajani2017ImprovedTO, Karras2018ProgressiveGO, Zhang2019SelfAttentionGA, Brock2019LargeSG, karras2019style, karnewar2020msg, karras2020analyzing, liu2020towards, jiang2021transgan, zhao2021improved, karras2021alias, esser2021taming, hudson2021generative, Sauer2021NEURIPS, lee2022vitgan, zhang2022styleswin, sauer2022stylegan, kang2023gigagan}. Gulrajani~\etal~\cite{Gulrajani2017ImprovedTO} develop a ResNet-style generator and discriminator. Zhang~\etal~\cite{Zhang2019SelfAttentionGA} show that adding self-attention layers into GAN can improve image generation performance (SAGAN). Brock~\etal~\cite{Brock2019LargeSG} scale up the ResNet-style architecture with hierarchical embedding~(BigGAN and BigGAN-Deep). Apart from the above stream, Karras~\etal have proposed a series of GAN architectures starting from Progressive GAN~\cite{Karras2018ProgressiveGO}, to StyleGAN~\cite{karras2019style}, StyleGAN2~\cite{karras2020analyzing}, and StyleGAN3~\cite{karras2021alias}. The StyleGAN family specializes in generating images with a low inter-class variation using more inductive biases in the architecture structure than SAGAN and BigGAN. \newtext{Beyond the narrow distribution modeling capabilities of StyleGAN-family models, Kang~\etal~\cite{kang2023gigagan} and Sauer~\etal~\cite{Sauer2023ARXIV} develop GigaGAN and StyleGAN-T, featuring much bigger model capacity, and demonstrate GAN's potential for a more generic image synthesis task, \textit{i.g,} text-to-image synthesis.}

\subsection{Conditional image generation}
\label{conditional_image_generation_taxonomy}
Conditional image generation is a task to synthesize an intended image provided that conditional information is given. The conditional information could be an image~\cite{isola_i2i, zhu2017unpaired, yi2017dualgan, liu2017unsupervised, kim2017learning, Choi_2018_CVPR, huang2018multimodal, mo2019instagan, lee2018diverse, liu2019few, Kim2020U-GAT-IT, choi2020starganv2, park2020contrastive, casanova2021instanceconditioned}, a categorical label~\cite{Mirza2014ConditionalGA, Odena2017ConditionalIS, Salimans2016ImprovedTF, zhou2018activation, Miyato2018cGANsWP, Siarohin2019WhiteningAC, NIPS2019_8414, kavalerov2021multi, kang2020contragan, zhou2020omni, hou2021cgans, kang2021rebooting}, or a text~\cite{xu2018attngan, zhu2019dm, zhang2021cross, ramesh2021zero, ramesh2022hierarchical, saharia2022photorealistic}. With GAN's surging popularity, conditional GAN~(cGAN) has become a representative framework for conditional image generation, and the cGANs specialized in each conditional information are being studied as separate research topics. 

The majority of conditioning methods for the generator are conducted by conditional batch normalization (cBN)~\cite{Dumoulin2017ALR, de_Vries} or simple concatenation operation. Unlike the generator, the conditioning techniques for the discriminator are developed based on objective functions accompanied by a slight modification of the architecture. ACGAN~\cite{Odena2017ConditionalIS} adopts an auxiliary classifier on top of the discriminator and tries to minimize the adversarial loss and classification objective together. PD-GAN (Projection Discriminator GAN)~\cite{Miyato2018cGANsWP} combines projected categorical embeddings with discriminator's embeddings, so adversarial training is performed both on the discriminative process and on the categorical classification. ReACGAN~\cite{kang2021rebooting} succeeds in the spirit of ACGAN and adopts data-to-data cross-entropy (D2D-CE) loss for class conditioning. The primary motivation for ReACGAN invention is ACGAN's gradient exploding problem, and the exploding issue can be significantly alleviated by normalizing the discriminator's features onto a unit hypersphere. 

\subsection{Adversarial losses and regularizations}
\label{adversarial_losses_and_regularizations_taxonomy}
The unstable nature of GAN dynamics is a major obstacle in high-quality image generation. To unearth the source of the instability, many researchers strive to provide interpretations of the unstable training issue and propose prescriptions in the form of adversarial loss and regularization. Arjovsky~\etal~\cite{Arjovsky2017WassersteinG, Arjovsky2017TowardsPM} state that this unstable characteristic attributes to the inherent gradient vanishing problem of the vanilla GAN loss. According to their papers, a vanilla GAN, which tries to minimize the Jensen-Shannon divergence between real and fake distributions, cannot provide helpful gradient signals to the discriminator if the distribution of fake images is far away from the real data distribution. To detour the gradient vanishing problem, Arjovsky~\etal propose WGAN that minimizes 1-Wasserstein distance instead of the Jessen-Shannon criterion. Since training WGAN requires restricting the discriminator to a 1-Lipschitz function, several WGAN variants are proposed with the change of imposing the restriction: WGAN-GP~\cite{Gulrajani2017ImprovedTO}, WGAN-DRA~\cite{Kodali2018OnCA}, and WGAN-LP~\cite{petzka2017regularization}. On the other side, GANs based on Integral Probability Measure~(IPM)~\cite{muller1997integral} are proposed by many researchers. WGAN~\cite{Arjovsky2017WassersteinG}, McGAN~\cite{mroueh2017mcgan}, MMD GAN~\cite{li2017mmd}, Fisher GAN~\cite{mroueh2017fisher}, Cram\'er GAN~\cite{bellemare2017cramer}, and SphereGAN~\cite{park2020spheregan} are representative IPM-based GANs that require a restricted discriminator architecture. However, those models cannot produce successful results on large-scale image datasets, such as ImageNet~\cite{Deng2009ImageNetAL}. Meanwhile, Miyato~\etal~\cite{Miyato2018SpectralNF} show successful results in ImageNet generation by devising spectral normalization, which bounds the gradient w.r.t an input image to 1.0. Consequently, modern GANs can be divided into two streams: (1) SNGAN~\cite{Miyato2018SpectralNF}, SAGAN~\cite{Zhang2019SelfAttentionGA}, BigGAN~\cite{Brock2019LargeSG}, and ReACGAN~\cite{kang2021rebooting} that require spectral normalization for stable training and (2) StyleGAN2~\cite{karras2020analyzing}, and StyleGAN3~\cite{karras2021alias} that use R1 regularization~\cite{Mescheder2018ICML} which also shrinks the norm of gradient significantly.

\subsection{Techniques for data-efficient Training}
\label{data_efficient_training_taxonomy}
Brock~\etal~\cite{Brock2019LargeSG} found that the discriminator of GAN tends to memorize the training dataset and provide an authenticity score for a given image, disregarding the realism of the generated images. This observation implies that if the training dataset is not enough, the discriminator can quickly memorize the training dataset, accelerating the training collapse. To this empirical discovery, data-efficient training has become one of the popular research trends in the GAN community. Researchers apply data augmentations on real and fake images to prevent the discriminator's overfitting. Differentiable augmentation (DiffAug)~\cite{zhao2020differentiable}, adaptive discriminator augmentation (ADA)~\cite{Karras2020TrainingGA}, and adaptive pseudo augmentation (APA)~\cite{deceived2021} are some notable studies that deal with data augmentations for data-efficient learning. Unlike the augmentation-based approaches, Tseng~\etal presents LeCam regularization~\cite{lecam2021}, which transforms the WGAN training procedure to minimize LeCam divergence under mild assumptions. Since LeCam divergence is known to be more robust under the limited data situation than other f-divergence variants, GAN-imposed LeCam is expected to be data-efficient. 
\begin{figure*}[ht]
    \centering
    \includegraphics[width=0.95\linewidth]{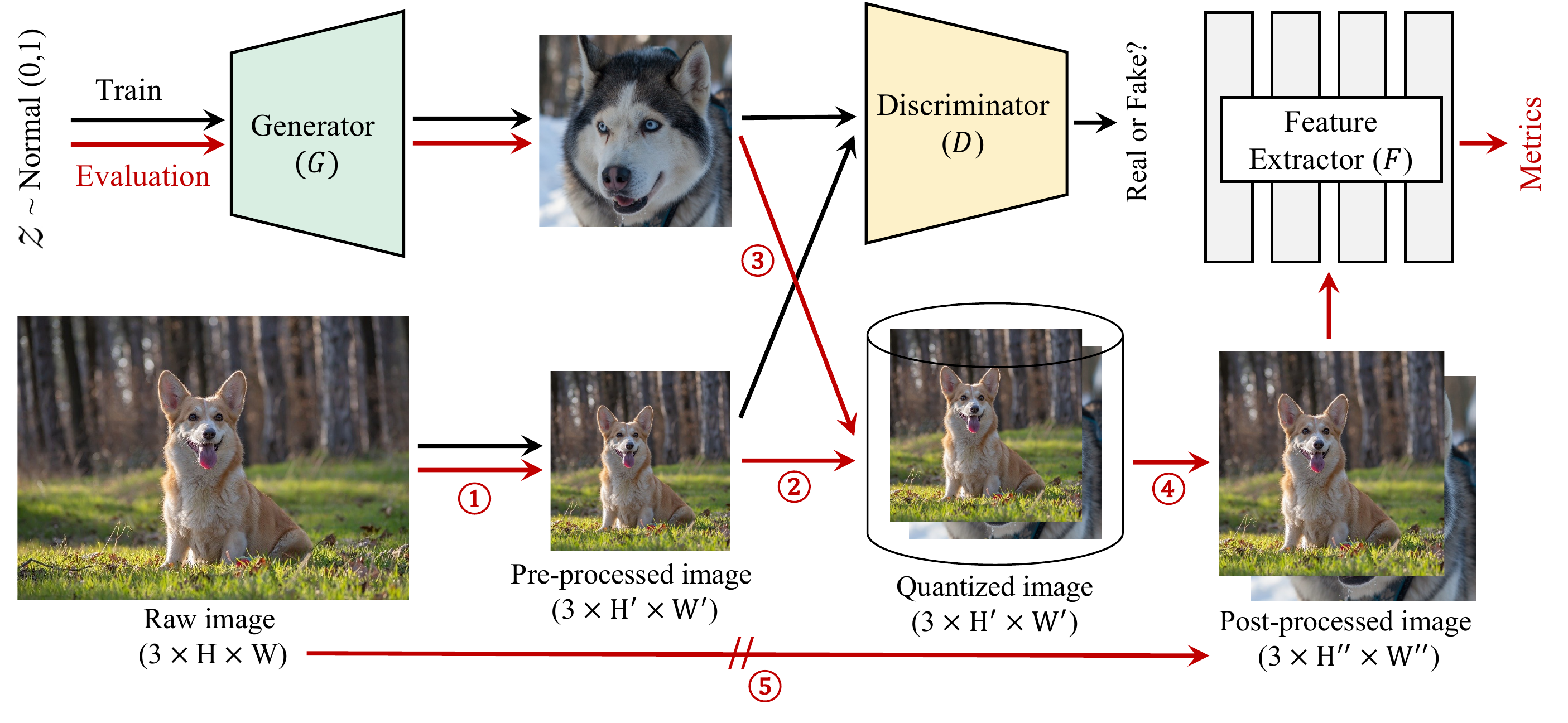}
    \vspace{-5.5mm}
    \caption{The schematic describes how real and fake images are processed during training and evaluation. Resizing and normalization operations are applied to the raw image in the {\small \textcircled{\scriptsize 1}}  procedure. Then, the pre-processed and generated images are fed into the discriminator to perform adversarial training. When evaluating GAN, it is recommended to quantize both real and generated images to 8-bit representation to save images into a disk ({\small \textcircled{\scriptsize 2}} and {\small \textcircled{\scriptsize 3}}) as stated in \cite{parmar2021cleanfid}. Finally, the quantized images are resized and normalized before being fed to the evaluation backbone~({\small \textcircled{\scriptsize 4}}). We suggest using an architecture-friendly anti-aliasing resizer instead of directly adopting a high-quality resizing filter. Note that skipping the procedures {\small \textcircled{\scriptsize 1}} and {\small \textcircled{\scriptsize 2}} and processing the raw image using {\small \textcircled{\scriptsize 5}} procedure should not be allowed since it will make a discrepancy between the generator's ideal distribution and the real data distribution so leading to biased evaluation.}
    \label{fig:pipeline}
    \vspace{-2mm}
\end{figure*}

\section{Inappropriate practices}
\label{sec:issues_in_training}
GAN approaches have been improved remarkably. However, there are inappropriate practices in data preprocessing and evaluation of GAN that can be boiled down to four points as follows:
\begin{itemize}
\item Adopting an improper image resizer and applying an unnecessary quantization process during the preprocessing step. (Sec.~\ref{sec:aliasing_quantization})
\item The image resizer applied before the evaluation backbone~(Feature Extractor, $F$) is not the same as the resizer used for training the evaluation backbone. (Sec.~\ref{sec:pil_bicubic_is_good})
\item Evaluating GAN only with IS and FID that are based on the pre-trained InceptionV3. (Sec.~\ref{sec:backbone} and Sec.~\ref{sec:metrics})
\item Lacking the description of the type of normalization operation in batch normalization layers, although it can significantly affect the results. (Sec.~\ref{sec:normalization})
\end{itemize}
Following the conclusions from these sections, we suggest protocols for training and evaluating GANs in Sec.~\ref{sec:protocol_suggestion}.

\subsection{When processing training images}
\label{sec:aliasing_quantization}
The improper practices in preprocessing image data can be summarized into three points: (1) inherent aliasing artifacts in the training dataset due to improper resizing functions, (2) sampling error occurred by adopting a poor interpolation filter for both up- and down-sampling, and (3) quantization error that raises when converting a training dataset into other data I/O formats (\textit{i.e.}, HDF5).

We highlight less attention to avoiding aliasing artifacts in training images. \emptext{As Parmar~\etal~\cite{parmar2021cleanfid} and Karras~\etal stated~\cite{karras2021alias}, processing training dataset using not only anti-aliasing but also a high-quality resizer is necessary for realistic image generation.} However, most GAN approaches do not report what kind of resizer is used to preprocess images. 
By investigating original implementations, we identify that many methods use a \textbf{bilinear interpolation} for the image resizer, so resized images are occasionally of poor quality and exhibit aliasing artifacts. Furthermore, some studies utilize Tiny-ImageNet~\cite{Tiny} dataset which contains aliasing and noise artifacts (Figure~A1).

This work proposes to use either \textbf{bicubic}~\cite{Keys1981CubicCI} or \textbf{Lanczos~\cite{Turkowski1990FiltersFC} interpolation} for image up- and down-sampling to preprocess images. \emptext{Interestingly, Parmar~\etal~\cite{parmar2021cleanfid} discovered that resizers implemented in popular TensorFlow~\cite{tensorflow2015-whitepaper}, PyTorch~\cite{NEURIPS2019_9015}, and OpenCV~\cite{opencv_library} libraries could cause aliasing artifacts in resulting images. According to this finding, we recommend using python imaging library (PIL)~\cite{umesh2012image}, such as PIL.BICUBIC or PIL.LANCZOS resizer.}

Lastly, we discover that some popular approaches have an unnecessary or improper quantization process during the data preprocessing step. For example, software platforms built upon PyTorch-BigGAN~\cite{brock2018large} convert an image from PNG or JPEG to HDF5 format for fast I/O. The software platforms perform unnecessary normalization and wrong quantization operations during the conversion, raising quantization errors. Practitioners should skip the unnecessary normalization process and use a proper quantization operation to avoid the quantization error.

For example, if we use NumPy~\cite{harris2020array} framework for uint8 quantization, it is necessary to add 0.5 to each pixel since NumPy adopts the round-down operation for uint8 quantization. We express one of the possible solutions for image quantization in the NumPy environment as follows:
\begin{align*}
    \rvx_{\text{q}} = \text{clip}(127.5\rvx_{\text{n}} + 128, 0, 255),
    \label{eq:eq4}\tag{4}
\end{align*}
where clip$(\cdot,~\text{lower bound},~\text{upper bound})$ is a clipping function, $\rvx_{\text{q}}$ is an image quantized in unit8 type, and $\rvx_{\text{n}}$ is an normalized image with a pixel range of [-1.0, 1.0].

\subsection{When processing generated images}
\label{sec:pil_bicubic_is_good}
We observe another hazard during the GAN evaluation. It attributes to the choice of image resizer applied to the generated images when feeding them to the evaluation backbone (\textit{i.e.}, TensorFlow-InceptionV3 to acquire FID and IS scores). Regarding this issue, Parmar~\etal~\cite{parmar2021cleanfid} confirmed that the improper resizing operation provokes aliasing artifacts, and they suggest clean-metric (clean-FID) by using the PIL.BICUBIC resizer~\cite{umesh2012image}.

In this work, we suggest using \textbf{backbone friendly resizer} for the evaluation. Our key finding is that the proper image resizer depends on the backbone network. 
For instance, feeding bicubic-interpolated images to the InceptionV3 network is improper since the InceptionV3 network was originally trained using a different resizer. 
Adopting the same resize filter used in training the TensorFlow InceptionV3 network cannot be the solution. This is because the InceptionV3 network was trained using the direct route \circled{5} as depicted in Figure~\ref{fig:pipeline} while real images for GAN should be resized using the anti-aliasing bicubic/Lanczos resizer through the route \circled{1}. This fact makes the entire data processing procedures of a recognition model and GAN inevitably different. \emptext{Also, using a resizer implemented in TensorFlow for \circled{5} method can induce aliasing artifacts as discovered by Parmar~\etal~\cite{parmar2021cleanfid}, so both generated and real images may lose details. For that matter, we recommend using a resizer in the PIL library as suggested by Parmar~\etal~\cite{parmar2021cleanfid}, but with a different interpolation filter applied.}

 To identify which resizer is more suitable for the specific backbone, we experiment with the assumption that the resizer is better if it allows better image classification accuracy. If the evaluation backbone shows better recognition ability, the backbone is likely to extract more disentangled features. 
The assumption correlates with the reason why InceptionV3 network is widely applied as an evaluation backbone network. The authors of IS~\cite{Salimans2016ImprovedTF} selected InceptionV3~\cite{Szegedy2016RethinkingTI} as a feature extractor since it was one of the cutting-edge models at that time. 

\begin{table*}[!htp]
\caption{Experimental results to check IS~\cite{Salimans2016ImprovedTF} and classification accuracies of an evaluation backbone according to the resizer type. We use ImageNet~\cite{Deng2009ImageNetAL} valid dataset to measure the metrics. We resize a raw image using PIL.LANCZOS resizer~(procedure {\small \textcircled{\scriptsize 1}}) and quantize the image~(procedure {\small \textcircled{\scriptsize 2}}). The numbers shown on the left side of the table~(128, 256, and 512) indicate the resolution of the quantized image, and each image is resized once again to adjust the resolution for the evaluation backbone~(procedure {\small \textcircled{\scriptsize 4}}). The final resolution depends on the types of evaluation backbones: 299 for TensorFlow-InceptionV3~\cite{Szegedy2016RethinkingTI}, 224 for PyTorch-SwAV~\cite{Caron2020UnsupervisedLO}, and 224 for PyTorch-Swin-T~\cite{liu2021swin}. We measure Inception Score (IS), SwAV Score (SS), Swin-Transformer Score (TS), Top-1 accuracy, and Top-5 accuracy. SS and TS have the same conceptual scores as IS, except that the InceptionV3 backbone is replaced with SwAV and Swin-T, respectively. The numbers in bold-faced denote the best performances, and \newtext{yellow colored blocks indicate architecture-friendly resizers for each evaluation backbone.}}
\vspace{-2.5mm}
\centering
\resizebox{0.93\textwidth}{!}{
\begin{tabular}{clccccccccccc}
\cmidrule[0.75pt]{2-13}
& \multirow{2}*[-0.5ex]{\textsc{Resizer}~$\circled4$} & \multicolumn{3}{c}{TensorFlow-InceptionV3~\cite{Szegedy2016RethinkingTI}} & & \multicolumn{3}{c}{PyTorch-SwAV~\cite{Caron2020UnsupervisedLO}}  & & \multicolumn{3}{c}{PyTorch-Swin-T~\cite{liu2021swin}} \\\cmidrule[0.75pt]{3-5}\cmidrule[0.75pt]{7-9}\cmidrule[0.75pt]{11-13}

&  & \multirow{1}*[0.0ex]{\textsc{IS}~$\uparrow$} & \multirow{1}*[0.0ex]{\textsc{Top-1}~$\uparrow$} & \multirow{1}*[0.0ex]{\textsc{Top-5}~$\uparrow$} & & \multirow{1}*[0.0ex]{\textsc{SS}~$\uparrow$} & \multirow{1}*[0.0ex]{\textsc{Top-1}~$\uparrow$} & \multirow{1}*[0.0ex]{\textsc{Top-5}~$\uparrow$}  & & \multirow{1}*[0.0ex]{\textsc{TS}~$\uparrow$} & \multirow{1}*[0.0ex]{\textsc{Top-1}~$\uparrow$} & \multirow{1}*[0.0ex]{\textsc{Top-5}~$\uparrow$}

\\\cmidrule[0.75pt]{2-5}\cmidrule[0.75pt]{7-9}\cmidrule[0.75pt]{11-13}
\parbox[t]{0.5mm}{\multirow{4}{*}{\rotatebox[origin=c]{90}{\text{128}}}}
& \textsc{Nearest}&  167.06  &  67.88   & 88.07 & & 153.46  & 59.71 & 82.80 & & \textbf{106.32} & 75.35 & 92.79 \\
& \textsc{Bilinear} &  \cellcolor{yellow!40}\textbf{182.57}  &  \cellcolor{yellow!40}\textbf{72.03}   & \cellcolor{yellow!40}\textbf{90.09} &  & \cellcolor{yellow!40}\textbf{276.67} & \cellcolor{yellow!40}\textbf{69.78} & \cellcolor{yellow!40}\textbf{89.46} & & 61.88 & 78.21 & 94.81\\
& \textsc{Bicubic}~\cite{Keys1981CubicCI} & 178.45 & 71.84 & 89.80 & & 273.76 & 69.26 & 88.99 & & \cellcolor{yellow!40}71.70 & \cellcolor{yellow!40}78.93 & \cellcolor{yellow!40}95.16\\ 
& \textsc{Lanczos}~\cite{Turkowski1990FiltersFC} & 173.79 & 71.42 & 89.53 & & 254.81 & 68.36 & 88.37 & & 74.55 & \textbf{79.65} & \textbf{95.27}\\ 
\cmidrule[0.75pt]{2-13}
\parbox[t]{0.5mm}{\multirow{4}{*}{\rotatebox[origin=c]{90}{\text{256}}}}
& \textsc{Nearest} & \textbf{260.17} & 77.65  & 93.98 & & 299.90 & 72.79 & 91.15 & & \textbf{213.70} & 84.87 & 97.45 \\
& \textsc{Bilinear} & \cellcolor{yellow!40}259.55 &  \cellcolor{yellow!40}78.11  & \cellcolor{yellow!40}94.08 &  & \cellcolor{yellow!40}\textbf{327.90} & \cellcolor{yellow!40}\textbf{74.39} & \cellcolor{yellow!40}\textbf{92.06} &  & 154.52 & 84.14 & 97.13\\
& \textsc{Bicubic}~\cite{Keys1981CubicCI} & 259.99 & 78.26 & \textbf{94.16} & & 321.78 & 74.04 & 91.85 & & \cellcolor{yellow!40}194.74 & \cellcolor{yellow!40}\textbf{85.13} & \cellcolor{yellow!40}\textbf{97.51} \\ 
& \textsc{Lanczos}~\cite{Turkowski1990FiltersFC} & 259.31 & \textbf{78.27} & 94.13 & & 316.78 & 73.68 & 91.64 & & 202.29 & 85.00 & 97.39\\ 
\cmidrule[0.75pt]{2-13}
\parbox[t]{0.5mm}{\multirow{4}{*}{\rotatebox[origin=c]{90}{\text{512}}}}
& \textsc{Nearest}& 262.30 & 77.89 & 94.00 & & 302.56 & 72.57 & 91.22 & & \textbf{212.28} & 84.51 & 97.29 \\
& \textsc{Bilinear} & \cellcolor{yellow!40}\textbf{265.22} & \cellcolor{yellow!40} \textbf{78.48} & \cellcolor{yellow!40}\textbf{94.30} &  & \cellcolor{yellow!40}\textbf{341.60} & \cellcolor{yellow!40}\textbf{74.66} & \cellcolor{yellow!40}\textbf{92.25} &  & 156.38 & 84.28 & 97.18\\
& \textsc{Bicubic}~\cite{Keys1981CubicCI} & 265.14 & 78.44 & \textbf{94.30} & & 334.83 & 74.34 & 92.04 & & \cellcolor{yellow!40}192.10 & \cellcolor{yellow!40}\textbf{85.12} & \cellcolor{yellow!40}\textbf{97.48}\\ 
& \textsc{Lanczos}~\cite{Turkowski1990FiltersFC} & 264.32 & 78.41 & 94.26 & & 326.64 & 73.99 & 91.78 & & 201.03 & \textbf{85.12} & 97.44 \\ 
\cmidrule[0.75pt]{2-13}
\end{tabular}}
\label{table:post_processing}
\vspace{-4mm}
\end{table*} 

In Table~\ref{table:post_processing}, we measure Inception Score~\cite{Salimans2016ImprovedTF}, Top-1, and Top-5 image classification accuracies on ImageNet~\cite{Deng2009ImageNetAL} validation dataset with various image resizers applied for the procedure~\circled{4} in Figure~\ref{fig:pipeline}. Interestingly, the results are sensitive depending on the choice of image resizer, which indicates GAN comparison should be carefully designed. 
Experimental results also show that PIL.BILINEAR exhibits the coherent Inception Score (IS) and recognition accuracies in most cases for TensorFlow-InceptionV3~\cite{Szegedy2016RethinkingTI}.
To see the effects of image resizer in modern backbone networks, we perform the same experiment with PyTorch-SwAV~\cite{Caron2020UnsupervisedLO} and PyTorch-Swin-T~\cite{liu2021swin}. In the case of PyTorch-SwAV, PIL.BILINEAR is considered a reasonable choice. One worth mentioning point is that processed images using PIL.NEARNEST resizer gives exceptionally high Swin Transformer Scores~(TS). For this reason, we weight Top-1 and Top-5 accuracies to select the proper resizer for Swin-T and decide to use PIL.BICUBIC resizer. 

\subsection{Backbone networks for GAN evaluation}
\label{sec:backbone} 
Conventionally, almost all scores for generative model evaluation are computed using the features extracted from the TensorFlow-InceptionV3 network. Since 
Fr\'echet Distance~(FD) measured using TensorFlow-InceptionV3 network~(FID) is sample-inefficient~\cite{Morozov2021OnSI} and is biased towards ImageNet classification task~\cite{Kynkaanniemi2022TheRO}, it sometimes does not capture the discrepancy between distributions well. Therefore, we investigate GAN evaluation using a modern recognition backbone (Swin-T~\cite{liu2021swin}) and self-supervised backbones (SwAV~\cite{Caron2020UnsupervisedLO} and DINO~\cite{caron2021emerging}). 

\begin{figure}[!t]
    \centering
    \includegraphics[width=0.99\linewidth]{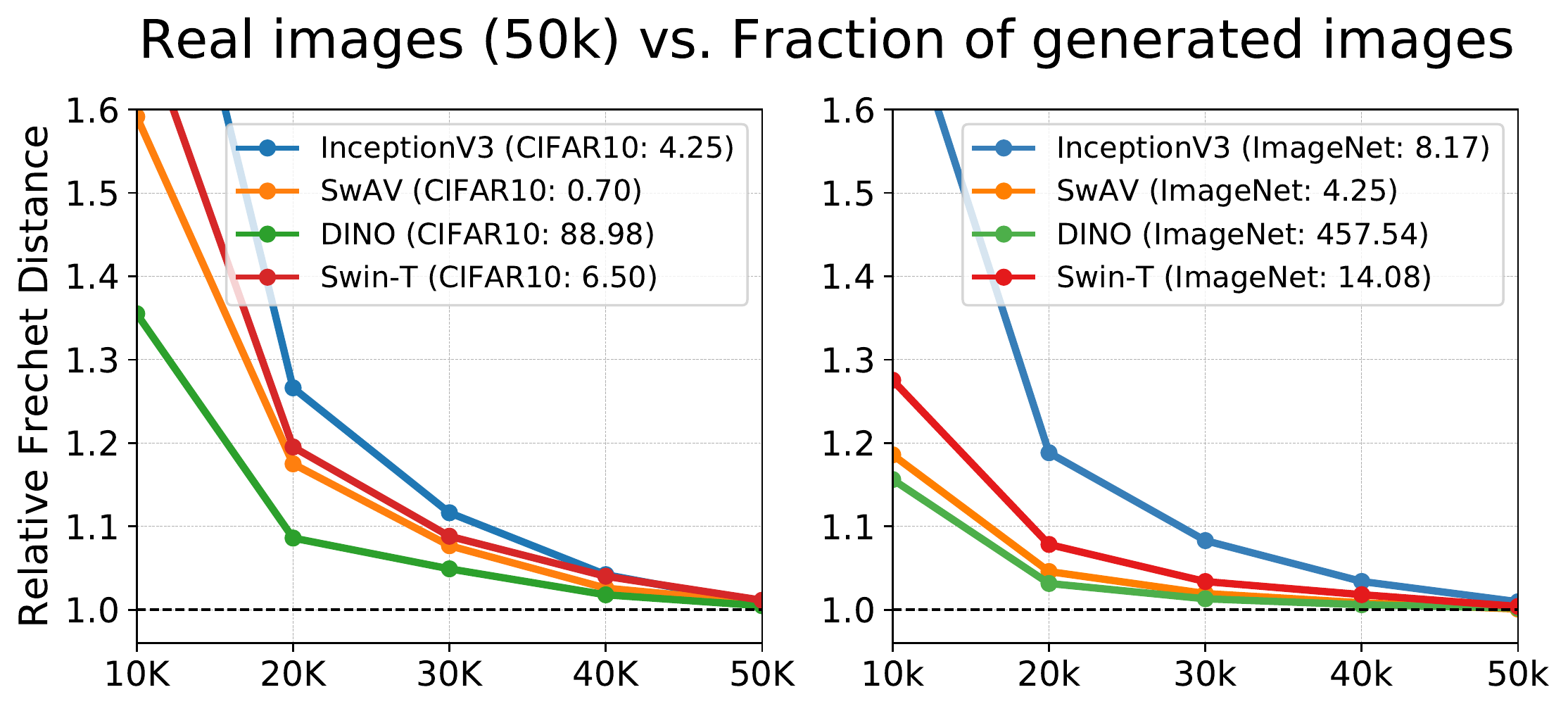}
    \vspace{-2mm}
    \caption{Trends of relative Fr\'echet Distance~(FD) between real images (50k) and a fraction of generated images. To calculate the target FD, we sample 50k generated images using BigGAN trained using StudioGAN, and the computed FD is used as the reference for calculating the relative FD. The reference values are written in the legends. The figures show how efficiently each backbone can measure FD.}
    \label{fig:real_vs_fake_fid}
    \vspace{-2mm}
\end{figure}

Let $\mX_{S}$ and $\mX_{T}$ be the source and target datasets, and $\mX_{t} \subset \mX_{T}$ be a subset of the target dataset. Then, \textbf{relative-FD} can be written as $\frac{\text{FD}(\mX_{S}, \mX_{t}, F)}{\text{FD}(\mX_{S}, \mX_{T}, F)}$ to analyze the sample efficiency of FD metrics across different backbones $F$. Figure~\ref{fig:real_vs_fake_fid} shows relative-FD computed using pre-trained BigGAN models~\cite{Brock2019LargeSG} on CIFAR10~\cite{Krizhevsky2009LearningML} and ImageNet~\cite{Deng2009ImageNetAL}. Lower is better because it indicates we can evaluate small number of generated images reasonably. The figure implies that the evaluated FDs using the InceptionV3 network are sample-inefficient compared to the other networks, as stated in~\cite{binkowski2018demystifying}. Note that SwAV, DINO, and Swin-T give relatively sample-efficient results on CIFAR10 and ImageNet datasets. 

\begin{figure}[!t]
    \centering
    \includegraphics[width=0.99\linewidth]{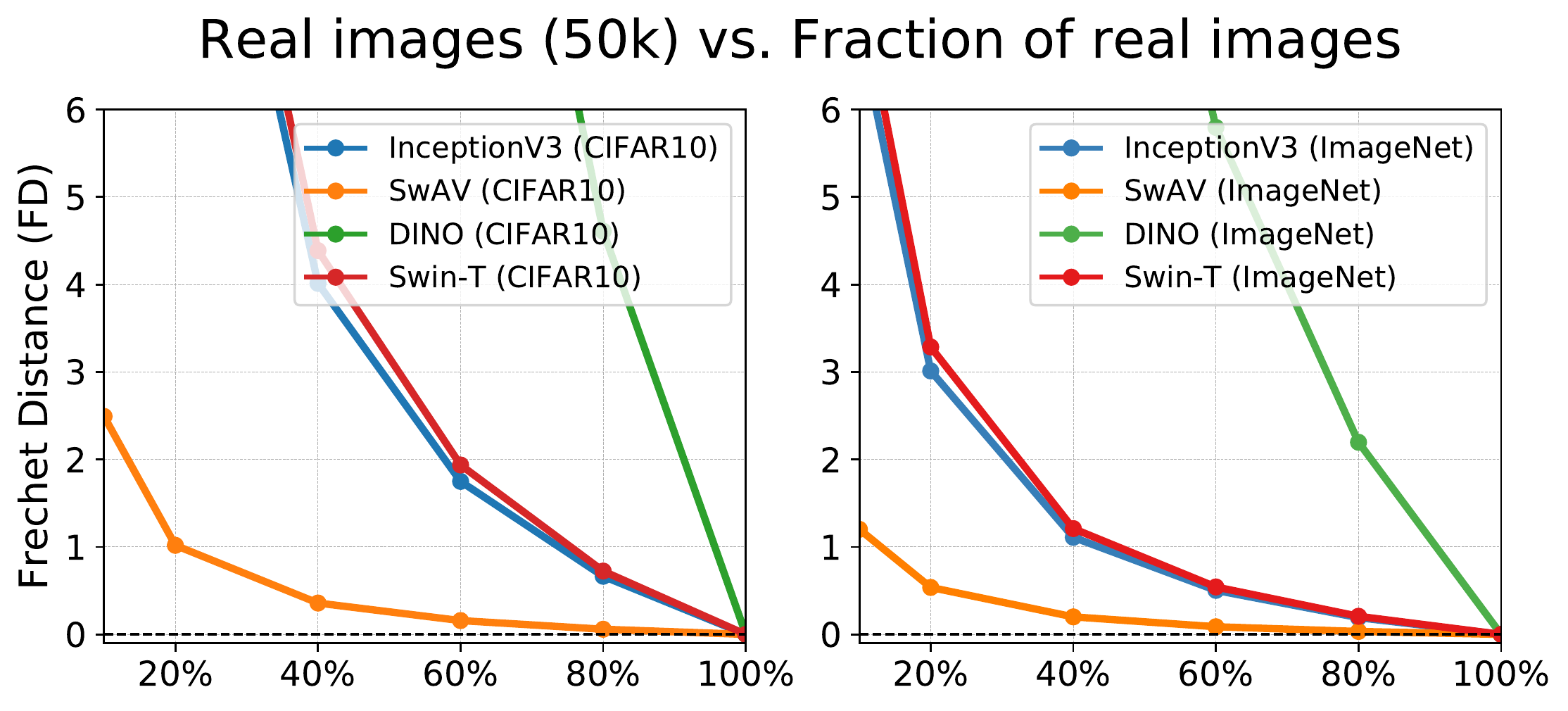}
    \vspace{-2mm}
    \caption{Trends of Fr\'echet Distance~(FD) between real images (50k) and a fraction of the real images. The Figures show how efficiently each backbone can measure real-to-real-FD.}
    \label{fig:real_vs_real_fid}
    \vspace{-3mm}
\end{figure}

To explore the effectiveness of FD, we visualize trends of \textbf{real-to-real-FD}s as the number of samples increases in Figure~\ref{fig:real_vs_real_fid}. Using various backbone networks, we measure the feature distance between real images and the fraction of real images. Using the evaluation backbones, we calculate real images' moments (mean and variance). Then, we sample fractions of the real images and calculate FDs. The real-to-real-FD can be written as FD$(\mX_{S}, \mX_{s}, F)$, where $\mX_{s} \subset \mX_{S}.$ Unlike the results in Figure~\ref{fig:real_vs_fake_fid}, Figure~\ref{fig:real_vs_real_fid} shows that DINO does not measure the real-to-real-FD efficiently. We speculate this phenomenon attributes to the property of DINO capturing the layout of a scene. As a result, we believe that \textbf{SwAV, and Swin-T} are good backbone candidates for GAN evaluation in addition to the InceptionV3 network.


\subsection{Evaluation metrics}
\label{sec:metrics}

Currently, GAN evaluation hugely depends on the value of Fr\'echet Inception Distance (FID)~\cite{Heusel2017GANsTB}. While FID has been verified by subsequent papers~\cite{Miyato2018SpectralNF, Brock2019LargeSG, karras2020analyzing, karras2021alias}, it has several limitations that FID informs the general ability of a generative model instead of the model's specialties~\cite{sajjadi2018assessing, Kynknniemi2019ImprovedPA, ferjad2020icml}. Also, FID is sometimes not coherent with human judgment~\cite{liu2018improved, Morozov2021OnSI, Sauer2021NEURIPS}. To complement FID, Precision \& Recall metrics~\cite{sajjadi2018assessing, Kynknniemi2019ImprovedPA, ferjad2020icml} are proposed. However, those metrics are not as widely used as FID, and the criterion for computing the metrics has not been established concretely~\cite{sauer2022stylegan}. Therefore, we emphasize that evaluating GAN using various metrics with concrete criteria is necessary to analyze the various aspect of GAN.

\subsection{Normalization operations}
\label{sec:normalization}
\begin{table}[t]
    \caption{Experiments to check whether the normalization operation of batch normalization layers can affect the evaluation results. We utilize a pre-trained BigGAN~\cite{Brock2019LargeSG} and evaluate the model by changing a normalization operation.}
    \vspace{-0.3cm}
    \centering
    \resizebox{0.48\textwidth}{!}{
    \begin{tabular}{clrrrr}
    \cmidrule[1.0pt]{2-6}
    \parbox[t]{2mm}{\multirow{5}{*}{\rotatebox[origin=c]{90}{\text{CIFAR10}}}}
    & Normalization & \text{IS}~$\uparrow$ & \text{FID}~$\downarrow$ & \text{Precision}~$\uparrow$ & \text{Recall}~$\uparrow$  \\
    \cmidrule[1.0pt]{2-6}
    & Moving average~\cite{pmlr-v37-ioffe15}  & 9.92 & 4.65 & 0.74 & 0.64 \\ 
    & Batch statistics~\cite{pmlr-v37-ioffe15} & 9.93 & 4.12 & 0.74 & 0.65 \\
    & Standing statistics~\cite{Brock2019LargeSG} & 9.96 & 4.16 & 0.74 & 0.65\\
    \cmidrule[1.0pt]{2-6}
    \parbox[t]{2mm}{\multirow{3}{*}{\rotatebox[origin=c]{90}{\text{ImgNet}}}}
    & Moving average~\cite{pmlr-v37-ioffe15} & 100.69 & 10.23 & 0.74 & 0.58 \\ 
    & Batch statistics~\cite{pmlr-v37-ioffe15} & 87.62 & 10.86 & 0.73 & 0.60 \\
    & Standing statistics~\cite{Brock2019LargeSG} & 98.51 & 8.54 & 0.75 & 0.60 \\
    \cmidrule[1.0pt]{2-6}
    \end{tabular}}
    \vspace{-0.3cm}
    \label{table:normalizing}
\end{table}
We experimentally discover that the normalization operation of batch normalization layers can significantly affect the generation quality in testing time if the generator contains a batch normalization layer. Specifically, the generation results vary depending on the way of calculating statistics (mean and variance of features) for the normalization operation: (1) using updated moving average statistics~\cite{pmlr-v37-ioffe15}, (2) adopting batch statistics~\cite{pmlr-v37-ioffe15}, and (3) standing statistics~\cite{Brock2019LargeSG}. The moving average statistics are computed as the moving average of batch statistics during training time, exactly how the batch normalization layer works in testing time. On the other hand, batch statistics work in the same fashion as batch normalization in training time. Unlike the previous two approaches, standing statistics proposed by Brock~\etal~\cite{Brock2019LargeSG} compute the statistics by accumulating statistics of multiple batches~(\textit{number of repetition}) whose batch size is randomly determined from 1 to \textit{maximum batch size}. As shown in Table~\ref{table:normalizing}, (1) and (2) give poor synthesis outputs compared to (3) based on various metrics, including IS, FID, and Precision $\&$ Recall. In addition, when we evaluate GAN using (2), the generation performance differs according to the batch size, resulting in inconsistent evaluation. We suggest using the \textbf{standing statistics} for GAN evaluation.

\subsection{Reproducibility}
\label{sec:other_issues}
In addition to the previous issues, we observe that open-sourced GANs have some mistakes and missing statements. For instance, when evaluating a generative model during training for monitoring purposes, it is necessary to switch the generative model to the evaluation mode that freezes the moving average moments in the batch normalization layers. This prevents the model from cheating information in the validation dataset.
In addition, many GAN approaches do not mention which dataset split is used for the evaluation. The lack of description induces misleading comparison since the metrics are sensitive to the reference distribution. Therefore, GAN approaches should clarify the dataset type (train, validation, and test) and the number of reference data for a fair evaluation. If a subset of a dataset is used in the assessment, providing the subset is required for reproducibility.

Moreover, training results dramatically change depending on the choice of hyperparameters, hidden details, and even the specific release version of machine learning framework~(Caffe~\cite{jia2014caffe}, TensorFlow~\cite{tensorflow2015-whitepaper}, PyTorch~\cite{NEURIPS2019_9015}, \textit{etc.}). 
As a result, it is hard to reproduce the reported performance of GAN even with the authors' official implementation. For example, the authors' implementation of BigGAN~\cite{brock2018large} is not guaranteed to reproduce the experiment on ImageNet dataset~\cite{biggan_issue, Zhao2020FeatureQI}. Moreover, image generation benchmarks on CIFAR10~\cite{Krizhevsky2009LearningML} and ImageNet~\cite{Deng2009ImageNetAL} are not compared carefully, so the reported performances in BigGAN~\cite{Brock2019LargeSG}, TACGAN~\cite{NIPS2019_8414}, OmniGAN~\cite{zhou2020omni}, ReACGAN~\cite{kang2021rebooting}, and ADCGAN~\cite{hou2021cgans} are not consistent, resulting in an unfair and improper comparison between models.
\section{Protocol Suggestion}
\label{sec:protocol_suggestion}
With the discussions and observations in Sec.~\ref{sec:issues_in_training}, it is obvious to build a standard protocol before making a fair comparison. Therefore, we propose \newtext{the training and evaluation protocols to train and evaluate a variety of GAN models.}

\vspace{2mm}
\noindent \textbf{Training protocol.} When training a GAN model for high-quality image generation, it is important to keep the quality of training images as good as possible:
\begin{enumerate}
\item If an image resizer is required in training, use a high-quality resizer (\textit{e.g.,} PIL.BICUBIC and PIL.LANCZOS) that minimizes aliasing artifacts.
\item If there is a need to save processed images into a disk, keep the images with lossless compression (\textit{e.g.} png) or with little lossy compression (\textit{e.g.} high-quality jpeg).
\item Avoid applying any unnecessary quantization operation to the training dataset. 
\item When evaluating a generative model during training, switch the generative model to freeze the batch normalization layers in the generative model.
\end{enumerate}\leavevmode

\vspace{2mm}
\noindent \textbf{Evaluation protocol.} When evaluating a GAN model, reproducibility and fairness are the crucial factors. Therefore, we present an evaluation protocol as follows:
\begin{enumerate}
\item Save generated images in the lossless png format to a disk using a proper image quantization operation.
\item Resize real and fake images using the backbone-friendly resizer described in section~\ref{sec:pil_bicubic_is_good}.
\item Evaluation of a generative model should include FD~(\textit{e.g.} FID) and metrics that can quantify fidelity~(\textit{e.g.} Precision) and diversity~(\textit{e.g.} Recall) of generated samples~\cite{Kynknniemi2019ImprovedPA, ferjad2020icml}.

\item If the generator contains batch normalization layers, apply the standing statistics technique proposed by~\cite{Brock2019LargeSG}.
\item Clarify the dataset name and split (training, validation, and test) and the number of reference data for a fair evaluation.
\item Specify the version of the deep learning framework and the included library to reproduce the results.
\end{enumerate}

\section{StudioGAN}
\label{sec:studiogan}

This section introduces StudioGAN, a software library that provides reproducible implementations of representative GANs. StudioGAN is motivated by practical needs, where reproducibility has been a ball and chain around the ankle of GAN development. StudioGAN supports over \textbf{30 popular GANs} and \textbf{GAN-related modules} based on the PyTorch framework~(refer to the appendix for the full list). 

In short, StudioGAN has the following key features:
\begin{itemize}
\item \textbf{Coverage:} StudioGAN is a self-contained library that provides 7 GAN architectures, 9 conditioning methods, 4 adversarial losses, 13 regularization modules, 3 augmentation modules, 8 evaluation metrics, and 5 evaluation backbones. Among these configurations, we formulate 30 GANs as representatives.
\item \textbf{Flexibility:} Each modularized option is managed through a configuration system that works through a YAML file, so users can train a large combination of GANs by mix-matching distinct options.
\item \textbf{Reproducibility:} With StudioGAN, users can compare and debug various GANs with the unified computing environment without concerning about hidden details and tricks. Table~\ref{table:reproducibility} shows that StudioGAN successfully reproduces the most representative GANs on four benchmark datasets.

\item \textbf{Versatility:} StudioGAN supports 5 types of acceleration methods with synchronized batch normalization for training: a single GPU training, data-parallel training (DP), distributed data-parallel training~(DDP), multi-node distributed data-parallel training (MDDP), and mixed-precision training~\cite{micikevicius2018mixed}.
\end{itemize}
\emptext{The implemented methods in StudioGAN are marked with colored boxes in Figure~\ref{fig:taxonomy}. We gently note that StudioGAN is not the only attempt to provide various implementations. Repositories such as MMGeneration~\cite{2021mmgeneration} and Mimicry~\cite{lee2020mimicry} provide 16 and 6 GAN implementations, respectively, with IS and FID values. While these platforms are beneficial, they evaluate GAN with limited metrics and have little flexibility to mix-match diverse backbones and recent training techniques.}

\begin{table}[t]
\caption{We check the reproducibility of GANs implemented in StudioGAN by comparing IS~\cite{Salimans2016ImprovedTF} and FID~\cite{Heusel2017GANsTB} with the original papers. We identify our platform successfully reproduces most of representative GANs except for PD-GAN~\cite{Miyato2018cGANsWP}, ACGAN~\cite{Odena2017ConditionalIS}, LOGAN~\cite{Wu2019LOGANLO}, SAGAN~\cite{Zhang2019SelfAttentionGA}, and BigGAN-Deep~\cite{Brock2019LargeSG}. FQ means Flickr-Faces-HQ Dataset (FFHQ)~\cite{karras2019style}. The resolutions of ImageNet~\cite{Deng2009ImageNetAL}, AFHQv2~\cite{choi2020starganv2, karras2021alias}, and FQ datasets are 128, 512, and 1024, respectively. }
    \vspace{-2mm}
    \centering
    \setlength\tabcolsep{2.6pt}
    \resizebox{0.47\textwidth}{!}{
    \begin{tabular}{clcccc}
    \cmidrule[1.0pt]{2-6}
    & \multirow{3}*[-0.5ex]{\normalsize{Method}} & \multicolumn{2}{c}{\multirow{2}{*}{\normalsize{Paper}}} & \multicolumn{2}{c}{\text{Our StudioGAN}} \\
    &  & \multicolumn{2}{c}{} & \multicolumn{2}{c}{\text{Implementation}}\\
    \cmidrule[1.0pt]( r){3-4}
    \cmidrule[1.0pt]( l){5-6}
    & & \text{IS}~$\uparrow$ & \text{FID}~$\downarrow$ & \text{IS}~$\uparrow$ & \text{FID}~$\downarrow$ \\
    \cmidrule[1.0pt]{2-6}
    \parbox[t]{2mm}{\multirow{14}{*}{\rotatebox[origin=c]{90}{CIFAR10}}}
    & WGAN-GP~\cite{Gulrajani2017ImprovedTO, Miyato2018SpectralNF} & 6.68 & 40.1 & 6.71~(\textcolor{teal}{$\uparrow$ 0.03}) & 36.84~(\textcolor{teal}{$\downarrow$ 3.26}) \\
    & PD-GAN~\cite{Miyato2018cGANsWP} & 8.62 & 17.5 & 7.26~(\textcolor{purple}{$\downarrow$ 1.36}) & 29.79~(\textcolor{purple}{$\uparrow$ 12.29})\\
    & SNDCGAN~\cite{Miyato2018SpectralNF} & 7.42 & 29.3 & 8.20~(\textcolor{teal}{$\uparrow$ 0.78}) & 15.66~(\textcolor{teal}{$\downarrow$ 13.64})\\
    & SNResGAN~\cite{Miyato2018SpectralNF} & 8.22 & 21.7 & 8.98~(\textcolor{teal}{$\uparrow$ 0.76}) & 9.51~(\textcolor{teal}{$\downarrow$ 12.19}) \\
    & BigGAN~\cite{Brock2019LargeSG} & 9.22  & 14.73 & 9.97~(\textcolor{teal}{$\uparrow$ 0.75}) & 4.16~(\textcolor{teal}{$\downarrow$ 10.57}) \\
    & StyleGAN2~\cite{karras2020analyzing} & 9.53 & 6.96 & 10.28~(\textcolor{teal}{$\uparrow$ 0.75}) & 3.69~(\textcolor{teal}{$\downarrow$ 3.27}) \\
    & MHGAN~\cite{kavalerov2021multi, Karras2020TrainingGA} & 9.58 &  6.40 & 10.13~(\textcolor{teal}{$\uparrow$ 0.55}) & 3.93~(\textcolor{teal}{$\downarrow$ 2.47})\\
    & StyleGAN2 + ADA~\cite{Karras2020TrainingGA} & 10.14 & 2.42 & 10.46~(\textcolor{teal}{$\uparrow$ 0.32}) &  2.31~(\textcolor{teal}{$\downarrow$ 0.11})  \\
    & CRGAN~\cite{Zhang2019ConsistencyRF} & - &  11.48 & 10.38 & 7.18~(\textcolor{teal}{$\downarrow$ 4.3}) \\
    & ICRGAN~\cite{Zhao2020ImprovedCR} & - & 9.21 & 10.15 & 7.43~(\textcolor{teal}{$\downarrow$ 1.78}) \\\
    & BigGAN-DiffAug~\cite{zhao2020differentiable} & 9.25 & 8.59 & 9.78~(\textcolor{teal}{$\uparrow$ 0.53}) & 7.16~(\textcolor{teal}{$\downarrow$ 1.43}) \\
    & BigGAN-LeCam~\cite{lecam2021} & 9.31 & 8.31 & 10.13~(\textcolor{teal}{$\uparrow$ 0.82}) & 7.36~(\textcolor{teal}{$\downarrow$ 0.95})\\
    & ACGAN~\cite{Odena2017ConditionalIS, zhou2018activation} & 8.25  & - & 7.33~(\textcolor{purple}{$\downarrow$ 0.92}) & 31.72 \\
    & LOGAN~\cite{Wu2019LOGANLO} & 8.67 & 17.7 &   7.66~(\textcolor{purple}{$\downarrow$ 1.01}) & 20.65~(\textcolor{purple}{$\uparrow$ 2.95})\\
    \cmidrule[1.0pt]{2-6}
    \parbox[t]{2mm}{\multirow{5}{*}{\rotatebox[origin=c]{90}{ImageNet}}}
    & SNGAN~\cite{Miyato2018SpectralNF, shim2020circlegan} & 36.80 & 27.62 & 32.40~(\textcolor{purple}{$\downarrow$ 4.4}) & 28.39~(\textcolor{purple}{$\uparrow$ 0.23})\\
    & SAGAN~\cite{Zhang2019SelfAttentionGA} & 51.43 & 18.28 & 19.16~(\textcolor{purple}{$\downarrow$ 32.27}) & 51.82~(\textcolor{purple}{$\uparrow$ 33.54}) \\
    & BigGAN~(B.S.=2048)~\cite{Brock2019LargeSG} & 98.8 & 8.7 & 97.76~(\textcolor{purple}{$\downarrow$ 1.04}) & 8.35~(\textcolor{teal}{$\downarrow$ 0.35}) \\
    & StyleGAN3-t~\cite{sauer2022stylegan} & 15.30 & 53.57 & 20.99~(\textcolor{teal}{$\uparrow$ 5.69}) & 36.21~(\textcolor{teal}{$\downarrow$ 17.36}) \\
    & BigGAN~(B.S.=256)~\cite{Brock2019LargeSG, NIPS2019_8414} & 38.05  & 22.77 & 43.97~(\textcolor{teal}{$\uparrow$ 5.92}) & 16.36~(\textcolor{teal}{$\downarrow$ 6.41}) \\
    \cmidrule[1.0pt]{2-6}
    \parbox[t]{2mm}{\multirow{3}{*}{\rotatebox[origin=c]{90}{\scriptsize{AFHQv2}}}}
    & StyleGAN2 + ADA~\cite{Karras2020TrainingGA} & - & 4.62 & 12.13 & 4.58~(\textcolor{teal}{$\downarrow$ 0.04}) \\
    & StyleGAN3-t + ADA ~\cite{karras2021alias} & - & 4.04 & 11.84 & 4.47~(\textcolor{purple}{$\uparrow$ 0.43})\\
    & StyleGAN3-r + ADA ~\cite{karras2021alias} & - & 4.40 & 11.97 & 4.68~(\textcolor{purple}{$\uparrow$ 0.28})\\
    \cmidrule[1.0pt]{2-6}
    \parbox[t]{2mm}{\multirow{1}{*}{\rotatebox[origin=c]{90}{FQ}}}
    & StyleGAN2 ~\cite{karras2020analyzing} & - & 2.70  & 5.17 & 2.76~(\textcolor{purple}{$\uparrow$ 0.06}) \\
    \cmidrule[1.0pt]{2-6}
    \end{tabular}}
    \label{table:reproducibility}
    \vspace{-4mm}
\end{table}
\section{Benchmark}
\label{sec:benchmark}
\subsection{Reproducibility of StudioGAN}
\label{sec:reproduce}
\newtext{Before presenting a benchmark, we examine whether our software library can reproduce evaluation results of well-known GAN papers. We utilize CIFAR10, ImageNet, AFHQv2, and FFHQ (FQ) datasets, along with Inception Score (IS) and Fr\'echet Inception Distance (FID), for the reproducibility check. StudioGAN generally presents superior image synthesis results in most cases, with the exceptions of PD-GAN~\cite{Miyato2018cGANsWP}, ACGAN~\cite{Odena2017ConditionalIS}, LOGAN~\cite{Wu2019LOGANLO}, and SAGAN~\cite{Zhang2019SelfAttentionGA}. Investigating the performance discrepancies between original implementations and StudioGAN would give researchers insights into successful GAN training; however, pinpointing exact causes is extremely demanding, as implemented libraries, used datasets, and minor details vary from paper to paper. For example, StudioGAN seems to successfully replicate the results of SNGAN~\cite{Miyato2018SpectralNF} on ImageNet but struggles to reproduce the results of SAGAN~\cite{Zhang2019SelfAttentionGA}. The only differences between SNGAN and SAGAN are the self-attention layers~\cite{tfsagan} and changed learning rate as suggested by the SAGAN paper~\cite{Zhang2019SelfAttentionGA}. We notice such small difference results in the reproducibility issue.}

\newtext{BigGAN~\cite{Brock2019LargeSG} and StyleGAN-family models~\cite{karras2019style, karras2020analyzing, Karras2020TrainingGA, karras2021alias, sauer2022stylegan} have become the standard for GAN developments. Since BigGAN and StyleGAN-family models are open sourced by the authors, IS~\cite{Salimans2016ImprovedTF} and FID~\cite{Heusel2017GANsTB} of these models have shown continuous improvement. Regarding StyleGAN2 on CIFAR10 and ImageNet, we discover that the performance discrepancies between StudioGAN and the original implementation are due to the difference in data pre-processing methods. Despite our efforts to match model parameters, hyperparameter settings, and data processing setup, some results still differ from original implementations. We attribute this to slight differences in low-level implementation details. From our experience, even minor changes, such as loading sequence of tensors to CUDA, can lead to considerable variations in training results. For BigGAN, FID scores on CIFAR10 are different across papers (14.73 in~\cite{Brock2019LargeSG}, 7.27 in~\cite{zhou2020omni}, and 4.16 in our paper). We speculate that the superior performance of BigGAN from StudioGAN can be attributed to the different hyperparameter settings and the fine-details mentioned above as well as the employment of standing statistics~\cite{Brock2019LargeSG}, and synchronized batch normalization~\cite{mao2018bn}.}

\subsection{Baselines}
\label{sec:baselines}
We provide a comprehensive GAN benchmark using the StudioGAN library. We train and evaluate GANs spanning primitive DCGAN~\cite{Radford2016UnsupervisedRL} to recent StyleGAN3~\cite{karras2021alias}, built upon four different types of backbone for the generator and the discriminator (DCGAN, ResNetGAN, BigGAN, and StyleGAN). Some GANs~(such as CRGAN~\cite{Zhang2019ConsistencyRF} and DiffAug~\cite{zhao2020differentiable}) can be applied to multiple GAN backbones, but we use the same generator and discriminator architectures described in the original paper to provide evaluation results close to the original study.

BigGAN and StyleGANs are representative GANs for natural image synthesis. While those models are being actively studied, BigGAN and StyleGANs tend to be studied separately in the literature because StyleGANs are designed for unconditional image generation, whereas BigGAN is designed for conditional image generation. To compare those models in a fair way, we adopt the class conditioning version of StyleGAN2~\cite{Karras2020TrainingGA}, StyleGAN3-t~\cite{karras2021alias}, and StyleGAN3-r~\cite{karras2021alias} for all experiments except for the cases on AFHQv2 and FFHQ datasets in Tables~\ref{table:inception}, A3, and A4. Because of the poor convergence issue, StyleGAN3-r results on ImageNet are omitted from Tables~\ref{table:inception}, A3, and A4. Here, we utilize InceptionV3, SwAV, and Swin-T for the evaluation backbone to avoid biased conclusions.

\subsection{Datasets}
We use four standard datasets for the conditional image generation: \textbf{CIFAR10}~\cite{Krizhevsky2009LearningML}, \textbf{ImageNet}~\cite{Deng2009ImageNetAL}, \textbf{AFHQv2}~\cite{choi2020starganv2, karras2021alias}, and \textbf{FFHQ}~\cite{karras2019style} and three of our proposed ImageNet subsets: \textbf{Baby/Papa/Grandpa-ImageNet}. 
We create the subsets of ImageNet for a small-scale ImageNet experiment because training GAN on the full ImageNet requires extensive computational resources. The proposed subsets of the ImageNet are made according to the classification difficulty, enabling researchers to analyze behaviors of conditional GANs based on the level of classification difficulty. Baby-ImageNet contains the easiest images to classify, whereas the Granpa-ImageNet contains the hardest ones. We attach a detailed explanation of datasets in the appendix. Tiny-ImageNet~\cite{Tiny} is also intended for the small-scale ImageNet experiment, but we decided not to use it for the benchmark because Tiny-ImageNet is known to have severe aliasing artifacts and degradation. 

\subsection{Evaluation metrics}
\label{sec:datasets_and_evaluation_metrics}
We utilize 7 evaluation metrics: Inception Score (IS)~\cite{Salimans2016ImprovedTF} (Eq.~\ref{eq:IS}), Fr\'echet Inception Distance (FID)~\cite{Heusel2017GANsTB} (Eq.~\ref{eq:FID}), Precision \& Recall~\cite{Kynknniemi2019ImprovedPA}, Density \& Coverage~\cite{ferjad2020icml}, and Intra-class FID~(IFID)~\cite{Zhang2019SelfAttentionGA} that is the average of class-wise FID. Each metric is computed using TensorFlow-InceptionV3 network in which weights are borrowed from Improved-GAN GitHub~\cite{openaiinception}. We use PyTorch implementations of IS~(we implemented), FID~\cite{pytorchttur}, Precision \& Recall~\cite{officialprdc}, and Density \& Coverage~\cite{officialprdc} to construct seamless training and evaluation procedures.

\textbf{Precision} $\&$ \textbf{Recall} are used to quantify the fidelity and diversity of generated images based on the estimated support of real and fake distributions. Precision is defined as the portion of randomly generated images that falls within the support of real data distribution. On the other hand, Recall is the portion of real images falling within the support of fake data distribution. High Precision and Recall mean that the generated images are high-quality and diverse. Mathematically, Precision $\&$ Recall can be written as follows:
\begin{align*}
    & \text{Precision}(\mX_{s}, \mX_{t}, F, k) := 
    \frac{1}{M}\sum_{i=1}^{M}\mathbb{I}\bigg(\vf_{t, i}\!\in\! \mathcal{M}\Big(F(\mX_{s}), k\Big)\!\bigg),
    \label{eq:precision}\tag{5}\\
    & \text{Recall}(\mX_{s}, \mX_{t}, F, k)\!:=\! \frac{1}{N}\sum_{i=1}^{N}\mathbb{I}\bigg(\!\vf_{s, i}\!\in\! \mathcal{M}\Big(F(\mX_{t}), k\Big)\!\bigg),
    \label{eq:recall}\tag{6}
\end{align*}
where $\mathcal{M}(\mF, k) := \bigcup_{\vf \in \mF}B\big(\vf, \text{NN}_{k}(\mF, \vf, k)\big)$, $\mF$ is a collection of features from $F$, $\mX_{s} = \{\vx_{s, 1},..., \vx_{s, N}\}$ is a source dataset~(real images), $\mX_{t} = \{\vx_{t, 1},..., \vx_{t, M}\}$ is a target dataset~(generated images), $B(\vf, r)$ is the n-dimensional sphere in which $\vf = F(\vx) \in \mathbb{R}^{n}$ is the center and $r$ is the radius, $\text{NN}_{k}(\mF, \vf, k)$ is the distance from $\vf$ to the $k$-th nearest embedding in $\mF$, and $\mathbb{I}(\cdot)$ is a indicator function. 

Recently, Naeem~\etal~\cite{ferjad2020icml} show that the manifold construction process using the nearest neighbor function is vulnerable to outlier samples, so it often results in an overestimated manifold of the distribution. With the overestimation problem fixed by sample counting, Naeem~\etal propose \textbf{Density} $\&$ \textbf{Coverage} metrics. Mathematically, these metrics can be expressed as follows:
\begin{align*}
    & \text{Density}(\mX_{s}, \mX_{t}, F, k) := \\ 
    & \quad\frac{1}{kM}\sum_{j=1}^{M}\sum_{i=1}^{N}\mathbb{I}\bigg(\vf_{t, j} \!\in\! B\Big(\vf_{s, i}, \text{NN}_{k}(F(\mX_{s}), \vf_{s, i}, k)\Big)\!\bigg),
    \label{eq:eq8}\tag{8}\\
    & \text{Coverage}(\mX_{s}, \mX_{t}, F, k) := \\     
    & \frac{1}{N}\sum_{i=1}^{N} \mathbb{I}\bigg(\exists j \text{ s.t. } \vf_{t, j} \in B\Big(\vf_{s, i}, \text{NN}_{k}(F(\mX_{s}), \vf_{s, i}, k)\Big)\bigg).
    \label{eq:eq9}\tag{9}
\end{align*}

We perform additional evaluations by replacing the InceptionV3 with a pre-trained SwAV~\cite{officialswav} and Swin-T~\cite{officialswin} as an alternative evaluation backbone. Table~\ref{table:name_of_metrics} summarizes the names of all metrics w.r.t various backbone networks.
Note that such a comprehensive re-evaluation of representative GANs with these metrics is not presented so far.

\newtext{In addition to the 7 metrics listed above, various other metrics are also considered to evaluate the performance of image synthesis models. Notable metrics include MSE, PSNR, SSIM~\cite{wang2004image}, LPIPS~\cite{zhang2018unreasonable}, Realism score~\cite{Kynknniemi2019ImprovedPA}, and GIQA~\cite{gu2020giqa}. MSE, PSNR, SSIM, and LPIPS measure the reconstruction difference between ground truth and targeted images. Realism score and GIQA measure how close a given image is located from the true data distribution (i.e., how realistically it has been generated). Consequently, these metrics are less suitable for establishing a benchmark for class-conditional image synthesis, which is the primary objective of this paper. Although CAS~\cite{Ravuri2019ClassificationAS} can measure class-conditional precision and recall, it requires time-consuming ResNet training. Instead, we evaluate models using improved Precision \& Recall and Density \& Coverage.}

\begin{table}[t]
\setlength{\tabcolsep}{3pt}
    \caption{Metric names w.r.t. various backbone networks.}
    \vspace{-2mm}
    \centering
    \resizebox{0.5\textwidth}{!}{
    \begin{tabular}{lcccccc}
    \cmidrule[1.0pt]{1-7}
    \multirow{2}*[-0.5ex] {\normalsize{Backbone}} & \multirow{2}*[-0.5ex]{\normalsize{Score}} & Fr\'echet & \multirow{2}*[-0.5ex]{\normalsize{Precision}} &
    \multirow{2}*[-0.5ex]{\normalsize{Recall}} & \multirow{2}*[-0.5ex]{\normalsize{Density}} &
    \multirow{2}*[-0.5ex]{\normalsize{Coverage}} \\
     & & Distance & & \vspace{1mm}\\
    \cmidrule[1.0pt]{1-7}
    \normalsize{InceptionV3}~\cite{Szegedy2016RethinkingTI} & \normalsize{IS} & \normalsize{FID}              & \normalsize{Precision} & \normalsize{Recall} & \normalsize{Density} & \normalsize{Coverage} \vspace{1mm}\\
    \normalsize{SwAV}~\cite{Caron2020UnsupervisedLO}           & \normalsize{SS} & \normalsize{FSD}              & \normalsize{S-Precision} & \normalsize{S-Recall} & \normalsize{S-Density} & \normalsize{S-Coverage} \vspace{1mm}\\
    \normalsize{Swin-T}~\cite{caron2021emerging}         & \normalsize{TS} & \normalsize{FTD}              & \normalsize{T-Precision} & \normalsize{T-Recall} & \normalsize{T-Density} & \normalsize{T-Coverage} \\
    \cmidrule[1.0pt]{1-7}
    \end{tabular}
    }
    \label{table:name_of_metrics}
    \vspace{-2mm}
\end{table}

\subsection{Findings from the evaluation results}
\label{sec:large_scale_results}
Tables~\ref{table:inception},~A3, and A4 show quantitative results using InceptionV3, SwAV, and Swin-T, respectively. Thin solid lines are used to group GANs with the similar training backbone~(DCGAN, ResNet, BigGAN, and StyleGAN). We also visualize the quantitative results in~Figures~A3,~A4,~A5,~A6,~A7,~A8,~A9,~A10, and~A11 to analyze results better. For the details about the experiment setup, please refer to the appendix. We provide our key findings from the experiments below.

\begin{table*}[!htp]
    \caption{Benchmark table evaluated using \textbf{TensorFlow-InceptionV3}~\cite{Szegedy2016RethinkingTI}. The resolutions of CIFAR10~\cite{Krizhevsky2009LearningML}, Baby/Papa/Grandpa ImageNet, ImageNet~\cite{Deng2009ImageNetAL}, AFHQv2~\cite{choi2020starganv2, karras2021alias}, and FQ~\cite{karras2019style} datasets are 32, 64, 128, 512, and 1024, respectively. Top-1 and Top-2 performances are indicated in red and blue, respectively.}
    \vspace{-0.2cm}
    \centering
    \resizebox{0.88\textwidth}{!}{
    \begin{tabular}{clrrrrrrr}
    \cmidrule[1.0pt]{2-9}
    & \textbf{TensorFlow-InceptionV3}~\cite{Szegedy2016RethinkingTI} & \text{IS}~$\uparrow$ & \text{FID}~$\downarrow$ & \text{Precision}~$\uparrow$ & \text{Recall}~$\uparrow$ & \text{Density}~$\uparrow$ & \text{Coverage}~$\uparrow$ & \text{IFID}~$\downarrow$ \\
    \cmidrule[1.0pt]{2-9}
    \parbox[t]{2mm}{\multirow{29}{*}{\rotatebox[origin=c]{90}{\text{CIFAR10}}}}
    & DCGAN~\cite{Goodfellow2014GenerativeAN} & 6.11 & 48.29 & 0.61 & 0.21 & 0.52 & 0.28 & 117.00\\ 
    & LSGAN~\cite{Mao2017LeastSG} & 6.55 & 40.22 & 0.60 & 0.34 & 0.49 & 0.35 & 111.51 \\ 
    & Geometric GAN~\cite{Lim2017GeometricG} & 6.61 & 43.16 & 0.58 & 0.25 & 0.48 & 0.31 & 114.59 \\ 
    \cmidrule[0.25pt]{2-9}
    & ACGAN-Mod~\cite{Odena2017ConditionalIS} & 7.19 & 33.39 & 0.61 & 0.21 & 0.52 & 0.36 & 72.75 \\ 
    & WGAN~\cite{Arjovsky2017TowardsPM} & 3.59 & 107.68 & 0.45 & 0.02 & 0.29 & 0.08 & 158.11 \\ 
    & DRAGAN~\cite{Kodali2018OnCA} & 6.59 & 34.69 & 0.63 & 0.47 & 0.56 & 0.36 & 106.46 \\ 
    & WGAN-GP~\cite{Gulrajani2017ImprovedTO} & 5.22 & 53.98 & 0.68 & 0.26 & 0.67 & 0.27 & 119.22 \\ 
    & PD-GAN~\cite{Miyato2018cGANsWP} & 7.25 & 31.54 & 0.61 & 0.28 & 0.53 & 0.39 & 65.04 \\ 
    & SNGAN~\cite{Miyato2018SpectralNF} & 8.83 & 9.01 & 0.70 & 0.62 & 0.79 & 0.75 & 22.85 \\ 
    & SAGAN~\cite{Zhang2019SelfAttentionGA} & 8.64 & 10.35 & 0.69 & 0.63 & 0.73 & 0.71 & 25.00 \\ 
    & TACGAN~\cite{NIPS2019_8414} & 7.68 & 29.51 & 0.62 & 0.28 & 0.57 & 0.41 & 64.52 \\ 
    & LOGAN~\cite{Wu2019LOGANLO} & 7.89 & 20.45 & 0.65 & 0.63 & 0.61 & 0.54 & 95.82 \\ 
    \cmidrule[0.25pt]{2-9}
    & BigGAN~\cite{Brock2019LargeSG} & 9.96 & 4.16 & 0.74 & 0.65 & 0.98 & 0.88 & 14.29 \\ 
    & CRGAN~\cite{Zhang2019ConsistencyRF} & 10.09 & \placeholder{\topthree{3.16}} & 0.74 & 0.66 & 0.97 & \placeholder{\topthree{0.90}} & 12.90 \\ 
    & MHGAN~\cite{kavalerov2021multi} & 10.07 & 3.95 & 0.73 & 0.65 & 0.92 & 0.88 & 14.42 \\ 
    & ICRGAN~\cite{Zhao2020ImprovedCR} & 10.14 & 3.54 & \textbf{\topone{0.75}} & 0.64 & 0.98 & \placeholder{\topthree{0.90}} & 13.37 \\ 
    & ContraGAN~\cite{kang2020contragan} & 9.78 & 6.01 & 0.74 & 0.63 & 0.95 & 0.84 & 119.63 \\ 
    & BigGAN + DiffAug~\cite{zhao2020differentiable} & 10.03 & 3.35 & 0.74 & 0.67 & 0.97 & \placeholder{\topthree{0.90}} & \placeholder{\topthree{12.63}} \\ 
    & BigGAN + LeCam~\cite{lecam2021} & 10.17 & 3.23 & 0.73 & 0.67 & 0.95 & 0.89 & 13.11\\ 
    & ADCGAN~\cite{hou2021cgans} & 9.91 & 3.93 & 0.72 & 0.67 & 0.92 & 0.88 & 14.27 \\ 
    & ReACGAN~\cite{kang2021rebooting} & 9.85 & 3.87 & \textbf{\topone{0.75}} & 0.62 & 1.01 & 0.88 & 15.17 \\ 
    \cmidrule[0.25pt]{2-9}
    & StyleGAN2~\cite{karras2020analyzing} & 10.17 & 3.78 & 0.72 & 0.67 & 0.97 & 0.89 & 14.04 \\ 
    & StyleGAN2 + DiffAug~\cite{karras2020analyzing, zhao2020differentiable} & \textbf{\topone{10.55}} & \textbf{\toptwo{2.39}} & 0.74 & \textbf{\toptwo{0.68}} & \textbf{\toptwo{1.02}} & \textbf{\toptwo{0.92}} & \textbf{\topone{11.56}} \\ 
    & StyleGAN2 + ADA~\cite{Karras2020TrainingGA} & \textbf{\toptwo{10.53}} & \textbf{\topone{2.31}} & \textbf{\topone{0.75}} & \textbf{\topone{0.69}} & \textbf{\topone{1.04}} & \textbf{\topone{0.93}} & \textbf{\toptwo{11.82}} \\ 
    & StyleGAN2 + LeCam~\cite{karras2020analyzing, lecam2021} & 10.04 & 4.65 & 0.71 & 0.66 & 0.90 & 0.86 & 15.66 \\ 
    & StyleGAN2 + APA~\cite{deceived2021} & 10.05 & 4.04 & 0.70 & \textbf{\toptwo{0.68}} & 0.89 & 0.87 & 54.40 \\ 
    & StyleGAN2 + D2D-CE~\cite{kang2021rebooting} & \placeholder{\topthree{10.43}} & 3.44 & 0.74 & 0.64 & \textbf{\toptwo{1.02}} & \placeholder{\topthree{0.90}} & 14.14 \\ 
    & StyleGAN3-r + ADA~\cite{karras2021alias}  & 9.72 & 10.83 & 0.68 & 0.53 & 0.84 & 0.75 & 27.32 \\ 
    \cmidrule[1.0pt]{2-9}
    
    \parbox[t]{2mm}{\multirow{8}{*}{\rotatebox[origin=c]{90}{\text{Baby-ImageNet}}}}
    & SNGAN~\cite{Miyato2018SpectralNF} & 18.21 & 36.31 & 0.54 & \textbf{\toptwo{0.58}} & 0.38 & 0.39 & 93.10 \\ 
    & SAGAN~\cite{Zhang2019SelfAttentionGA} & 16.47 & 42.10 & 0.50 & \placeholder{\topthree{0.53}} & 0.33 & 0.32 & 111.11 \\
    \cmidrule[0.25pt]{2-9}
    & BigGAN~\cite{Brock2019LargeSG} & \textbf{\topone{27.47}} & \textbf{\topone{18.64}} & \textbf{\topone{0.63}} & 0.47 & \textbf{\topone{0.59}} & \textbf{\toptwo{0.57}} & \textbf{\topone{62.22}} \\ 
    & ContraGAN~\cite{kang2020contragan} & 23.14 & 24.67 & \placeholder{\topthree{0.62}} & 0.47 & 0.53 & 0.45 & 220.72 \\ 
    & ReACGAN~\cite{kang2021rebooting} & \textbf{\toptwo{26.18}} & \textbf{\toptwo{19.75}} & \textbf{\topone{0.63}} & 0.43 & \placeholder{\topthree{0.54}} & \placeholder{\topthree{0.50}} & \placeholder{\topthree{77.95}} \\ 
    \cmidrule[0.25pt]{2-9}
    & StyleGAN2~\cite{karras2020analyzing} & \placeholder{\topthree{23.40}} & \placeholder{\topthree{22.21}} & \placeholder{\topthree{0.62}} & \textbf{\topone{0.59}} & \textbf{\toptwo{0.58}} & \textbf{\topone{0.62}} & \textbf{\toptwo{70.83}} \\ 
    & StyleGAN3-t~\cite{karras2021alias} & 9.99 & 102.24 & 0.37 & 0.01 & 0.17 & 0.06 & 235.20 \\ 
    \cmidrule[1.0pt]{2-9}
    
    \parbox[t]{2mm}{\multirow{8}{*}{\rotatebox[origin=c]{90}{\text{Papa-ImageNet}}}}
    & SNGAN~\cite{Miyato2018SpectralNF} & 15.04 & 41.06 & 0.52 & \textbf{\toptwo{0.55}} & 0.38 & 0.37 & 107.13 \\ 
    & SAGAN~\cite{Zhang2019SelfAttentionGA} & 12.86 & 51.55 & 0.48 & \placeholder{\topthree{0.53}} & 0.31 & 0.27 & 130.10 \\ 
    \cmidrule[0.25pt]{2-9}
    & BigGAN~\cite{Brock2019LargeSG} & \textbf{\toptwo{21.62}} & \placeholder{\topthree{24.99}} & \placeholder{\topthree{0.60}} & 0.47 & \textbf{\toptwo{0.54}} & \textbf{\toptwo{0.53}} & \textbf{\toptwo{79.35}} \\ 
    & ContraGAN~\cite{kang2020contragan} & 19.22 & 25.33 & \textbf{\topone{0.62}} & 0.48 & \textbf{\toptwo{0.54}} & 0.46 & 203.33 \\ 
    & ReACGAN~\cite{kang2021rebooting} & \textbf{\topone{21.80}} & \textbf{\topone{21.74}} & \textbf{\topone{0.62}} & 0.47 & \textbf{\topone{0.55}} & \placeholder{\topthree{0.51}} & \placeholder{\topthree{91.76}} \\ 
    \cmidrule[0.25pt]{2-9}
    & StyleGAN2~\cite{karras2020analyzing} & \placeholder{\topthree{19.28}} & \textbf{\toptwo{23.28}} & 0.58 & \textbf{\topone{0.61}} & 0.52 & \textbf{\topone{0.58}} & \textbf{\topone{76.06}} \\ 
    & StyleGAN3-t~\cite{karras2021alias} & 9.45 & 88.74 & 0.47 & 0.00 & 0.34 & 0.11 & 234.71 \\ 
    \cmidrule[1.0pt]{2-9}

    \parbox[t]{2mm}{\multirow{8}{*}{\rotatebox[origin=c]{90}{\text{Grandpa-ImageNet}}}}
    & SNGAN~\cite{Miyato2018SpectralNF} & 12.70 & 41.74 & 0.49 & \textbf{\toptwo{0.52}} & 0.35 & 0.37 & 110.47\\ 
    & SAGAN~\cite{Zhang2019SelfAttentionGA} & 11.96 & 50.02 & 0.44 & 0.47 & 0.29 & 0.29 & 131.93 \\ 
    \cmidrule[0.25pt]{2-9}
    & BigGAN~\cite{Brock2019LargeSG} & \placeholder{\topthree{17.64}} & 24.40 & \placeholder{\topthree{0.58}} & \placeholder{\topthree{0.48}} & \placeholder{\topthree{0.55}} & \placeholder{\topthree{0.57}} & \textbf{\toptwo{78.11}} \\ 
    & ContraGAN~\cite{kang2020contragan} & \textbf{\toptwo{17.86}} & \textbf{\toptwo{21.14}} & \textbf{\topone{0.64}} & 0.43 & \textbf{\topone{0.64}} & 0.54 & 193.93 \\ 
    & ReACGAN~\cite{kang2021rebooting} & \textbf{\topone{19.53}} & \textbf{\topone{18.65}} & \textbf{\topone{0.64}} & 0.41 & \textbf{\topone{0.64}} & \textbf{\toptwo{0.58}} & \placeholder{\topthree{92.39}} \\ 
    \cmidrule[0.25pt]{2-9}
    & StyleGAN2~\cite{karras2020analyzing} & 15.71 & \placeholder{\topthree{21.88}} & 0.56 & \textbf{\topone{0.58}} & 0.51 & \textbf{\topone{0.59}} & \textbf{\topone{77.66}} \\ 
    & StyleGAN3-t~\cite{karras2021alias} & 7.79 & 93.60 & 0.49 & 0.00 & 0.36 & 0.10 & 243.25 \\ 
    \cmidrule[1.0pt]{2-9}

    \parbox[t]{2mm}{\multirow{7}{*}{\rotatebox[origin=c]{90}{\text{ImageNet}}}}
    & SNGAN-256~\cite{Miyato2018SpectralNF} & \placeholder{\topthree{31.42}} & 31.49 & 0.56 & \textbf{\topone{0.67}} & 0.43 & \placeholder{\topthree{0.39}} & \textbf{\topone{112.68}} \\ 
    \cmidrule[0.25pt]{2-9}
    & BigGAN-256~\cite{Brock2019LargeSG} & \textbf{\toptwo{40.78}} & \textbf{\toptwo{20.57}} & \placeholder{\topthree{0.66}} & \textbf{\toptwo{0.65}} & \textbf{\toptwo{0.63}} & \textbf{\toptwo{0.52}} & \textbf{\toptwo{121.32}} \\ 
    & ContraGAN-256~\cite{kang2020contragan} & 25.23 & \placeholder{\topthree{29.59}} & \textbf{\toptwo{0.69}} & 0.52 & \placeholder{\topthree{0.62}} & 0.31 & 162.80 \\ 
    & ReACGAN-256~\cite{kang2021rebooting} & \textbf{\topone{66.67}} & \textbf{\topone{15.65}} & \textbf{\topone{0.77}} & 0.38 & \textbf{\topone{0.88}} & \textbf{\topone{0.53}} & \placeholder{\topthree{127.31}} \\ 
    \cmidrule[0.25pt]{2-9}
    & StyleGAN2~\cite{karras2020analyzing} & 22.54 & 33.40 & 0.54 & \placeholder{\topthree{0.63}} & 0.40 & 0.34 & 135.19 \\ 
    & StyleGAN3-t~\cite{karras2021alias} & 21.06 & 36.51 & 0.53 & 0.60 & 0.38 & 0.30 & 142.49 \\ 
    \cmidrule[1.0pt]{2-9}

    \parbox[t]{2mm}{\multirow{4}{*}{\rotatebox[origin=c]{90}{\text{AFHQv2}}}}
    & StyleGAN2~\cite{karras2020analyzing} & \textbf{\topone{11.94}} & 6.08 & \textbf{\topone{0.85}} & 0.49 & \textbf{\topone{1.32}} & 0.81 & 6.02 \\ 
    & StyleGAN2 + ADA~\cite{Karras2020TrainingGA} & \placeholder{\topthree{11.70}} & \placeholder{\topthree{4.75}} & \textbf{\toptwo{0.83}} & \placeholder{\topthree{0.58}} & \textbf{\toptwo{1.20}} & \placeholder{\topthree{0.83}} & \placeholder{\topthree{4.73}} \\
    & StyleGAN3-t + ADA~\cite{karras2021alias}& 11.63 & \textbf{\topone{4.53}} & \placeholder{\topthree{0.81}} & \textbf{\topone{0.68}} & \placeholder{\topthree{1.05}} & \textbf{\topone{0.84}} & \textbf{\toptwo{4.59}} \\ 
    & StyleGAN3-r + ADA~\cite{karras2021alias} & \textbf{\toptwo{11.74}} & \textbf{\toptwo{4.62}} & \placeholder{\topthree{0.81}} & \textbf{\toptwo{0.66}} & 1.04 & \textbf{\topone{0.84}} & \textbf{\topone{4.55}} \\ 
    \cmidrule[1.0pt]{2-9}
    \parbox[t]{2mm}{\multirow{1}{*}{\rotatebox[origin=c]{90}{\text{FQ}}}}
    & StyleGAN2~\cite{karras2020analyzing} & 5.27 & 3.16 & 0.77 & 0.60 & 1.09 & 0.90 & - \\
    \cmidrule[1.0pt]{2-9}
    \end{tabular}}
    \vspace{-0.5cm}
    \label{table:inception}
\end{table*}
\begin{figure*}[ht!]
    \quad\quad\:\normalsize{StyleGAN2 on \textsc{FFHQ~($1024^{2}$)}} 
    \qquad\qquad\qquad\qquad\qquad\quad\
    \normalsize{StyleGAN3-t + ADA on \textsc{AFHQv2~($512^{2}$)}}\par\medskip
    \includegraphics[width=0.99\linewidth]{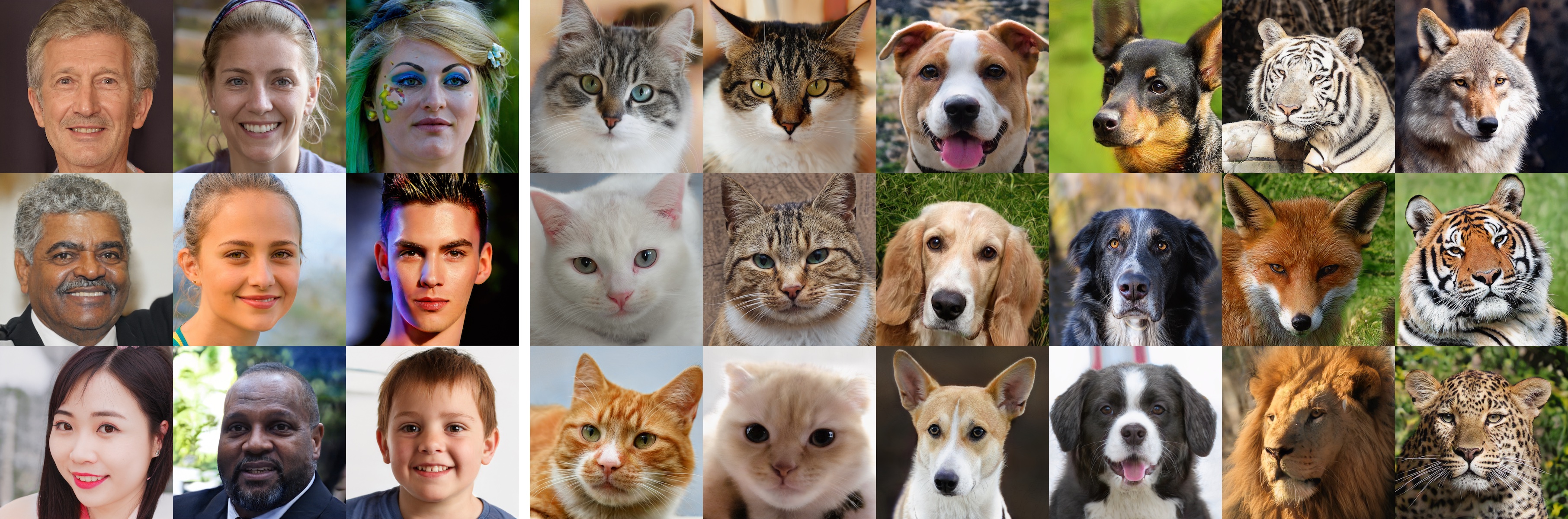}
\end{figure*}
\begin{figure*}[ht!]
    \qquad\qquad\qquad\qquad\qquad\qquad\qquad\quad\normalsize{ReACGAN on \textsc{ImageNet~($128^{2}$)}} 
    \qquad\qquad\qquad\quad\normalsize{StyleGAN2 + ADA on \textsc{CIFAR10~($32^{2}$)}}\par\medskip
    \includegraphics[width=0.99\linewidth]{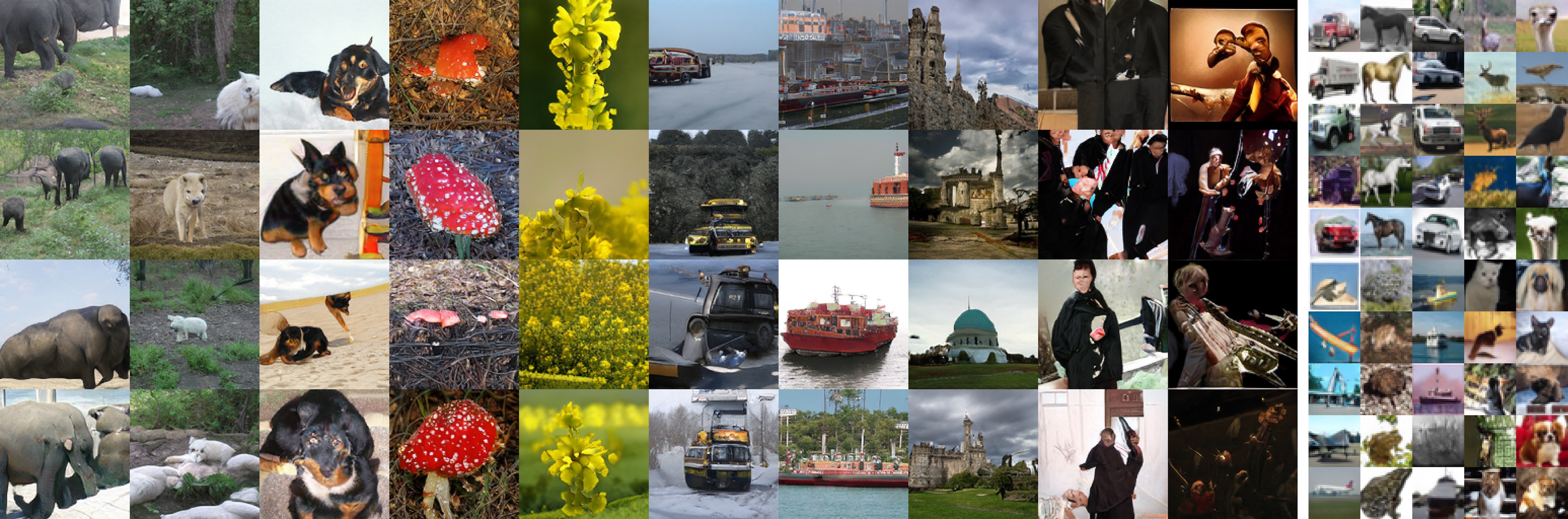}
    \vspace{-0.1cm}
    \caption{Generated images from GANs implemented in StudioGAN library. We select the best performing GAN on each dataset based on the average rank and generate fake images using those GANs. StyleGAN2~\cite{karras2020analyzing} is used for FFHQ~\cite{karras2019style}, StyleGAN3-t + ADA~\cite{karras2021alias} is used for AFHQv2~\cite{choi2020starganv2, karras2021alias}, ReACGAN~\cite{kang2021rebooting} is used for ImaegNet, and StyleGAN2 + ADA~\cite{Karras2020TrainingGA} is used for CIFAR10~\cite{Krizhevsky2009LearningML}.}
    \label{fig:generated_images}
    \vspace{-0.3cm}
\end{figure*}

\vspace{2mm} \noindent \textbf{Rooms for improvement in CIFAR10.} CIFAR10 experiment is a starting point for generative model development and has been highly explored by numerous GANs. Interestingly, we find that the generation performances of representative GANs can still be improved with our StudioGAN implementations. Although FID has several drawbacks mentioned in Sec.~\ref{sec:metrics}, it has long been conventionally adopted as a primary metric for evaluating generative models. Thus we decide to compare FID in this section, as a lot of existing papers only present FID scores. As can be seen in Tables~\ref{table:reproducibility} and \ref{table:inception}, FID values of SNGAN, BigGAN, StyleGAN2, MHGAN, CRGAN, ICRGAN, and BigGAN-DiffAug are significantly lower than the originally reported numbers. We believe StudioGAN achieves better training dynamics with numerically stable spectral normalization implemented in PyTorch and dedicated hyperparameter selection.

\vspace{2mm} \noindent \textbf{StyleGAN2 can generate more diverse images compared to BigGAN.} 
Experiments using Baby/Papa/Grandpa-ImageNet shown in Table~\ref{table:inception} and~A3 present that StyleGAN2 models are apt to exhibit higher recall and coverage values than BigGAN counterparts. Figures~A9 and~A10 in Appendix verify the statement again by showing the orange markers~(StyleGAN2) are located above~(high recall and coverage) the red markers~(BigGAN). However, based on a few examples, we notice that the recall and coverage values of StyleGAN2 are sometimes lower than those of BigGAN. For example, StyleGAN2 trained using ImageNet shows lower density (BigGAN: 0.65 $\rightarrow$ StyleGAN2: 0.63) and coverage~(BigGAN: 0.52 $\rightarrow$ StyleGAN2: 0.34) than BigGAN as can be seen in Table~\ref{table:inception}. We speculate that this is because StyleGAN's generator design is too restrictive to model complex data distributions like ImageNet.

\vspace{2mm} \noindent \textbf{StyleGAN family struggles with ImageNet.} As shown in Tables~\ref{table:inception} and A3, StyleGAN2 and StyleGAN3-t present FID values of 33.4 and 36.51 and FSD values of 5.61 and 6.07 on ImageNet generation experiments, respectively, which are worse than those of BigGAN, ContraGAN, and ReACGAN. In contrast, StyleGAN2 has a better synthesis ability than BigGAN~(refer to Table~A1) with AFHQv2 dataset. Such results imply that StyleGAN2 and StyleGAN3-t are tailored to center-aligned and high-resolution images, and the StyleGAN-family models are not good at generating images having high inter-class variations. \brown{This observation becomes more evident when comparing the FID values of cGAN models trained on subsets of ImageNet (Baby/Papa/Grandpa-ImageNet) and the original ImageNet. StyleGAN2 models trained on Baby-, Papa-, and Grandpa-ImageNet achieve FID scores of 22.21, 23.28, and 21.88, respectively, which are comparable to or even better than those achieved by BigGAN. However, when StyleGAN2 is trained on the full ImageNet, it exhibits a significantly higher FID score (33.40) compared to other cGAN models (BigGAN: 20.57, ContraGAN: 29.59, and ReACGAN: 15.65). This indicates that training conditional StyleGAN2 becomes increasingly challenging as the number of classes in the training data increases.}

\vspace{2mm} \noindent \textbf{Fr\'echet SwAV Distance favors StyleGAN2 over BigGAN.}
As can be seen in Tables~\ref{table:inception} and~A3, we discover that FID and FSD show different preferences across BigGAN, ContraGAN, ReACGAN, and StyleGAN2. Specifically, SwAV tends to take the side of StyleGAN2 while InceptionV3 is apt to take the side of BigGAN and ReACGAN. We believe that StyleGAN variants can generate more realistic images because Morozov~\etal~\cite{Morozov2021OnSI} demonstrates that FSD and Precision \& Recall measured through SwAV provide more consistent evaluation results with human evaluation. However, given that FID~\cite{Heusel2017GANsTB} is still broadly adopted, we think that further analysis of FID, FSD, Precision $\&$ Recall, and S-Precision $\&$ S-Recall is necessary.

\vspace{2mm} \noindent \textbf{Image generation quality vs. image classification accuracy.} As shown in Table~\ref{table:inception}, evaluation results using the InceptionV3 network on Baby/Papa/Grandpa-ImageNet imply no strong correlation between classification difficulty and generation quality. However, Table~A3 shows interesting trends that the FSD value of the generated images gets lower if training images are easily classifiable. For instance, FSD values of BigGAN, ContraGAN, ReACGAN, and StyleGAN on Baby-ImageNet are 4.49, 4.47, 4.18, and 3.43, while FSD values of those models on Papa-ImageNet are 5.3, 4.62, 4.3, and 3.7. This relationship holds for Grandpa-ImageNet too. 
Choosing the most suitable evaluation backbone is an open question, so we do not judge that well-classifiable images are well-generated images. However, we value the correlation between FSD and classification difficulty and believe it is worth exploring in the future.

\subsection{Potential hazards}
So far, we have described an extensive benchmark based on our newly suggested evaluation protocol. However, as we look over the results and inspect the details of each metric, we found some unclear points and the potential danger of misleading metrics. In this section, we elaborate few potential hazards for GAN community to be aware of in future research.

\vspace{2mm}
\noindent \textbf{Evaluation results are highly affected by evaluation backbone.}
As shown in Tables~\ref{table:inception}, A3, and A4, the rank of each metric varies as InceptionV3~\cite{Szegedy2016RethinkingTI}, SwAV~\cite{Caron2020UnsupervisedLO}, or Swin-T~\cite{liu2021swin} is used as an evaluation backbone. InceptionV3 network was adopted as an evaluation backbone because there was a belief that a cutting-edge recognition model is useful for extracting good image features. Still, there is no guarantee that InceptionV3 is the best choice for generative model evaluation. 
Therefore, we encounter a potential hazard that the evaluation using Inception Score (IS), Fr\'echet Inception Distance~(FID), Precision $\&$ Recall, and Density $\&$ Coverage can be biased toward the pre-trained InceptionV3 model. Therefore, future studies should be conducted to identify which type of an evaluation backbone is appropriate for generative model evaluation.

\vspace{2mm} \noindent \textbf{IFID may lead to wrong conclusion.}
Intra-class FID~(IFID) is the average of class conditional FIDs and is used to measure intra-class fidelity, diversity, and class conditioning performance in a scalar value. As Sajjadi~\etal~\cite{sajjadi2018assessing} stated, quantifying multiple characteristics into a scalar leads to confusing conclusions, and independently measuring each characteristic is required. For example, ContraGAN exhibits a lower FID value~(21.14) than BigGAN~(24.4) in the Grandpa-ImageNet generation experiment. However, the IFID value of ContraGAN~(193.93) is exceptionally high compared to the IFID value of BigGAN (78.11). \newtext{In Table~\ref{table:class_accuracy}, we notice that the exceptionally high IFID score of ContraGAN attributes to its limited capability in label conditioning, exemplifying that assessing generative models based solely on FID is not recommended.}

\vspace{2mm} \noindent \textbf{Conditional generative models overlook class conditioning performance.} 
While conditional GAN~(cGAN) is devised to perform conditional image generation, studies that deal with cGAN put conditional image generation aside and mainly focus on generating realistic images. Although Classifier Accuracy Score~(CAS)~\cite{Ravuri2019ClassificationAS} is proposed to quantify class-conditional precision and recall of a generative model, it is not widely adopted due to the heavy computational cost induced by training ResNet~\cite{He_2016_CVPR} from scratch. 
We provide Top-1 and Top-5 classification accuracies using the InceptionV3 network instead. 
Table~\ref{table:class_accuracy} indicates that ContraGAN models are poor at conditional image generation. In addition, BigGAN is better at class conditioning than ReACGAN, which differs from our expectation that classifier-based GANs would perform better in image conditioning. 

\begin{table}[t]
    \caption{ImageNet classification accuracies on generated images from GANs. We use TensorFlow-InceptionV3~\cite{Szegedy2016RethinkingTI} to compute Top-1 accuracy, Top-5 accuracy, and FID.}
    \vspace{-2mm}
    \centering
    \resizebox{0.48\textwidth}{!}{
    \begin{tabular}{clrrr}
    \cmidrule[0.75pt]{2-5}
    & Model   & \text{Top-1 acc.}~$\uparrow$ & \text{Top-5 acc.}~$\uparrow$ & \text{FID}~$\downarrow$\\
    \cmidrule[0.75pt]{2-5}
    \parbox[t]{2mm}{\multirow{4}{*}{\rotatebox[origin=c]{90}{\text{Baby}}}}
    & BigGAN~\cite{Brock2019LargeSG}  & 53.65 & 72.09 & 18.64 \\ 
    & ContraGAN~\cite{kang2020contragan}  & 4.44 & 6.52 &  24.67 \\ 
    & ReACGAN~\cite{kang2021rebooting} & 50.71 & 70.37 & 19.75  \\
    & StyleGAN2~\cite{karras2020analyzing} & 44.04 & 62.03 &  22.21 \\ 
    \cmidrule[0.25pt]{2-5}
    \parbox[t]{2mm}{\multirow{4}{*}{\rotatebox[origin=c]{90}{\text{Papa}}}}
    & BigGAN~\cite{Brock2019LargeSG}  & 31.37 & 53.45 & 24.99 \\ 
    & ContraGAN~\cite{kang2020contragan}  & 4.06 & 9.80 &  25.33 \\ 
    & ReACGAN~\cite{kang2021rebooting} & 26.75 & 50.17 & 21.74 \\
    & StyleGAN2~\cite{karras2020analyzing} & 24.90 & 43.84 & 23.28 \\ 
    \cmidrule[0.25pt]{2-5}
    \parbox[t]{2mm}{\multirow{4}{*}{\rotatebox[origin=c]{90}{\text{Grandpa}}}}
    & BigGAN~\cite{Brock2019LargeSG}  & 23.68 & 46.56 & 24.40 \\ 
    & ContraGAN~\cite{kang2020contragan}  & 2.63 & 7.46 &  21.14 \\ 
    & ReACGAN~\cite{kang2021rebooting} & 24.48 & 50.56 & 18.65 \\
    & StyleGAN2~\cite{karras2020analyzing} & 19.36 & 38.88 & 21.88 \\ 
    \cmidrule[0.25pt]{2-5}
    \parbox[t]{2mm}{\multirow{4}{*}{\rotatebox[origin=c]{90}{\text{ImageNet}}}}
    & BigGAN~\cite{Brock2019LargeSG}  & 37.28 & 61.96 & 20.57 \\ 
    & ContraGAN~\cite{kang2020contragan}  & 2.40 & 10.26 &  29.59 \\ 
    & ReACGAN~\cite{kang2021rebooting} & 23.14 & 50.31 & 15.65 \\
    & StyleGAN2~\cite{karras2020analyzing} & 17.97 & 38.17 & 33.40 \\ 
    \cmidrule[0.25pt]{2-5}
    \end{tabular}}
    \label{table:class_accuracy}
    \vspace{-3mm}
\end{table}

\vspace{2mm} \noindent \textbf{Precision \& Recall and Density \& Coverage exhibit inconsistent results.} Precision \& Recall and Density \& Coverage are pairs of metrics devised to disentangle the fidelity and diversity measurement from FID. However, we observe the Recall and Coverage metrics occasionally draw conflicting judgments, while precision and density metrics often reach a consensus. As shown in Table~\ref{table:inception}, InceptionV3-based Recall values of ReACGAN on CIFAR10, Baby/Papa/Grandpa-ImageNet, and ImageNet are 0.62, 0.43, 0.47, 0.41, and 0.38 respectively. Likewise, Recalls of BigGAN are 0.65, 0.47, 0.47, 0.48, and 0.65, respectively. These results indicate that generated images by ReACGAN generally have lower Recall values than BigGAN. Coverage values, however, imply that the generated images by ReACGAN hold better diversity than BigGAN. This phenomenon also holds for experiments using SwAV. As a result, we believe an extra study is required to accurately quantify the fidelity and diversity of generated images using those metrics.

\section{Other Approaches}
\label{evaluation_other_generative_models}
\begin{table*}[!htp]
    \centering
    \caption{Benckmark table for GANs~\cite{Brock2019LargeSG, esser2021taming, kang2021rebooting, karras2021alias, sauer2022stylegan}, autoregressive models~\cite{esser2021taming, chang2022maskgit, lee2022autoregressive}, and diffusion probabilistic models~\cite{ho2020denoising, song2021scorebased, dhariwal2021diffusion, vahdat2021score, dockhorn2022score}. We evaluate those generative models using TensorFlow-InceptionV3~\cite{Szegedy2016RethinkingTI} and the architecture-friendly resizer: PIL.BILINEAR for the postprocessing step. In the case of ImageNet-128 and ImageNet-256 experiments, we preprocess images using the best performing resizer among PIL.BILINEAR, PIL.BICUBIC, and PIL.LANCZOS based on Fr\'echet Distance. We mark $\dag$ if images are preprocessed using PIL.BILINEAR for evaluation, $\dag\dag$ for the case of PIL.BICUBIC, and $\dag\dag\dag$ for the case of PIL.LANCZOS. Inference time is computed using a single A100 GPU. Top-1 and Top-2 performances are indicated in red and blue, respectively.}
    \vspace{-0.2cm}
    \resizebox{0.93\textwidth}{!}{
    \begin{tabular}{clcccccccc}
    \cmidrule[0.75pt]{2-10}
    & Model & \# Param. & \text{IS}~$\uparrow$ & \text{FID}~$\downarrow$ & \text{Precision}~$\uparrow$ & \text{Recall}~$\uparrow$ & \text{Density}~$\uparrow$ & \text{Coverage}~$\uparrow$ & Inf. (s)\\
    \cmidrule[0.75pt]{2-10}
    \parbox[t]{2mm}{\multirow{11}{*}{\rotatebox[origin=c]{90}{\text{CIFAR10}}}}
    & ReACGAN + DiffAug (Ours)~\cite{kang2021rebooting} & 9.4M & 10.15 & 2.64 & 0.75 & 0.65 & 0.98 & 0.90 & 0.009\\ 
    & StyleGAN2-ADA~\cite{kang2020contragan} & 20.2M & 10.31 & 2.41 & 0.74 & 0.68 & 1.02 & 0.92 & \textbf{\toptwo{0.008}}\\ 
    & StyleGAN2-ADA (Ours)~\cite{kang2020contragan} & 20.2M & \textbf{\toptwo{10.53}} & 2.31 & 0.75 & 0.69 & 1.04 & 0.93 & \textbf{\toptwo{0.008}}\\ 
    & StyleGAN2 + DiffAug + D2D-CE (Ours)~\cite{kang2021rebooting} & 20.2M & \placeholder{\topthree{10.46}} & 2.30 & 0.76 & 0.68 & 1.03 & 0.93 & \textbf{\topone{0.007}}\\ 
    & DDPM~\cite{ho2020denoising} & 35.2M & 9.73 & 3.23 & \textbf{\toptwo{0.78}} & 0.67 & 1.10 & 0.93 & 15.422\\ 
    & DDPM++~\cite{song2021scorebased} & 106.6M & 9.90 & 2.49 & \textbf{\toptwo{0.78}} & 0.69 & \textbf{\toptwo{1.12}} & \textbf{\toptwo{0.94}} & 46.697\\ 
    & NCSN++~\cite{song2021scorebased} & 107.6M & 10.08 & \placeholder{\topthree{2.27}} & 0.77 & \textbf{\toptwo{0.70}} & 1.07 & \textbf{\toptwo{0.94}} & 99.304\\ 
    & LSGM~\cite{vahdat2021score} & - & 10.04 & 2.80 & \textbf{\topone{0.80}} & \textbf{\toptwo{0.70}} & \textbf{\topone{1.15}} & \textbf{\topone{0.95}} & -\\ 
    & LSGM-ODE~\cite{vahdat2021score} & - & 10.07 & \textbf{\toptwo{2.09}} & 0.77 & \textbf{\topone{0.71}} & 1.03 & \textbf{\toptwo{0.94}} & - \\ 
    & CLD-SGM~\cite{dockhorn2022score} & - & 9.88 & 2.38 & \textbf{\toptwo{0.78}} & 0.69 & \textbf{\toptwo{1.12}} & \textbf{\toptwo{0.94}} & - \\ 
    & StyleGAN-XL~\cite{sauer2022stylegan} & 18.0M & \textbf{\topone{11.03}} & \textbf{\topone{1.88}} & 0.77 & 0.59 & 1.08 & \textbf{\toptwo{0.94}} & 0.010 \\ 
    \cmidrule[0.75pt]{2-10}
    \parbox[t]{2mm}{\multirow{6}{*}{\rotatebox[origin=c]{90}{\text{ImageNet-128}}}}
    & BigGAN~\cite{Brock2019LargeSG}$^{\dag\dag}$ & 70M & 128.51 & 7.66 & 0.80 & 0.51 & \placeholder{\topthree{1.16}} & 0.84 & 0.139 \\ 
    & BigGAN (Ours)\cite{Brock2019LargeSG}$^{\dag}$ & 70M & 98.51 & 8.54 & 0.75 & \placeholder{\topthree{0.60}} & 0.96 & 0.82 & \textbf{\topone{0.015}} \\ 
    & BigGAN-Deep~\cite{Brock2019LargeSG}$^{\dag\dag}$ & 50M & \textbf{\toptwo{161.73}} & \placeholder{\topthree{5.90}} & \textbf{\topone{0.89}} & 0.45 & \textbf{\topone{1.66}} & \placeholder{\topthree{0.91}} & 0.086 \\ 
    & ReACGAN (Ours)~\cite{kang2021rebooting}$^{\dag}$ & 70M & 94.06 & 8.19 & 0.80 & 0.42 & 1.07 & 0.72 & \textbf{\topone{0.015}} \\ 
    & StyleGAN-XL~\cite{sauer2022stylegan}$^{\dag\dag\dag}$ & 159M & \textbf{\topone{221.04}} & \textbf{\topone{1.94}} & \textbf{\toptwo{0.82}} & \textbf{\toptwo{0.62}} & \textbf{\toptwo{1.17}} & \textbf{\toptwo{0.94}} & 0.030 \\ 
    & ADM-G~\cite{dhariwal2021diffusion}$^{\dag\dag}$ & 464M & \placeholder{\topthree{156.22}} & \textbf{\toptwo{3.05}} & \placeholder{\topthree{0.81}} & \textbf{\topone{0.68}} & 1.13 & \textbf{\topone{0.95}} & 23.513 \\ 
    \cmidrule[0.75pt]{2-10}
    \parbox[t]{2mm}{\multirow{8}{*}{\rotatebox[origin=c]{90}{\text{ImageNet-256}}}}
    & BigGAN~\cite{Brock2019LargeSG}$^{\dag}$ & 164M & 185.52 & 7.75 & 0.83 & 0.47 & \placeholder{\topthree{1.34}} & 0.85 & 0.324 \\ 
    & BigGAN-Deep~\cite{Brock2019LargeSG}$^{\dag}$ & 112M & 224.46 & 6.95 & \textbf{\topone{0.89}} & 0.38 & \textbf{\toptwo{1.67}} & 0.88 & \textbf{\toptwo{0.177}} \\ 
    & VQGAN~\cite{esser2021taming}$^{\dag}$ & 1.5B & \textbf{\toptwo{314.61}} & 5.20 & 0.81 & 0.57 & 1.03 & 0.85 & 6.551 \\ 
    & StyleGAN-XL~\cite{sauer2022stylegan}$^{\dag\dag\dag}$ & 166M & \placeholder{\topthree{297.62}} & \textbf{\topone{2.32}} & 0.82 & \placeholder{\topthree{0.61}} & 1.16 & 0.93 & \textbf{\topone{0.040}} \\ 
    & ADM-G~\cite{dhariwal2021diffusion}$^{\dag}$ & 608M & 207.86 & 4.48 & 0.84 & \textbf{\topone{0.62}} & 1.19 & \textbf{\toptwo{0.94}} & 50.574 \\ 
    & ADM-G-U~\cite{dhariwal2021diffusion}$^{\dag}$ & 673M & 240.24 & \placeholder{\topthree{4.01}} & \placeholder{\topthree{0.85}} & \textbf{\topone{0.62}} & 1.22 & \textbf{\topone{0.95}} & 49.609 \\ 
    & MaskGIT~\cite{chang2022maskgit}$^{\dag}$ & 227M  & 216.38 & 5.40 & \textbf{\toptwo{0.87}} & 0.60 & \textbf{\toptwo{1.37}} & \textbf{\toptwo{0.94}} & 5.313 \\ 
    & RQ-Transformer~\cite{lee2022autoregressive}$^{\dag}$ & 3.9B & \textbf{\topone{339.41}} & \textbf{\toptwo{3.83}} & \placeholder{\topthree{0.85}} & 0.60 & 1.18 & 0.91 & 3.816 \\ 
    \cmidrule[0.75pt]{2-10}
    \end{tabular}}
    \label{table:other_generative_models_inception}
    \vspace{-2mm}
\end{table*}

This chapter evaluates other popular generative models, such as GANs with excessive parameters, autoregressive models, and probabilistic diffusion models. These approaches require excessive computational resources to be trained from scratch, so we utilize pre-trained models provided by authors to get generated images.
We follow our proposed evaluation protocol and preprocess the training images using the best-performed resizer among PIL.BILINEAR, PIL.BICUBIC, and PIL.LANCZOS based on FD value. The evaluation results are presented in Table~\ref{table:other_generative_models_inception}.

\vspace{2mm} \noindent \textbf{StyleGAN-XL.} StyleGAN-XL~\cite{sauer2022stylegan} tackles the limitation of the StyleGAN approaches that struggle with generating a diverse class of images, such as ImageNet.
In the CIFAR10 generation experiment, StyleGAN-XL exhibits the FID of 1.88, being first to break the wall of 2. Despite the record, StyleGAN-XL shows the lowest recall value of 0.59 compared to the other competitive models. This phenomenon is unnatural since images with a low recall are expected to exhibit a high FID. 

\newtext{A similar phenomenon is observed in the ImageNet-256 experiment. While StyleGAN-XL achieves the lowest FID value compared to other models, ADM-G-U presents higher precision and recall scores than StyleGAN-XL. This contradicts our expectation that higher precision and recall scores should be correlated with a lower FID. We leave the analysis on this matter as a future work because a more careful and thorough analysis should take place to understand this corner case of evaluation metrics.}

\vspace{2mm} \noindent \textbf{Auto-regressive (AR) and diffusion models.}
In Table~\ref{table:other_generative_models_inception}, RQ-Transformer \cite{lee2022autoregressive} shows the second best FID on ImageNet-256 followed by diffusion models ADM-G and ADM-G-U~\cite{dhariwal2021diffusion}. While StyleGAN-XL and RQ-Transformer exhibit lower FID values than ADM-G-U, we empirically confirm that ADM-G-U synthesizes more visually plausible images than StyleGAN-XL and RQ-Transformer. As can be seen in Figures~A12,~A13, and~A14 in Appendix, the generated images from ADM-G-U appear to reflect \newtext{the overall structure of objects more faithfully than StyleGAN-XL and RQ-Transformer.} To our knowledge, ADM-G-U is the first model to generate reasonable facial images among ImageNet-trained generative models.

\noindent \textbf{Comparing GANs with AR and diffusion models.} \newtext{\brown{Our experimental results suggest} that AR and diffusion models show comparable or slightly worse FIDs with StyleGAN-XL~(Table~\ref{table:other_generative_models_inception}). However, the number of model parameters in popular GAN models, such as BigGAN-Deep and StyleGAN-XL, is much smaller than AR and diffusion models. For instance, the number of parameters of BigGAN-Deep is 112M, but ADM-G and RQ-Transformer's parameters are 608M and 3.9B, respectively. This implies that GANs are parameter-efficient and can be scaled up to overcome the poor structural coherency~\cite{kang2023gigagan}. Furthermore, unlike AR and diffusion models, GANs do not require a consecutive inference process, enabling very fast sampling. For example, while it takes only 0.040 seconds to generate one sample with StyleGAN-XL, it takes upto 49.609 seconds with ADM-G-U (1000 DDPM).} \brown{While there have been efforts to reduce denoising steps~\cite{meng2023distillation, song2023consistency}, these models require a well-trained diffusion model for distillation. Moreover, the synthesis ability of the few-step diffusion models still lags behind the state-of-the-art GAN approach~\cite{kang2023gigagan}. Hence, we emphasize that GANs are highly efficient regarding model size, inference speed, and synthesis ability.}

\begin{figure*}[hbt!]
    StyleGAN-XL~\cite{sauer2022stylegan} on \textsc{ImageNet~(256 $\times$ 256)} \par\medskip
    \includegraphics[width=0.95\linewidth]{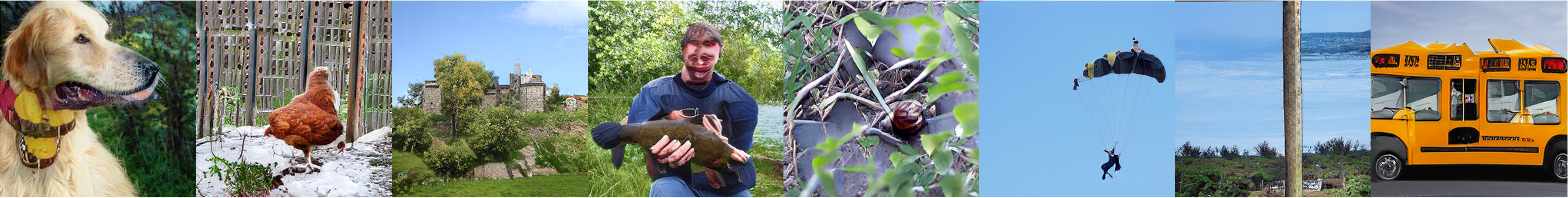} \\
    RQ-Transformer~\cite{lee2022autoregressive} on \textsc{ImageNet~(256 $\times$ 256)} \par\medskip
    \includegraphics[width=0.95\linewidth]{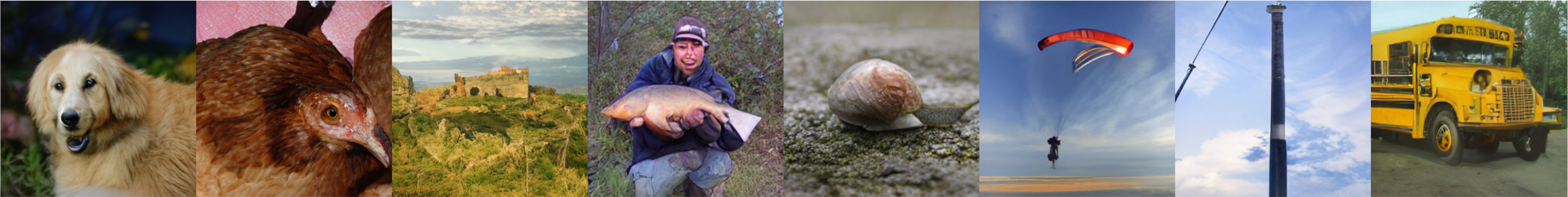} \\
    ADM-G-U~\cite{dhariwal2021diffusion} on \textsc{ImageNet~(256 $\times$ 256)} \par\medskip
    \includegraphics[width=0.95\linewidth]{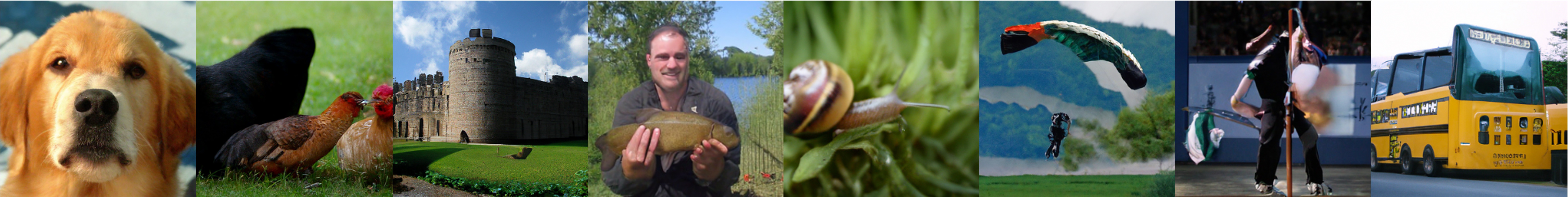} \\
    \vspace{-5mm}
    \caption{Generated images by StyleGAN-XL~(GAN)~\cite{sauer2022stylegan}, RQ-Transformer~(Auto Regressive Model)~\cite{lee2022autoregressive}, and ADM-G-U~(Denoising Diffusion Probabilistic Model)~\cite{dhariwal2021diffusion}. More qualitative results are displayed in Figures~A12, A13, and A14.}
    \label{fig:imagenet_other}
    \vspace{-4mm}
\end{figure*}

\section{Future research direction}
\label{sec:future_work}

\noindent \newtext{\textbf{Generative model evaluation.} \brown{Our paper primarily focuses on the training and evaluation protocols, as well as providing a comprehensive benchmark. However, it is still very difficult to explicitly specify which metrics researchers should use to develop and evaluate generative models. While widely adopted metrics like IS and FID have served as guiding principles in visual generative AI, several cases were reported where they do not perfectly align with the human evaluation of realism~\cite{borji2019pros}.} As a result, human evaluation is a crucial component for the accurate evaluation of generative models. Recently, several attempts, including Zhou~\etal~\cite{zhou2019hype}'s Human eYe Perceptual Evaluation (HYPE), try to introduce human-in-the-loop evaluation based on psychophysics. However, the time and cost involved in human evaluations limit the active use of this method in the community, and as a consequence, only a few selected models like SNGAN~\cite{Miyato2018SpectralNF}, BigGAN~\cite{Brock2019LargeSG}, and StyleGAN~\cite{karras2019style} offer human evaluation results. Given these circumstances, we believe developing more reliable human/automatic procedures will be critical in the future.}

\noindent \newtext{\textbf{Image synthesis on open-world visual images.} Recently, due to the amazing results from text-to-image synthesis models through a diffusion process, the generative modeling community is experiencing a rapid shift in technological trends from GANs to diffusion or auto-regressive (AR) models. With the development of text-to-image diffusion models, such as GLIDE~\cite{Nichol2022GLIDETP}, DALL$\cdot$E-2~\cite{ramesh2022hierarchical}, and Imagen~\cite{saharia2022photorealistic}, an enormous amount of GPUs are being devoted to training diffusion or AR models. GANs have yet to be actively utilized for text-to-image synthesis for their brittle nature of adversarial dynamics and their inability to synthesize structural details of objects within images faithfully. However, recent models, such as GigaGAN~\cite{kang2023gigagan} and StyleGAN-T~\cite{Sauer2023ARXIV} have demonstrated that GANs can be also scaled up and used for text-to-image synthesis on open-world datasets, such as LAION2B-en~\cite{schuhmann2022laion} and COYO-700M~\cite{kakaobrain2022coyo-700m}. This reveals that GANs can potentially serve as a breakthrough alternative to existing text-to-image models, which have suffered from slow inference time.}

\brown{On the other hand, there exist efforts to reduce the inference time of diffusion models through techniques like consistency distillation~\cite{song2023consistency} and progressive distillation~\cite{salimans2022progressive, meng2023distillation}. These approaches aim to make the synthesis process more efficient. We anticipate that these single-step generative models, including GANs, will be applied to various real-world applications. For example, one-step diffusion models can be effectively employed for customization tasks~\cite{kumari2023multi} that involve fine-tuning the model, while GANs can be utilized for image manipulation tasks~\cite{pan2023_DragGAN} by leveraging the linear latent space property.}
\section{Conclusion}
\label{sec:conclusion}

In this paper, we have introduced improper practices that arise when training and evaluating GAN. By correcting the above practices, we have presented training and evaluation protocols focusing on high-quality image generation and fair evaluation. Since current image synthesis benchmarks are not constructed using the proposed protocols, we have implemented extensive types of GANs based on our proposed GAN taxonomy to train and evaluate GANs from scratch. We provide StudioGAN, a software library of various GAN and relevant modules. Using the training and evaluation protocols and StudioGAN, we have offered a large-scale GAN evaluation benchmark and explained intriguing findings and potential hazards, which cast important questions for generative model development. In addition, we have evaluated recent image generation approaches such as ADM and RQ-Transformer. We hope StudioGAN serves as a cornerstone to explore GAN approaches and boosts the development of a next-generation generative AI.

\vspace{-1mm}
\ifCLASSOPTIONcompsoc
  \section*{Acknowledgments}
\else
  \section*{Acknowledgment}
\fi

This work is supported by IITP grants 2019-0-01906 (POSTECH AI Graduate School Program), 2021-0-00537 (Image Restoration), and 2022-0-00290 (Visual Intelligence) funded by the Korean government (MSIT).

\ifCLASSOPTIONcaptionsoff
  \newpage
\fi

\bibliographystyle{IEEEtran}
\bibliography{main_paper}
 
\nocite{Chen2016InfoGANIR}
\nocite{zhou2019lipschitz}
\nocite{Brock2017NeuralPE}
\nocite{Sinha2020TopkTO}

\begin{IEEEbiography}[{\includegraphics[width=1in,height=1.25in,clip,keepaspectratio]{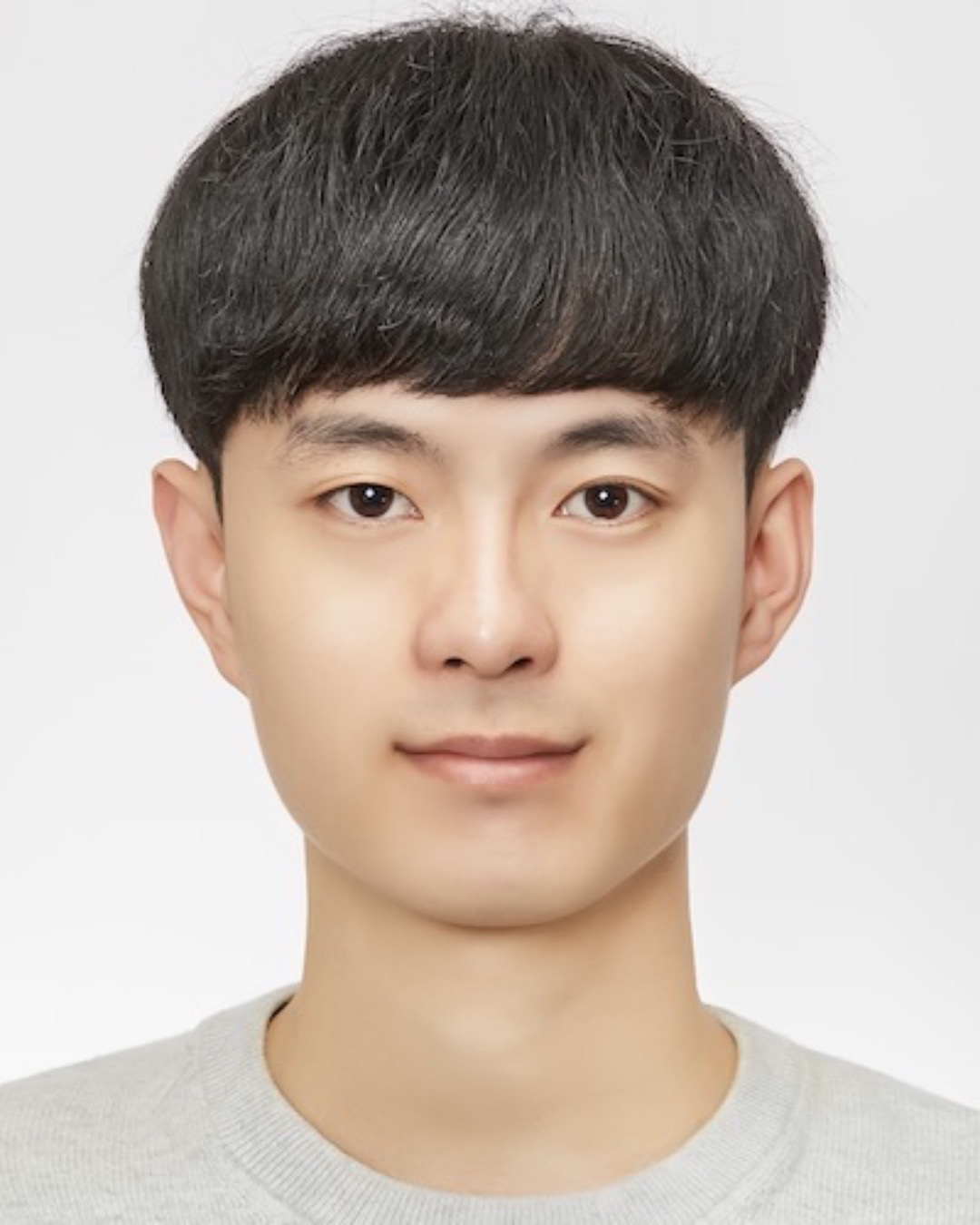}}]{Minguk Kang} received the BS degree from Pusan National University, South Korea, in 2019. He is currently working toward the PhD from the Graduate School of Artificial Intelligence, POSTECH, South Korea, where he created the StudioGAN library. His research interests include generative modeling, with applications in computer vision. He has authored three academic papers in NeurIPS and CVPR.
\end{IEEEbiography}
\vspace{-3mm}
\begin{IEEEbiography}[{\includegraphics[width=1in,height=1.25in,clip,keepaspectratio]{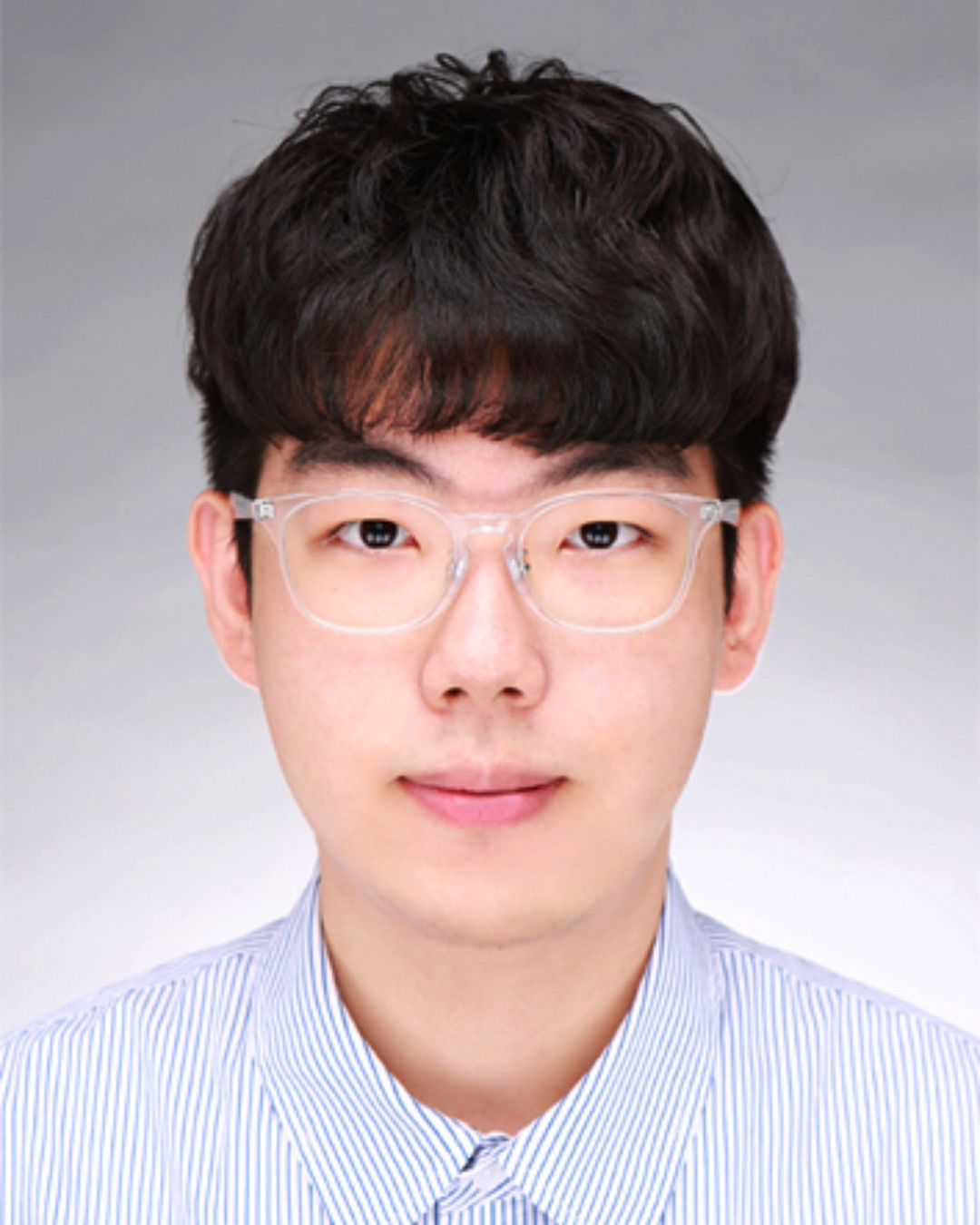}}]{Joonghyuk Shin} is a graduate student at the Department of Computer Science and Engineering, Seoul National University, South Korea. He received BS degree from the Department of Computer Science and Engineering, POSTECH in 2023. Along with Minguk, he is one of the core contributors of the StudioGAN library. His research interests include generative modeling and its diverse applications in computer vision. 
\end{IEEEbiography}
\vspace{-3mm}
\begin{IEEEbiography}[{\includegraphics[width=1in,height=1.25in,clip,keepaspectratio]{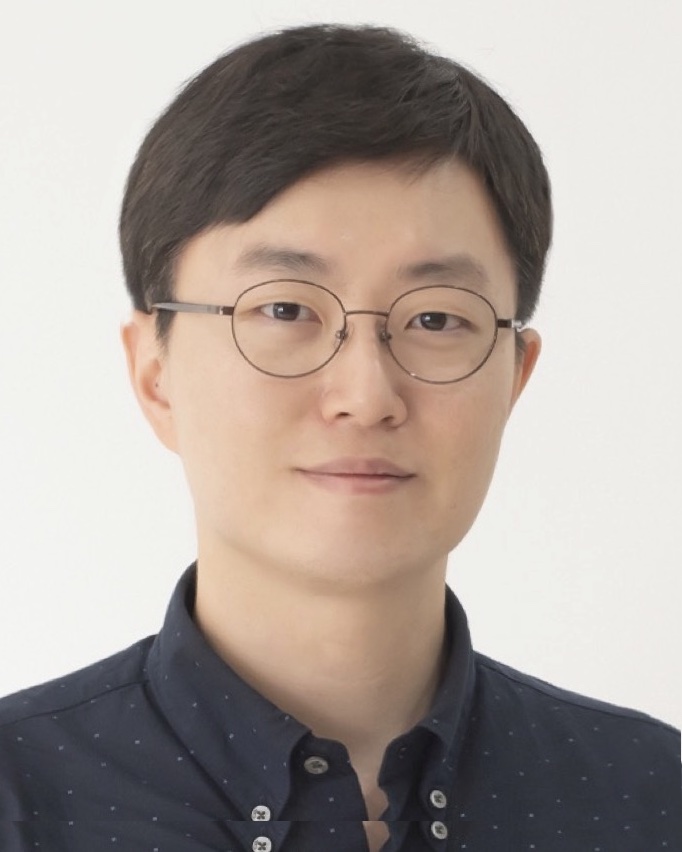}}]{Jaesik Park}
is an Assistant Professor at the Department of Computer Science and Engineering, Seoul National University. He received his Bachelor’s degree from Hanyang University (2009), and he received his Master’s degree (2011) and Ph.D. degree (2015) from KAIST. He worked at Intel as a research scientist (2015-2019), and he was a faculty member at POSTECH (2019-2023). His research interests include image synthesis and 3D scene understanding.
\end{IEEEbiography}

\clearpage
\section*{Appendix}
\renewcommand{\thetable}{\thesection\arabic{table}}
\renewcommand{\theHtable}{\thesection\arabic{table}}
\setcounter{table}{0}

\renewcommand{\thefigure}{\thesection\arabic{figure}}
\renewcommand{\theHfigure}{\thesection\arabic{figure}}
\setcounter{figure}{0}

\setcounter{section}{0}
\renewcommand{\thesection}{A\arabic{section}}
\renewcommand{\thesubsection}{\Alph{subsection}}
\renewcommand\thefigure{A\arabic{figure}}
\renewcommand{\thetable}{A\arabic{table}}

\setcounter{page}{1}
\section{Supported modules in StudioGAN}
\textbf{GANs}:
\begin{itemize}
    \item DCGAN~\cite{Radford2016UnsupervisedRL}
    \item InfoGAN~\cite{Chen2016InfoGANIR}
    \item LSGAN~\cite{Mao2017LeastSG}
    \item Geometric GAN~\cite{Lim2017GeometricG} 
    \item WGAN~\cite{Arjovsky2017WassersteinG} 
    \item WGAN-GP~\cite{Gulrajani2017ImprovedTO}
    \item WGAN-DRA~\cite{Kodali2018OnCA}
    \item ACGAN~\cite{Odena2017ConditionalIS} 
    \item PD-GAN~\cite{Miyato2018cGANsWP}
    \item SNGAN~\cite{Miyato2018SpectralNF}
    \item SAGAN~\cite{Zhang2019SelfAttentionGA}
    \item TACGAN~\cite{NIPS2019_8414}
    \item LGAN~\cite{zhou2019lipschitz}
    \item Unconditional BigGAN~\cite{Brock2019LargeSG}
    \item BigGAN~\cite{Brock2019LargeSG}
    \item BigGAN-Deep~\cite{Brock2019LargeSG}
    \item StyleGAN2~\cite{karras2020analyzing}
    \item CRGAN~\cite{Zhang2019ConsistencyRF}
    \item ICRGAN~\cite{Zhao2020ImprovedCR}
    \item LOGAN~\cite{Wu2019LOGANLO}
    \item ContraGAN~\cite{kang2020contragan}
    \item MHGAN~\cite{kavalerov2021multi}
    \item BigGAN-DiffAug~\cite{zhao2020differentiable}
    \item StyleGAN2-ADA~\cite{Karras2020TrainingGA}
    \item BigGAN-LeCam~\cite{lecam2021}, ADCGAN~\cite{hou2021cgans}
    \item ReACGAN~\cite{kang2021rebooting}
    \item StyleGAN2-APA~\cite{deceived2021}
    \item StyleGAN3-t~\cite{karras2021alias}
    \item StyleGAN3-r~\cite{karras2021alias}
\end{itemize}

\vspace{2mm} \noindent \textbf{Network Architectures}:
\begin{itemize}
    \item DCGAN~\cite{Radford2016UnsupervisedRL}
    \item ResNetGAN~\cite{Gulrajani2017ImprovedTO}
    \item SAGAN~\cite{Zhang2019SelfAttentionGA}
    \item BigGAN~\cite{Brock2019LargeSG}
    \item BigGAN-Deep~\cite{Brock2019LargeSG}
    \item StyleGAN2~\cite{karras2020analyzing}
    \item StyleGAN3~\cite{karras2021alias}.
\end{itemize} 

\vspace{2mm} \noindent \textbf{Conditioning methods}:
\begin{itemize}
    \item cGAN~\cite{Mirza2014ConditionalGA}
    \item ACGAN~\cite{Odena2017ConditionalIS}
    \item cBN~\cite{Dumoulin2017ALR, de_Vries, Miyato2018cGANsWP}
    \item PD-GAN~\cite{Miyato2018cGANsWP}
    \item TACGAN~\cite{NIPS2019_8414}
    \item MHGAN~\cite{kavalerov2021multi}
    \item ContraGAN~\cite{kang2020contragan}
    \item ADCGAN~\cite{hou2021cgans}
    \item ReACGAN~\cite{kang2021rebooting}
\end{itemize}

\vspace{2mm} \noindent \textbf{Adversarial losses}:
\begin{itemize}
    \item WGAN~\cite{Arjovsky2017WassersteinG}
    \item Vanilla GAN~\cite{Goodfellow2014GenerativeAN}
    \item LSGAN~\cite{Mao2017LeastSG}
    \item Geometric GAN~\cite{Lim2017GeometricG}
\end{itemize}

\vspace{2mm} \noindent \textbf{Regularizations}:
\begin{itemize}
    \item InfoGAN~\cite{Chen2016InfoGANIR}
    \item WGAN-GP~\cite{Gulrajani2017ImprovedTO}
    \item WGAN-DRA~\cite{Kodali2018OnCA}
    \item R1 Regularization~\cite{Mescheder2018ICML}
    \item SNGAN~\cite{Miyato2018SpectralNF}
    \item Orthogonal Regularization~\cite{Brock2017NeuralPE}
    \item PPL Regularization~\cite{karras2020analyzing}
    \item LGAN~\cite{zhou2019lipschitz}
    \item CRGAN~\cite{Zhang2019ConsistencyRF}
    \item LOGAN~\cite{Wu2019LOGANLO}
    \item Top-k Training~\cite{Sinha2020TopkTO}
    \item ICRGAN~\cite{Zhao2020ImprovedCR}
    \item LeCam~\cite{lecam2021}
\end{itemize}

\vspace{2mm} \noindent \textbf{Differentiable Augmentations}:
\begin{itemize}
    \item DiffAug~\cite{zhao2020differentiable}
    \item ADA~\cite{Karras2020TrainingGA}
    \item APA~\cite{deceived2021}
\end{itemize}

\section{Datasets}
\label{appendix:datasets_and_evaluation_metrics}
\begin{figure*}[ht!]
    \centering
    \includegraphics[width=0.49\linewidth]{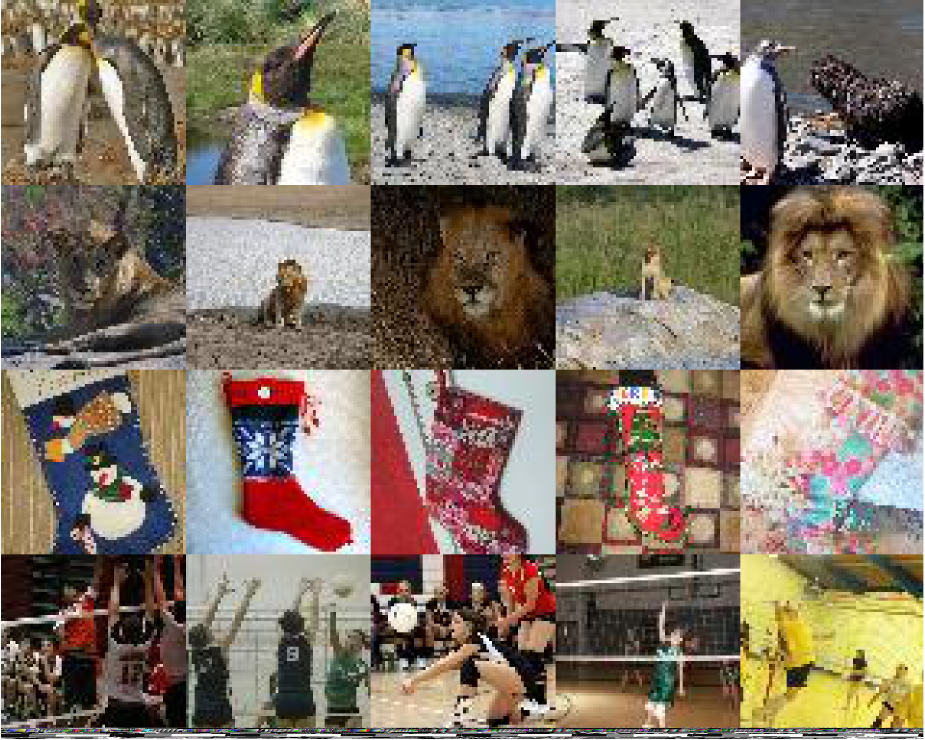}
    \includegraphics[width=0.49\linewidth]{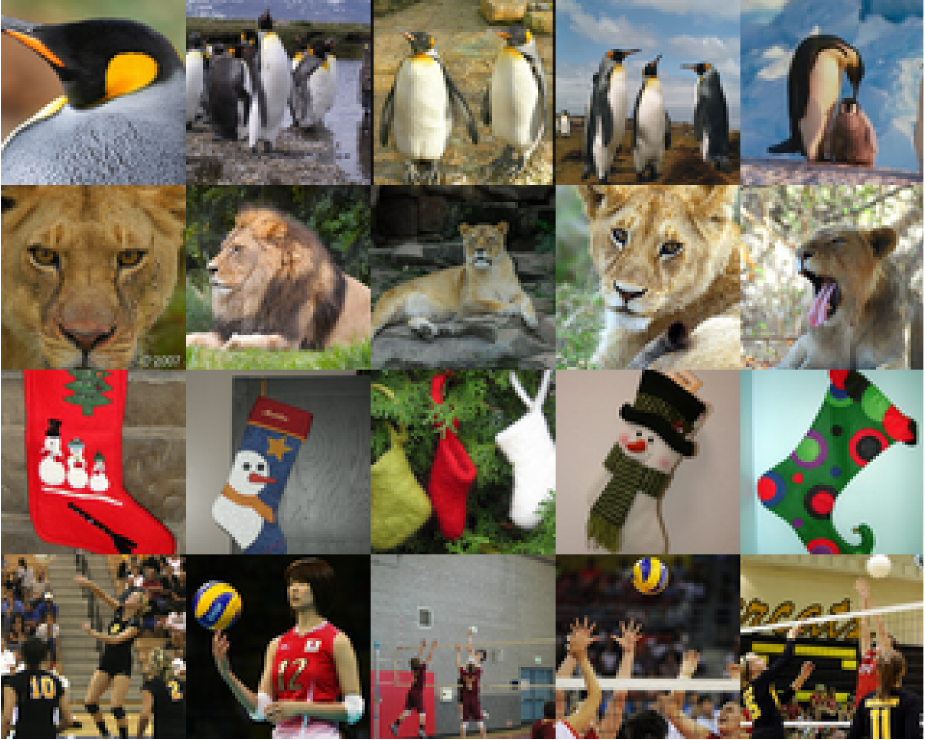}
    \caption{Images sampled from Tiny-ImageNet dataset (64$\times$64 pix.)~\cite{Tiny} (Shown on the left) and from Baby-ImageNet dataset (64$\times$64 pix.)~(Shown on the right). The images of Tiny-ImageNet exhibit severe image degradation due to improper data preprocessing. Images from Baby-ImageNet are resized to 64$\times$64 resolution using PIL.LANCZOS resizer, which results in better image quality.}
    \label{fig:baby_imagenet_visualization}
\end{figure*}
\noindent \textbf{CIFAR10}~\cite{Krizhevsky2009LearningML} is comprised of 50k train, and 10k test RGB images of $32 \times 32$ resolution. There are a total of ten classes, resulting in 5k train images and 1k test images per class. For the CIFAR10 10\% dataset, we randomly chose 500 images for each class. And for the CIFAR10 30\% dataset, we added 1000 more randomly selected images on top of the 10\% dataset. 

\vspace{2mm}
\noindent \textbf{Baby, Papa, and Grandpa ImageNet} are newly introduced subsets of the ImageNet dataset to replace old Tiny-ImageNet~\cite{Tiny}. Using InceptionV3 network's top-1 classification accuracy of ImageNet 1,000 classes, we sample 1$\sim$100 (most accurate), 451$\sim$550 (modest), and 901$\sim$1,000 (difficult) classes and groupe them into Baby, Papa, and Grandpa ImageNet, respectively. With Baby ImageNet being the easiest and Grandpa ImageNet being the hardest, images are center cropped and then resized into 64 x 64 resolution using the high-quality PIL.LANCZOS resizer~\cite{Turkowski1990FiltersFC} explained in section~\ref{sec:aliasing_quantization}. The resulting images show much less degradation and aliasing artifacts compared to the legacy Tiny-ImageNet as can be seen in Figure~\ref{fig:baby_imagenet_visualization}.

\vspace{2mm}
\noindent \textbf{ImageNet}~\cite{Deng2009ImageNetAL} provides around 1.2M and 50k RGB images in 1,000 classes for training and validation. We apply center crop to every image and resize images to 128 x 128 pixels using PIL.LANCZOS resizer as stated above.

\vspace{2mm}
\noindent \textbf{AFHQv2}~\cite{choi2020starganv2} is an improved version of the previous Animal Face High-Quality dataset (AFHQ)~\cite{choi2020starganv2} by utilizing Lanczos resampling instead of nearest neighbor downsampling. Total 15,803 RGB images (14,336 train, 1,467 test) with 512 x 512 resolution fall into three classes (cat, dog, and wild). Likewise, we resize images to 512 $\times$ 512 pixels using Lanczos resizer for Tables~\ref{table:reproducibility},~\ref{table:inception},~\ref{table:swav},~\ref{table:swin} and 256 $\times$ 256 pixels for Table~\ref{table:data_efficient} whose purpose is identifying the effectiveness of modules for data-efficient training.

\vspace{2mm}
\noindent \textbf{FFHQ (FQ for short)}~\cite{karras2019style} consists of 70k high-quality face images at 1024$\times$1024 resolution. The dataset contains abundant facial variations including gender, age, ethnicity, and image background.

\section{Training and evaluation setups}
\label{sec:training_setups}

For experiments, we report averaged value of three runs for CIFAR10, two runs for Baby/Papa/Grandpa-ImageNet, and a single run for ImageNet, AFHQv2, and FFHQ. All experiments were performed using mixed-precision training, and experiments that had collapsed at the early stage were re-executed. We empirically find that unexpected training failures occasionally occur when mixed precision is applied. We attribute this phenomenon to the extremely sensitive GAN dynamics. Using mixed precision requires scaling of floating points between FP16 and FP32, and some information loss is inevitable. While this may not be a big problem for most neural networks, we suspect that this has enormous effects on GAN dynamics which are susceptible to even the slightest gradient changes. 

With a lone exception of the CIFAR10 experiment that was done by a single GPU, all models were trained with 4-GPUs with proper accumulation for batch statistics. Since the variation during the evaluation process is relatively small, we evaluate every model once. (\textit{i.e.} One evaluation for each of the three runs for CIFAR10, summing up to three for each configuration) All evaluation was done on a single GPU.

Further details for training and evaluation are summarized below:

\begin{itemize}
\item We use PIL.LANCZOS resizer to preprocess training images (Sec.~\ref{sec:aliasing_quantization}) and apply a random horizontal flip.
\item We use PIL.BILINEAR resizer to process real and fake images for evaluation if InceptionV3 and SwAV backbones are applied. For the evaluation using Swin-T, we use PIL.BICUBIC resizer (Sec.~\ref{sec:pil_bicubic_is_good}).
\item For every experiment, we use the training set for the evaluation, not the validation set. 
\item We use the same number of generated images as the training images for Fr\'echet Distance (FD), Precision, Recall, Density, and Coverage calculation. For the experiments using Baby/Papa/Grandpa-ImageNet and ImageNet, we exceptionally use 50k fake images against a complete training set as real images.
\item When evaluating GANs, we always apply the standing statistics technique~\cite{Brock2019LargeSG} except for the experiments using StyleGAN2~\cite{karras2020analyzing} or StyleGAN3~\cite{karras2021alias}. As mentioned in Sec.~\ref{sec:normalization}, two hyperparameters (\textit{maximum batch size} and \textit{number of repetition}) are required for applying standing statistics. Empirically, we found that these hyperparameters perform nicely, when set to the same value of training batch size. (\textit{e.g.} If batch size was 256, we use 256 for both \textit{maximum batch size} and \textit{number of repetition}.)
\item We use PyTorch implementation of InceptionV3~\cite{pytorchttur} whose network weights are converted from OpenAI's TensorFlow-InceptionV3 network~\cite{openaiinception}. In the SwAV and Swin-T models, we use authors' official codes and pretrained network weights~\cite{officialswav, officialswin}.
\item When we calculate Intra-Class FD, we use the training images selected for each class as the reference and utilize the same amount of generated images for comparison.
\end{itemize}

\section{Evaluating Data-Efficient Techniques}
\label{guildline_for_data_efficient_training}
\begin{figure*}[!ht]
    \centering
    \vspace{2mm}
    \includegraphics[width=0.99\linewidth]{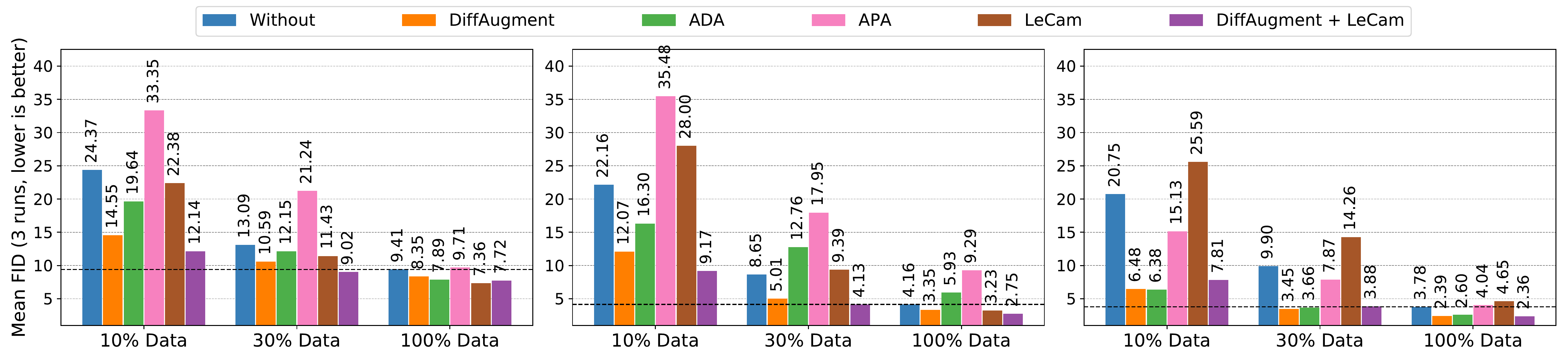}
    \caption{FID values of various data-efficient training methods applied to three GAN backbones: ResNet~\cite{Gulrajani2017ImprovedTO}, BigGAN~\cite{Brock2019LargeSG}, and StyleGAN2~\cite{karras2020analyzing}. The dotted line indicates the performance of GAN trained using the full images without any data-efficient training trick.}
    \label{fig:cifar10_data_efficient_FID}
\end{figure*}

\begin{table*}[!htp]
    \caption{Data-efficient training results. We resize AFHQv2 to 256 $\times$ 256 resolutions using PIL.LANCZOS resizer for training. Top-1 and Top-2 performances are indicated in red and blue, respectively.}
    \centering
    \vspace{-0.2cm}
    \resizebox{0.99\textwidth}{!}{
    \begin{tabular}{clrrrrrrrr}
    \cmidrule[0.75pt]{2-9}
    & Model + Data-efficient trick   & \text{IS}~$\uparrow$ & \text{FID}~$\downarrow$ & \text{Precision}~$\uparrow$ & \text{Recall}~$\uparrow$ & \text{Density}~$\uparrow$ & \text{Coverage}~$\uparrow$ & Avg. Rank\\
    \cmidrule[0.75pt]{2-9}
    \parbox[t]{2mm}{\multirow{12}{*}{\rotatebox[origin=c]{90}{\text{AFHQv2}}}}
    & BigGAN~\cite{Brock2019LargeSG}  & 7.64 & 29.13 & 0.86 & 0.02 & 1.30 & 0.45 & 8.00 \\ 
    & BigGAN + DiffAug~\cite{zhao2020differentiable} & 6.81 & 51.58 & 0.78 & 0.00 & 1.16 & 0.20 & 10.00 \\
    & BigGAN + ADA~\cite{Karras2020TrainingGA} & 11.70 & \placeholder{\topthree{5.52}} & 0.84 & \textbf{\toptwo{0.51}} & 1.32 & \textbf{\topone{0.85}} & 4.33 \\ 
    & BigGAN + LeCam~\cite{lecam2021} & 8.27 & 30.72 & \textbf{\topone{0.89}} & 0.05 & \placeholder{\topthree{1.49}} & 0.46 & 6.17 \\ 
    & BigGAN + APA~\cite{deceived2021} & 6.20 & 68.37 & 0.31 & 0.00 & 0.17 & 0.05 & 11.67 \\ 
    & BigGAN + DiffAug + LeCam~\cite{zhao2020differentiable, lecam2021} & 6.31 & 54.03 & 0.42 & 0.00 & 0.24 & 0.07 & 10.83 \\ 
    \cmidrule[0.25pt]{2-9}
    & StyleGAN2~\cite{karras2020analyzing} & 11.68 & 7.21 & 0.87 & 0.36 & 1.45 & 0.82 & 5.67 \\ 
    & StyleGAN2 + DiffAug~\cite{zhao2020differentiable} & \textbf{\toptwo{12.29}} & 6.13 & \textbf{\topone{0.89}} & 0.31 & \textbf{\topone{1.57}} & \placeholder{\topthree{0.84}} & \textbf{\toptwo{3.17}} \\ 
    & StyleGAN2 + ADA~\cite{Karras2020TrainingGA} & \textbf{\topone{12.31}} & \textbf{\topone{5.13}} & 0.86 & \placeholder{\topthree{0.48}} & 1.33 & \textbf{\topone{0.85}} &  \textbf{\topone{2.83}} \\ 
    & StyleGAN2 + LeCam~\cite{lecam2021} & 11.88 & 7.16 & 0.86 & 0.37 & 1.43 & 0.82 & 5.17 \\ 
    & StyleGAN2 + APA~\cite{deceived2021} & 11.7 & \textbf{\toptwo{5.44}} & 0.83 & \textbf{\topone{0.53}} & 1.18 & 0.83 & 5.17 \\ 
    & StyleGAN2 + DiffAug + LeCam~\cite{zhao2020differentiable, lecam2021} & \placeholder{\topthree{11.98}} & 6.05 & \placeholder{\topthree{0.88}} & 0.41 & \textbf{\toptwo{1.53}} & \placeholder{\topthree{0.84}} & \textbf{\toptwo{3.17}} \\ 
    \cmidrule[0.75pt]{2-9}
    \end{tabular}}
    \label{table:data_efficient}
\end{table*}
\noindent \textbf{Experiment setup.} We test various data-efficient training techniques to identify which technique works well in data-hungry situations. We select augmentation-based methods~(DiffAug~\cite{zhao2020differentiable}, ADA~\cite{Karras2020TrainingGA}, and APA~\cite{deceived2021}) and regularization-based method~(LeCam~\cite{lecam2021}). We also test whether an augmentation-based method can harmonize with the regularization-based method. We test DiffAug and LeCam with the same hyperparameters used in DiffAug and LeCam experiments. Figure~\ref{fig:cifar10_data_efficient_FID} shows results for 10\%, 30\%, and 100\% of CIFAR10 dataset on ResNet, BigGAN, and StyleGAN2 backbones, and Table~\ref{table:data_efficient} shows results for AFHQv2 dataset in 256$\times$256 resolution with BigGAN and StyleGAN2 backbones. We train CIFAR10 models for 500k iterations or fewer in case of early collapse for a fair evaluation. Note that the FID value of the StyleGAN2-ADA model continues to drop until 2.31 at around 1.5M iterations, as can be seen in Table~\ref{table:other_generative_models_inception}, but here we train all models up to 500k iterations to check the tendency. All AFHQv2 models were trained for 200k iterations, and it is reported in Table~\ref{table:other_generative_models_inception}.

\vspace{2mm} \noindent \textbf{Results.} \brown{The results reveal that as the amount of data decreases, the FID scores notably worsen. However, figure~\ref{fig:cifar10_data_efficient_FID} provides evidence that utilizing appropriate data-efficient training techniques can significantly alleviate the adverse effects of training GAN with limited data, leading to improved results.}  DiffAug and DiffAug + LeCam work best in generating CIFAR10 images in the extreme case when only 10\% or 30\% is provided. BigGAN with ADA occasionally exhibits worse performances than those without data-efficient techniques. However, StyleGAN2 consistently improves with ADA. \newtext{Furthermore, Table~\ref{table:data_efficient} shows that BigGAN can only successfully generate AFHQv2 images when trained with ADA, implying that choosing an appropriate data-efficient training method is essential in limited data scenarios, and ADA is often a favorable option.}

Figure~\ref{fig:cifar10_data_efficient_FID} shows that DiffAug with LeCam presents the best or second best FID values across three GAN backbones on CIFAR10 except for StyleGAN2 with 10\% and 30\% of data. It's also worthwhile to note that BigGAN with DiffAug + LeCam gives a decent FID of 4.13 with only 30\% of the data, which is quite impressive considering that BigGAN without any techniques displays FID of 4.16 with complete CIFAR10 data. While the combination of DiffAug and LeCam does not present better performance than ADA in the AFHQv2 experiment, we think that DiffAug can work in harmony with LeCam when appropriate hyperparameters for GAN training are selected.

We also observe APA and LeCam sometimes showing substandard performance when applied solely. In the case of APA, we think it is because APA was mainly targeted for the StyleGAN backbone concentrating on center-aligned datasets like FFHQ and AFHQv2. Figure~\ref{fig:cifar10_data_efficient_FID}(c) and Table~\ref{table:data_efficient} prove that APA works better with StyleGAN2 backbone and even better when combined with center aligned images like AFHQv2. LeCam, on the other hand, was originally devised for diversified class conditional datasets like CIFAR10 and ImageNet with considerations for both BigGAN and StyleGAN2 backbones. Authors of LeCam emphasize the power of this regularization when combined with augmentation methods. Our experiment results, although not perfectly fit, mostly correspond to the authors' idea that the combination of DiffAug + LeCam shows the best results for CIFAR10 across most settings. Although original paper~\cite{lecam2021} suggests LeCam, alone works better than DiffAug on 10\% ImageNet, our experiment results on 10\% and 30\% CIFAR10 display the limitation of this regularization-based method when applied solely in hugely data-hungry circumstances.

AFHQv2 generation experiments in Table~\ref{table:data_efficient} show that StyleGAN2 + ADA presents better Recall and Coverage values than that of StyleGAN2 + DiffAug. On the other hand, StyleGAN2 + ADA presents lower Precision and Density values than that of StyleGAN2 + DiffAug. DiffAug applies three types of image augmentations with fixed probability, whereas ADA applies diverse augmentations that are determined probabilistically. We speculate that such nature of DiffAug makes images more discriminative but less diverse than ADA.

\section{Impact of post-resizer}
\newtext{Table~\ref{table:other_generative_models_inception} presents Fr\'echet Distances for models trained on ImageNet-256 with varying combinations of the backbone network and post-resizer. While SwAV and InceptionV3 backbones are the most and second robust to the post-resizer change, we observe that Swin-T backbone evaluations are heavily affected by different post-resizing methods. For Swin-T backbone evaluation results, rankings among other models frequently change when a different post-resizer is adopted. VQGAN, StyleGAN-XL, and ADM-G-U notably perform better when choosing a bicubic resizer. Furthermore, if models are evaluated without a consistent post-resizer, we even observe BigGAN with bilinear resizer (19.15) outperforming BigGAN-Deep with Lanczos resizer (30.36) in the Swin-T backbone. For the InceptionV3 backbone, the bilinear resizer usually shows the lowest score with a noticeable gap with the Lanczos resizer. As a result, we summarize that model rankings can change not only by the different backbone but also by different post-resizers, adding more significance to a unified evaluation protocol with a consistent post-resizer.}

\begin{table*}[!htp]
    \centering
    \caption{Impact of post resizer. We evaluate various generative models trained on ImageNet-256 with varying combination of backbone network and post resizer. }
    \vspace{-0.2cm}
    \resizebox{1.0\textwidth}{!}{
    \begin{tabular}{clccccccccc}
    \cmidrule[0.75pt]{2-10}
    & & BigGAN~\cite{Brock2019LargeSG} & BigGAN-Deep~\cite{Brock2019LargeSG} & VQGAN~\cite{esser2021taming} & StyleGAN-XL~\cite{sauer2022stylegan} & ADM-G-U~\cite{dhariwal2021diffusion} & ADM-G~\cite{dhariwal2021diffusion} & MaskGIT~\cite{chang2022maskgit} & RQ-Transformer~\cite{lee2022autoregressive} \\
    \cmidrule[0.75pt]{2-10}
    \parbox[t]{2mm}{\multirow{3}{*}{\rotatebox[origin=c]{90}{\text{IncepV3}}}}
    & Bilinear & 7.75	& 6.95	& 5.20	& 2.51	& 4.01	& 4.48	& 5.40 & 3.83 \\ 
    & Bicubic~\cite{Keys1981CubicCI} & 8.11	& 7.26	& 5.27	& 2.34	& 4.10	& 4.94	& 5.74 & 4.00\\ 
    & Lanczos~\cite{Turkowski1990FiltersFC} & 8.26	& 7.41	& 5.33	& 2.32	& 4.16	& 5.16	& 5.92 & 4.14\\ 
    \cmidrule[0.75pt]{2-10}
    \parbox[t]{2mm}{\multirow{3}{*}{\rotatebox[origin=c]{90}{\text{SwAV}}}}
    & Bilinear & 4.02 & 3.57 & 3.03 & 1.10	& 1.78 & 1.54 & 2.48 & 2.14 \\ 
    & Bicubic~\cite{Keys1981CubicCI} & 4.02 & 3.60	& 3.03 & 1.08 & 1.81 & 1.60 & 2.49 & 2.16 \\ 
    & Lanczos~\cite{Turkowski1990FiltersFC} & 4.05 & 3.63 & 3.05 & 1.08 & 1.83 & 1.62 & 2.52 & 2.18 \\ 
    \cmidrule[0.75pt]{2-10}
    \parbox[t]{2mm}{\multirow{3}{*}{\rotatebox[origin=c]{90}{\text{Swin-T}}}}
    & Bilinear & 19.15 & 13.93 & 13.45 & 8.61 &  9.61 &  8.82 & 13.33 &  9.2  \\ 
    & Bicubic~\cite{Keys1981CubicCI} & 23.59 & 16.97 &  11.66 &  5.43 & 7.81 &  8.37 &  16.88 &   11.09 \\ 
    & Lanczos~\cite{Turkowski1990FiltersFC} & 30.36 &  22.5 &  13.79 &  6.18 &  9.76  &  11.16 &  22.61 &  16.07  \\ 
    \cmidrule[0.75pt]{2-10}
    \end{tabular}}
    \label{table:other_model_resizer}
\end{table*}

\section{Evaluation using other backbones}
\begin{figure*}[ht]
    \includegraphics[width=0.95\linewidth]{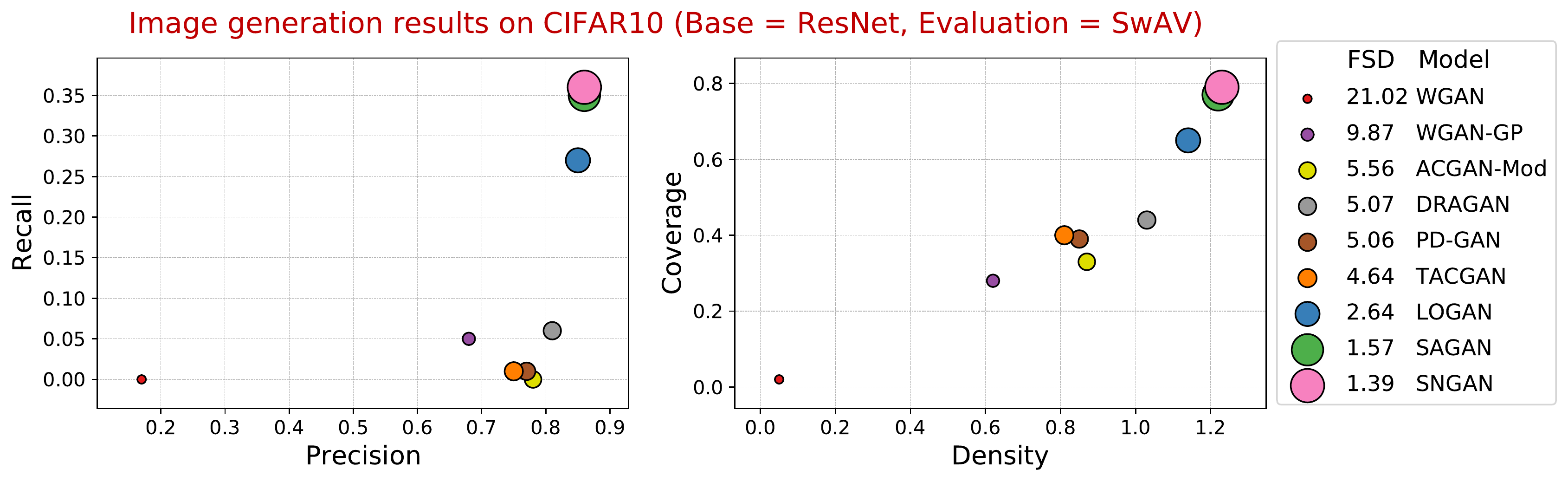}\\
    \includegraphics[width=0.95\linewidth]{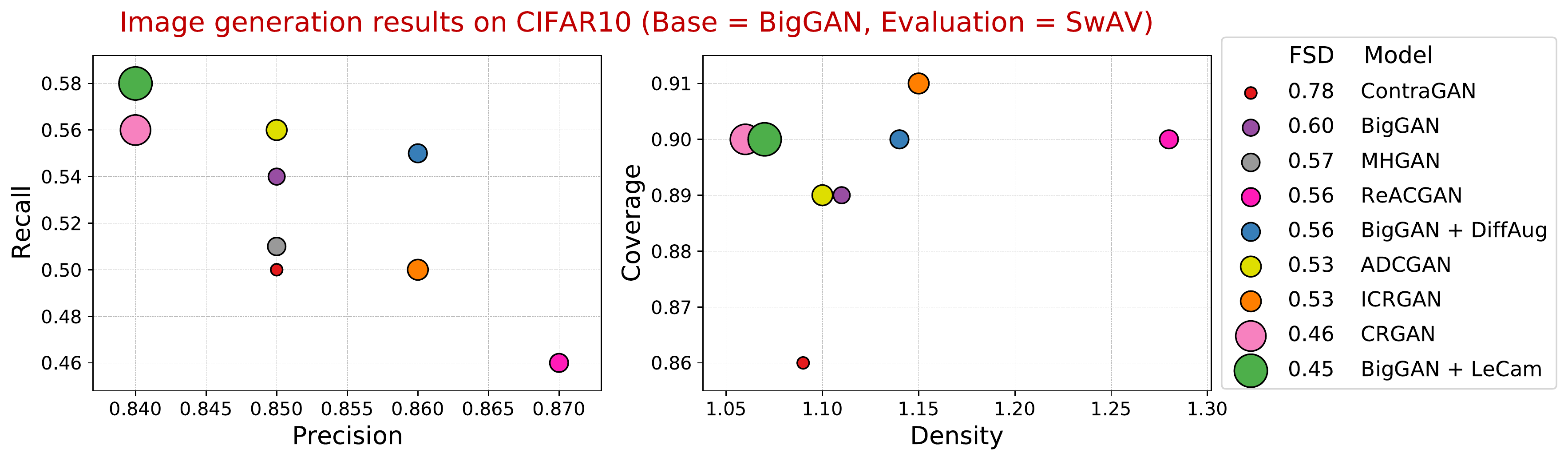}\\
    \includegraphics[width=0.95\linewidth]{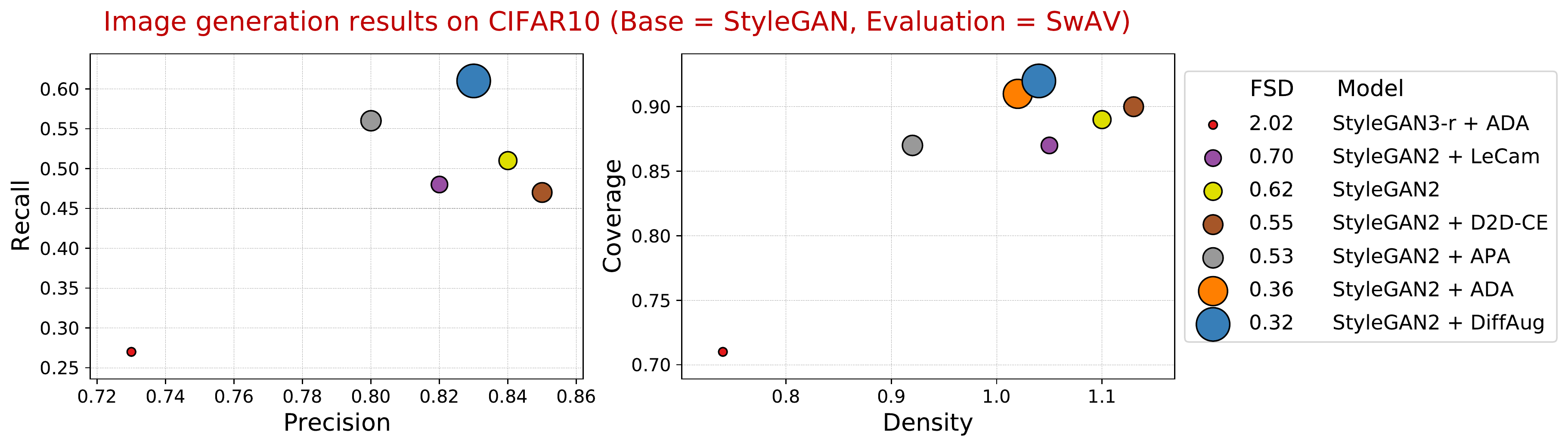}
    \caption{Scatter plots for analyzing GANs from the perspective of fidelity and diversity. We adopt PyTorch-SwAV~\cite{Caron2020UnsupervisedLO} as the evaluation backbone.}
    \label{fig:cifar10_swav_scatter}
\end{figure*}

\begin{figure*}[!ht]
    \includegraphics[width=0.95\linewidth]{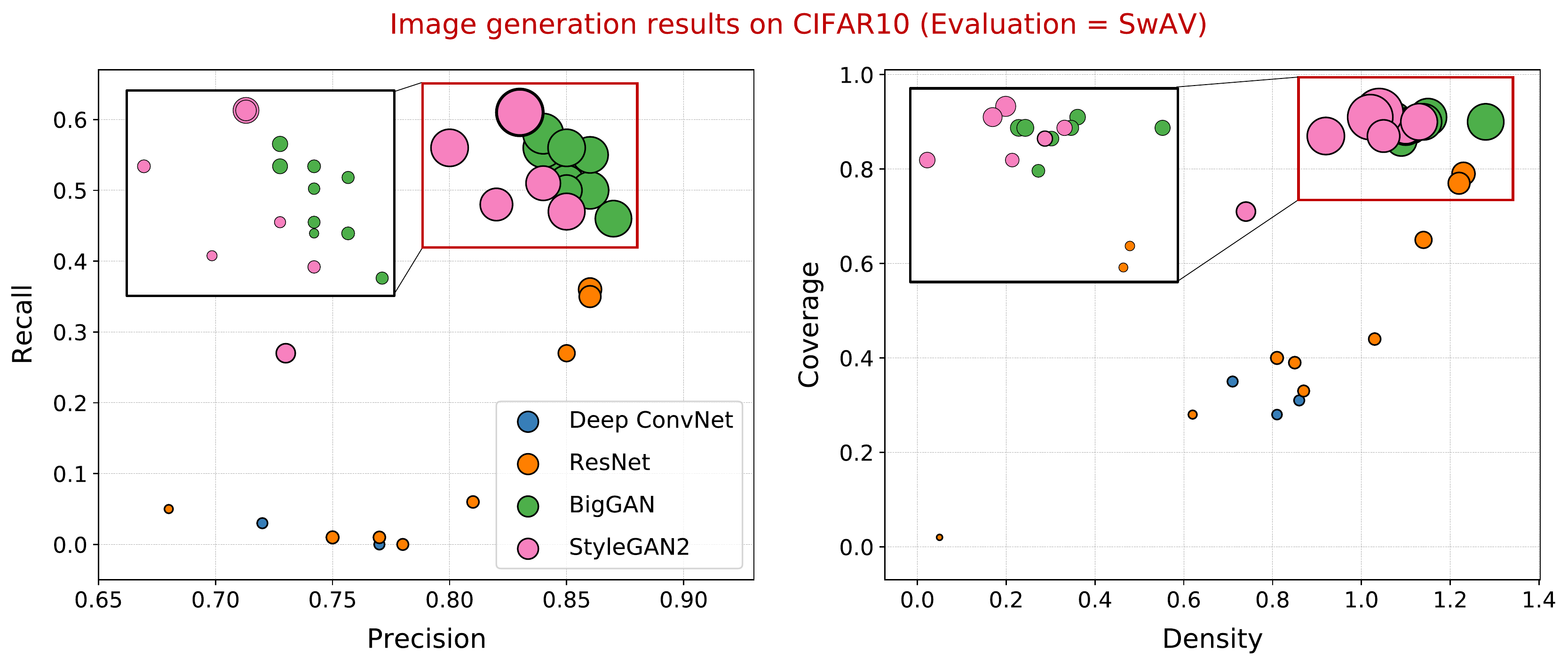}
    \vspace{-3mm}
    \caption{Scatter plots for analyzing the characteristics of distinct GAN architectures. We train GANs on CIFAR10~\cite{Krizhevsky2009LearningML} dataset and evaluate each GAN using SwAV~\cite{Caron2020UnsupervisedLO} network.}
    \label{fig:cifar10_swav_whole_scatter}
\end{figure*}

\begin{figure*}[!ht]
    \includegraphics[width=0.94\linewidth]{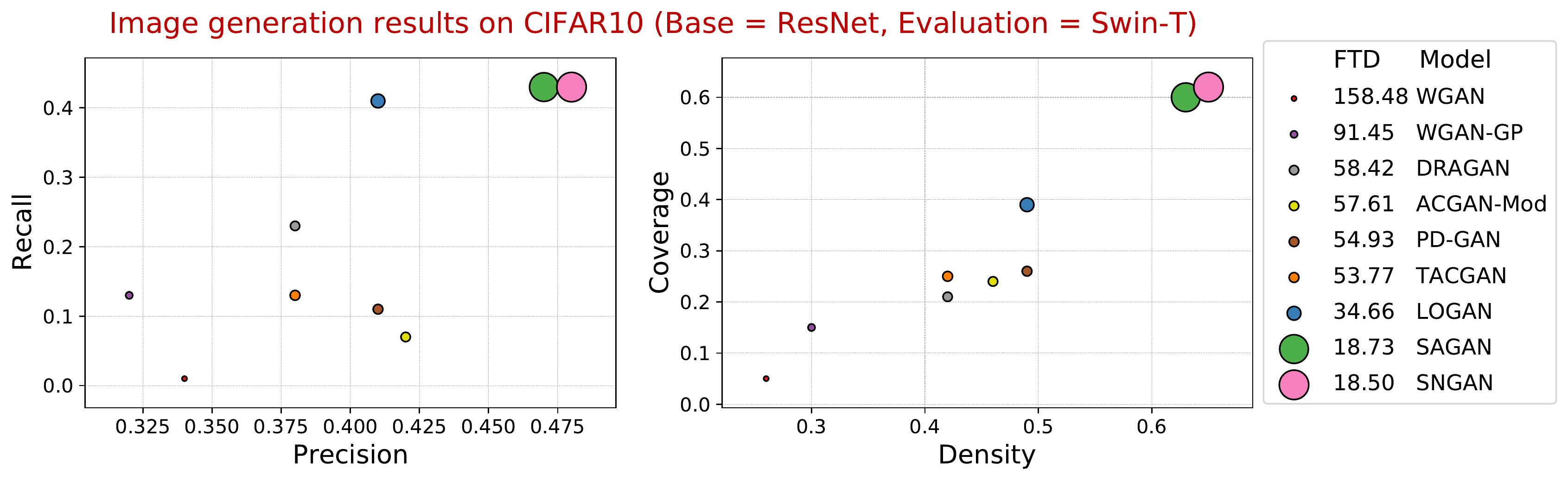}\\
    \includegraphics[width=0.94\linewidth]{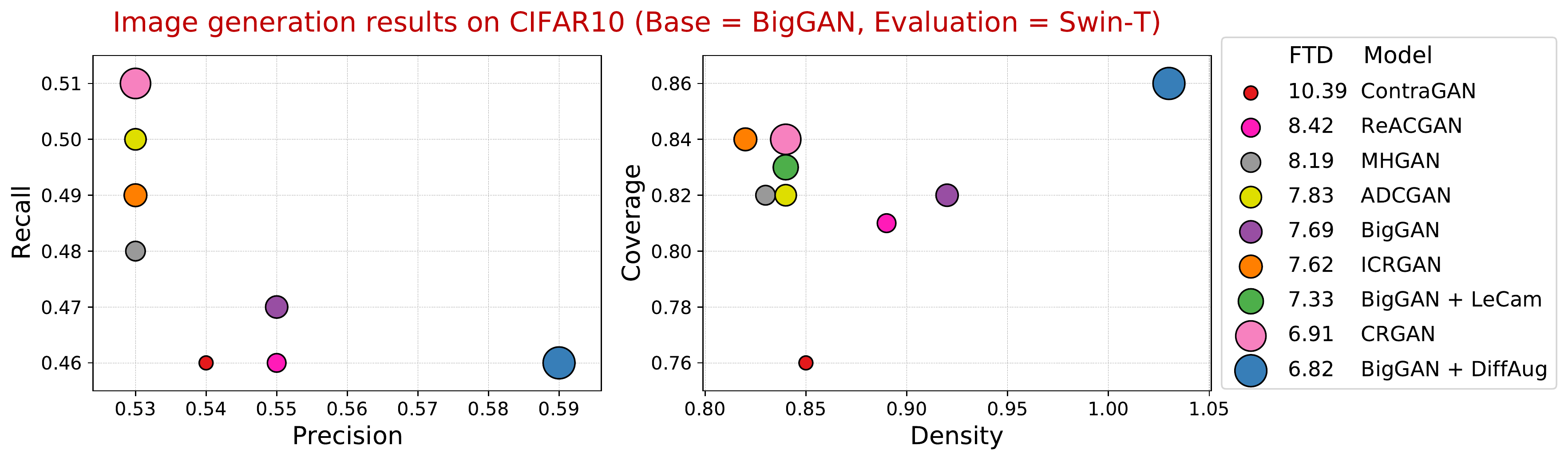}\\
    \includegraphics[width=0.94\linewidth]{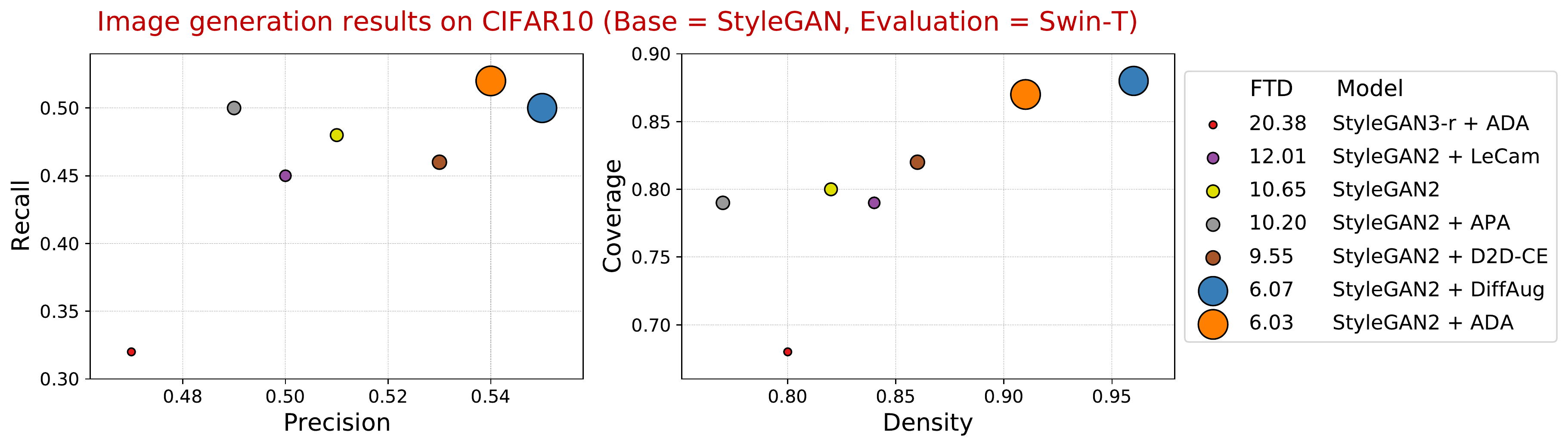}    
    \caption{Scatter plots for analyzing GANs from the perspective of fidelity and diversity. We use PyTorch-Swin-T~\cite{liu2021swin} as the evaluation backbone.}
    \label{fig:cifar10_swin_scatter}
\end{figure*}

\begin{figure*}[!ht]
    \includegraphics[width=0.93\linewidth]{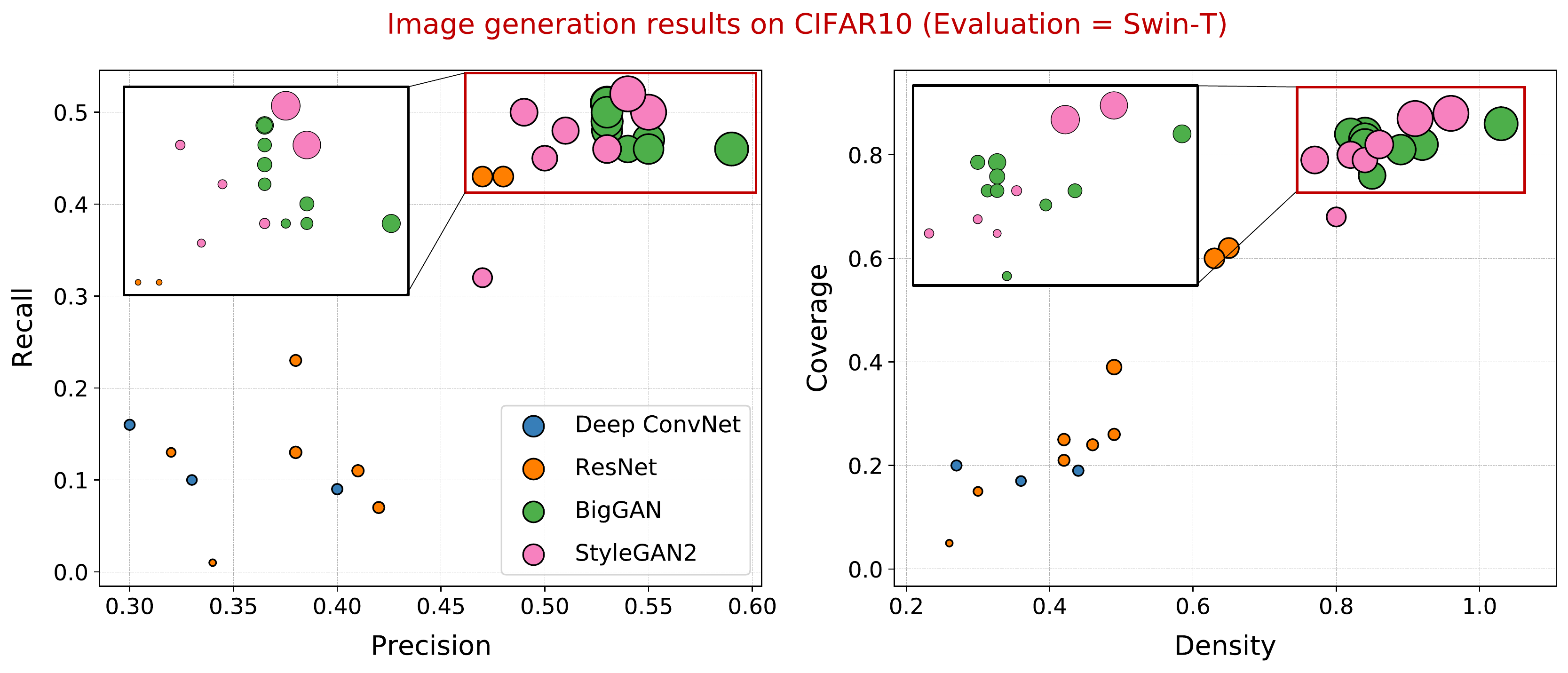}
    \vspace{-3mm}
    \caption{Scatter plots for analyzing the characteristics of distinct GAN architectures. We train GANs on CIFAR10~\cite{Krizhevsky2009LearningML} dataset and evaluate each GAN using Swin-T~\cite{liu2021swin} network.}
    \label{fig:cifar10_swin_whole_scatter}
\end{figure*}

\begin{figure*}[!ht]
    \includegraphics[width=0.95\linewidth]{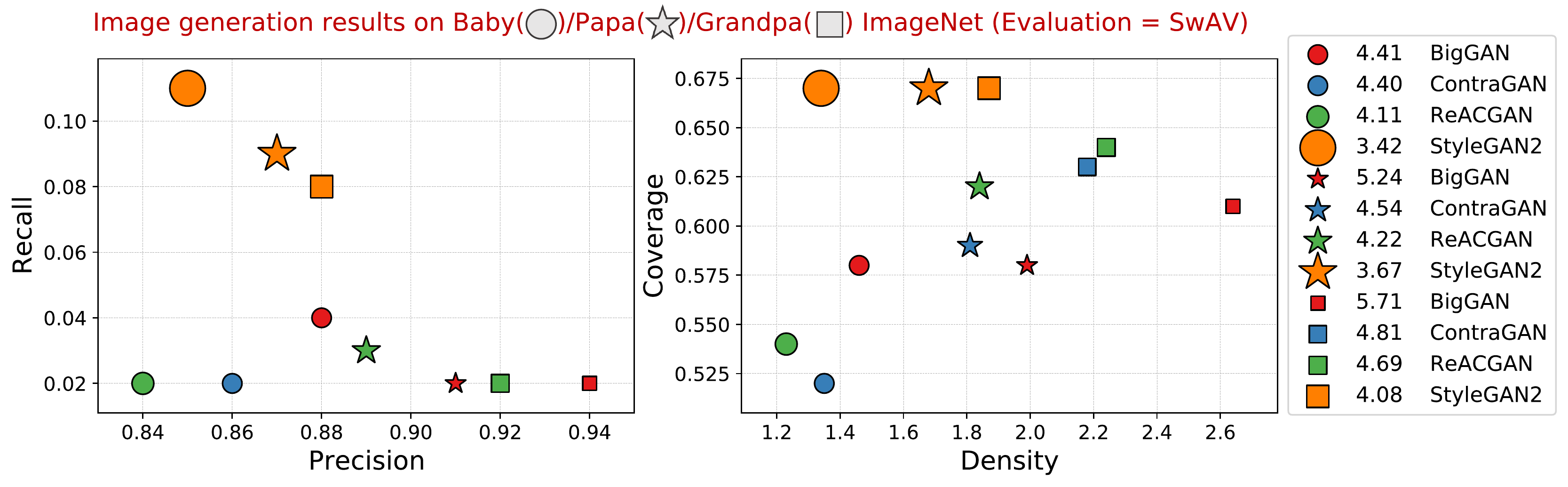}
    \vspace{-3mm}
    \caption{Scatter plots for analyzing the characteristics of distinct GAN architectures. We train GANs on Baby, Papa, and Grandpa ImageNet datasets and evaluate those GANs using PyTorch-SwAV~\cite{Caron2020UnsupervisedLO}.}
    \label{fig:imagenet_series_swav_scatter}
\end{figure*}

\begin{figure*}[!ht]
    \includegraphics[width=0.95\linewidth]{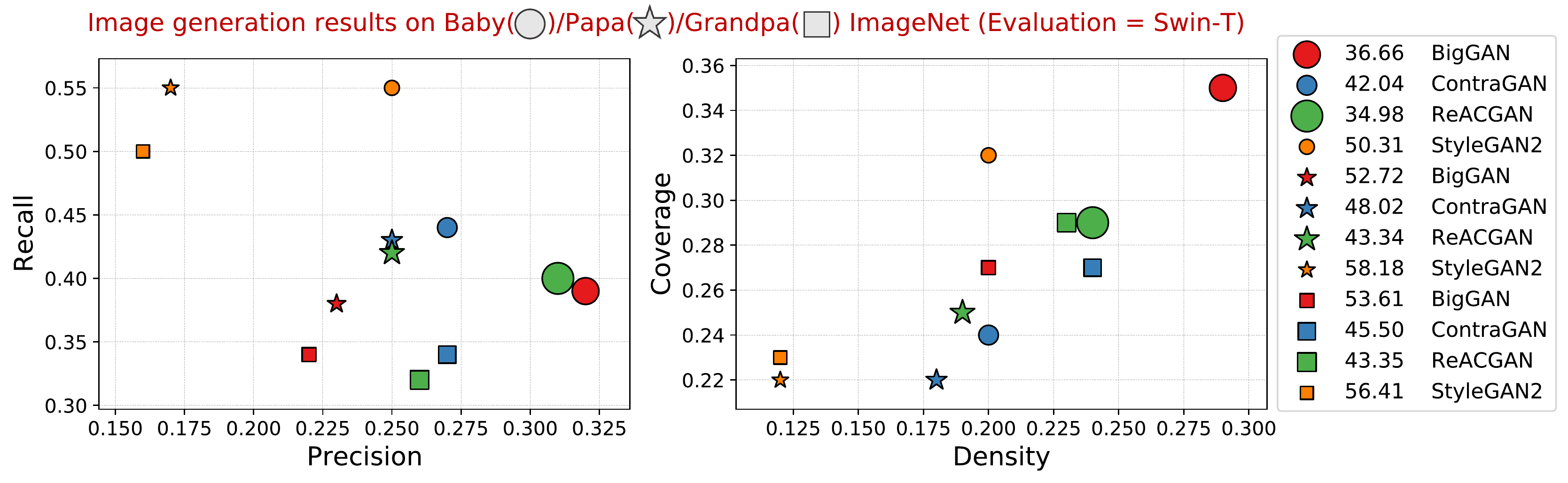}
    \vspace{-3mm}
    \caption{Scatter plots for analyzing the characteristics of distinct GAN architectures. We train GANs on Baby, Papa, and Grandpa ImageNet datasets and evaluate those GANs using PyTorch-Swin-T~\cite{liu2021swin}.}
    \label{fig:imagenet_series_swin_scatter}
\end{figure*}

We provide evaluation results using \textbf{SwAV}~\cite{Caron2020UnsupervisedLO} and \textbf{Swin-T}~\cite{liu2021swin} backbones in Tables~\ref{table:swav} and~\ref{table:swin}. Using the quantitative results, we visualize the results in the form of scatter plot in Figures~\ref{fig:cifar10_swav_scatter},~\ref{fig:cifar10_swin_scatter},~\ref{fig:cifar10_swav_whole_scatter},~\ref{fig:cifar10_swin_whole_scatter},~\ref{fig:imagenet_series_swav_scatter}, and~\ref{fig:imagenet_series_swin_scatter}.

\begin{table*}[!htp]
    \caption{Benchmark table evaluated using \textbf{PyTorch-SwAV}~\cite{Caron2020UnsupervisedLO}. The resolutions of CIFAR10~\cite{Krizhevsky2009LearningML}, Baby/Papa/Grandpa ImageNet, ImageNet~\cite{Deng2009ImageNetAL}, AFHQv2~\cite{choi2020starganv2, karras2021alias}, and FQ~\cite{karras2019style} datasets are 32, 64, 128, 512, and 1024, respectively. Top-1 and Top-2 performances are indicated in red and blue, respectively.}
    \vspace{-0.2cm}
    \centering
    \resizebox{0.88\textwidth}{!}{
    \begin{tabular}{clrrrrrrr}
    \cmidrule[1.0pt]{2-9}
    & \textbf{PyTorch-SwAV}~\cite{Caron2020UnsupervisedLO} & \text{SS}~$\uparrow$ & \text{FSD}~$\downarrow$ & \text{S-Precision}~$\uparrow$ & \text{S-Recall}~$\uparrow$ & \text{S-Density}~$\uparrow$ & \text{S-Coverage}~$\uparrow$ & \text{IFSD}~$\downarrow$\\
    \cmidrule[1.0pt]{2-9}
    \parbox[t]{2mm}{\multirow{29}{*}{\rotatebox[origin=c]{90}{\text{CIFAR10}}}}
    & DCGAN~\cite{Goodfellow2014GenerativeAN} & 6.32 & 7.28 & 0.77 & 0.00 & 0.81 & 0.28 & 13.13\\ 
    & LSGAN~\cite{Mao2017LeastSG} & 7.05 & 6.64 & 0.72 & 0.03 & 0.71 & 0.35 & 12.81\\ 
    & Geometric GAN~\cite{Lim2017GeometricG} & 6.65 & 6.67 & 0.77 & 0.00 & 0.86 & 0.31 & 12.77\\ 
    \cmidrule[0.25pt]{2-9}
    & ACGAN-Mod~\cite{Odena2017ConditionalIS} & 7.46 & 5.56 & 0.78 & 0.00 & 0.87 & 0.33 & 9.64\\ 
    & WGAN~\cite{Arjovsky2017WassersteinG} & 4.21 & 21.02 & 0.17 & 0.00 & 0.05 & 0.02 & 25.21\\ 
    & DRAGAN~\cite{Kodali2018OnCA} & 7.10 & 5.07 & 0.81 & 0.06 & 1.03 & 0.44 & 11.28\\ 
    & WGAN-GP~\cite{Gulrajani2017ImprovedTO} & 5.78 & 9.87 & 0.68 & 0.05 & 0.62 & 0.28 & 15.35\\ 
    & PD-GAN~\cite{Miyato2018cGANsWP} & 8.38 & 5.06 & 0.77 & 0.01 & 0.85 & 0.39 & 8.24\\ 
    & SNGAN~\cite{Miyato2018SpectralNF} & 10.79 & 1.39 & \textbf{\toptwo{0.86}} & 0.36 & \textbf{\toptwo{1.23}} & 0.79 & 2.61\\ 
    & SAGAN~\cite{Zhang2019SelfAttentionGA} & 10.26 & 1.57 & \textbf{\toptwo{0.86}} & 0.35 & \placeholder{\topthree{1.22}} & 0.77 & 2.85 \\ 
    & TACGAN~\cite{NIPS2019_8414} & 8.18 & 4.64 & 0.75 & 0.01 & 0.81 & 0.40 & 8.10 \\ 
    & LOGAN~\cite{Wu2019LOGANLO} & 8.61 & 2.64 & 0.85 & 0.27 & 1.14 & 0.65 & 9.45 \\ 
    \cmidrule[0.25pt]{2-9}
    & BigGAN~\cite{Brock2019LargeSG} & 13.63 & 0.60 & 0.85 & 0.54 & 1.11 & 0.89 & 1.53 \\ 
    & CRGAN~\cite{Zhang2019ConsistencyRF} & 14.21 & 0.46 & 0.84 & 0.56 & 1.06 & 0.90 & \placeholder{\topthree{1.35}} \\ 
    & MHGAN~\cite{kavalerov2021multi} & 13.61 & 0.57 & 0.85 & 0.51 & 1.10 & 0.89 & 1.62 \\ 
    & ICRGAN~\cite{Zhao2020ImprovedCR} & 14.00 & 0.53 & \textbf{\toptwo{0.86}} & 0.50 & 1.15 & \textbf{\toptwo{0.91}} & 1.44 \\ 
    & ContraGAN~\cite{kang2020contragan} & 13.27 & 0.78 & 0.85 & 0.50 & 1.09 & 0.86 & 13.36\\ 
    & BigGAN + DiffAug~\cite{zhao2020differentiable} & 13.46 & 0.56 & \textbf{\toptwo{0.86}} & 0.55 & 1.14 & 0.90 & 1.39\\ 
    & BigGAN + LeCam~\cite{lecam2021} & 13.98 & \placeholder{\topthree{0.45}} & 0.84 & \placeholder{\topthree{0.58}} & 1.07 & 0.90 & \placeholder{\topthree{1.35}} \\ 
    & ADCGAN~\cite{hou2021cgans} & 13.35 & 0.53 & 0.85 & 0.56 & 1.10 & 0.89 & 1.44 \\ 
    & ReACGAN~\cite{kang2021rebooting} & 13.44 & 0.56 & \textbf{\topone{0.87}} & 0.46 & \textbf{\topone{1.28}} & 0.90 & 1.71 \\ 
    \cmidrule[0.25pt]{2-9}
    & StyleGAN2~\cite{karras2020analyzing} & 13.83 & 0.62 & 0.84 & 0.51 & 1.10 & 0.89 & 1.53 \\ 
    &  StyleGAN2 + DiffAug~\cite{karras2020analyzing, zhao2020differentiable} &  \textbf{\topone{15.18}} &  \textbf{\topone{0.32}} &  0.83 &  \textbf{\topone{0.61}} &  1.04 &  \textbf{\topone{0.92}} &  \textbf{\topone{1.15}} \\ 
    & StyleGAN2 + ADA~\cite{Karras2020TrainingGA} & \textbf{\toptwo{15.02}} & \textbf{\toptwo{0.36}} & 0.83 & \textbf{\topone{0.61}} & 1.02 & \textbf{\toptwo{0.91}} & \textbf{\toptwo{1.20}} \\ 
    & StyleGAN2 + LeCam~\cite{karras2020analyzing, lecam2021} & 13.64 & 0.70 & 0.82 & 0.48 & 1.05 & 0.87 & 1.68 \\ 
    & StyleGAN2 + APA~\cite{deceived2021} & 14.09 & 0.53 & 0.80 & 0.56 & 0.92 & 0.87 & 5.81 \\ 
    & StyleGAN2 + D2D-CE~\cite{kang2021rebooting} & \placeholder{\topthree{14.50}} & 0.55 & 0.85 & 0.47 & 1.13 & 0.9 & 1.62 \\ 
    & StyleGAN3-r + ADA~\cite{karras2021alias} & 13.73 & 2.02 & 0.73 & 0.27 & 0.74 & 0.71 & 3.65 \\ 
    \cmidrule[1.0pt]{2-9}

    \parbox[t]{2mm}{\multirow{8}{*}{\rotatebox[origin=c]{90}{\text{Baby-ImageNet}}}}
    & SNGAN~\cite{Miyato2018SpectralNF} & 17.85 & 5.79 & \textbf{\toptwo{0.87}} & \placeholder{\topthree{0.03}} & \textbf{\toptwo{1.48}} & 0.45 & 10.87\\ 
    & SAGAN~\cite{Zhang2019SelfAttentionGA} & 14.52 & 6.42 & \textbf{\toptwo{0.87}} & 0.02 & \textbf{\topone{1.58}} & 0.40 & 12.02 \\ 
    \cmidrule[0.25pt]{2-9}
    &  BigGAN~\cite{Brock2019LargeSG} &  \textbf{\toptwo{34.47}} &  4.41 &  \textbf{\topone{0.88}} &  \textbf{\toptwo{0.04}} &  \placeholder{\topthree{1.46}} &  \textbf{\toptwo{0.58}} &  \textbf{\toptwo{9.08}} \\ 
    & ContraGAN~\cite{kang2020contragan} & 29.40 & \placeholder{\topthree{4.40}} & 0.86 & 0.02 & 1.35 & 0.52 & 21.53 \\ 
    & ReACGAN~\cite{kang2021rebooting} & \textbf{\topone{34.83}} & \textbf{\toptwo{4.11}} & 0.84 & 0.02 & 1.23 & \placeholder{\topthree{0.54}} & \placeholder{\topthree{10.45}} \\ 
    \cmidrule[0.25pt]{2-9}
    & StyleGAN2~\cite{karras2020analyzing} & \placeholder{\topthree{33.07}} & \textbf{\topone{3.42}} & 0.85 & \textbf{\topone{0.11}} & 1.34 & \textbf{\topone{0.67}} & \textbf{\topone{8.38}} \\ 
    & StyleGAN3-t~\cite{karras2021alias} & 9.03 & 11.59 & 0.68 & 0.00 & 0.68 & 0.08 & 22.44 \\ 
    \cmidrule[1.0pt]{2-9}
    \parbox[t]{2mm}{\multirow{8}{*}{\rotatebox[origin=c]{90}{\text{Papa-ImageNet}}}}
    & SNGAN~\cite{Miyato2018SpectralNF} & 12.65 & 6.36 & \textbf{\toptwo{0.89}} & 0.02 & 1.76 & 0.46 & 11.59 \\ 
    & SAGAN~\cite{Zhang2019SelfAttentionGA} & 10.26 & 6.90 & 0.88 & 0.02 & 1.80 & 0.41 & 12.65 \\ 
    \cmidrule[0.25pt]{2-9}
    & BigGAN~\cite{Brock2019LargeSG} & \placeholder{\topthree{19.81}} & 5.24 & \textbf{\topone{0.91}} & 0.02 & \textbf{\topone{1.99}} & 0.58 & \textbf{\toptwo{10.16}} \\ 
    & ContraGAN~\cite{kang2020contragan} & 19.72 & \placeholder{\topthree{4.54}} & \textbf{\toptwo{0.89}} & \textbf{\toptwo{0.03}} & \placeholder{\topthree{1.81}} & \placeholder{\topthree{0.59}} & 19.72 \\ 
    &  ReACGAN~\cite{kang2021rebooting} &  \textbf{\topone{22.47}} &  \textbf{\toptwo{4.22}} &  \textbf{\toptwo{0.89}} &  \textbf{\toptwo{0.03}} &  \textbf{\toptwo{1.84}} &  \textbf{\toptwo{0.62}} &  \placeholder{\topthree{11.00}} \\ 
    \cmidrule[0.25pt]{2-9}
    & StyleGAN2~\cite{karras2020analyzing}  & \textbf{\toptwo{20.84}} & \textbf{\topone{3.67}} & 0.87 & \textbf{\topone{0.09}} & 1.68 & \textbf{\topone{0.67}} & \textbf{\topone{8.38}} \\ 
    & StyleGAN3-t~\cite{karras2021alias} & 8.84 & 11.27 & 0.77 & 0.00 & 1.28 & 0.12 & 22.58 \\ 
    \cmidrule[1.0pt]{2-9}
    \parbox[t]{2mm}{\multirow{8}{*}{\rotatebox[origin=c]{90}{\text{Grandpa-ImageNet}}}}
    & SNGAN~\cite{Miyato2018SpectralNF} & 10.82 & 7.09 & 0.91 & \textbf{\toptwo{0.02}} & \textbf{\toptwo{2.26}} & 0.47 & 12.74 \\ 
    & SAGAN~\cite{Zhang2019SelfAttentionGA} & 10.05 & 8.03 & 0.90 & 0.01 & 2.19 & 0.42 & 14.61 \\ 
    \cmidrule[0.25pt]{2-9}
    & BigGAN~\cite{Brock2019LargeSG} & 16.29 & 5.71 & \textbf{\topone{0.94}} & \textbf{\toptwo{0.02}} & \textbf{\topone{2.64}} & 0.61 & \textbf{\toptwo{10.70}} \\ 
    & ContraGAN~\cite{kang2020contragan} & \textbf{\toptwo{19.67}} & \placeholder{\topthree{4.81}} & \textbf{\toptwo{0.92}} & \textbf{\toptwo{0.02}} & 2.18 & \placeholder{\topthree{0.63}} & 21.14 \\ 
    &  ReACGAN~\cite{kang2021rebooting} &  \textbf{\topone{20.64}} &  \textbf{\toptwo{4.69}} &  \textbf{\toptwo{0.92}} &  \textbf{\toptwo{0.02}} &  \placeholder{\topthree{2.24}} &  \textbf{\toptwo{0.64}} &  \placeholder{\topthree{12.37}} \\ 
    \cmidrule[0.25pt]{2-9}
    & StyleGAN2~\cite{karras2020analyzing} & \placeholder{\topthree{17.38}} & \textbf{\topone{4.08}} & 0.88 & \textbf{\topone{0.08}} & 1.87 & \textbf{\topone{0.67}} & \textbf{\topone{9.20}} \\ 
    & StyleGAN3-t~\cite{karras2021alias} & 7.30 & 13.28 & 0.80 & 0.00 & 1.25 & 0.09 & 25.74 \\ 
    \cmidrule[1.0pt]{2-9}
    \parbox[t]{2mm}{\multirow{7}{*}{\rotatebox[origin=c]{90}{\text{ImageNet}}}}
    & SNGAN-256~\cite{Miyato2018SpectralNF} & 29.94 & 6.59 & \textbf{\topone{0.94}} & \placeholder{\topthree{0.06}} & \textbf{\toptwo{2.97}} & 0.60 & \textbf{\toptwo{13.81}} \\ 
    \cmidrule[0.25pt]{2-9}
    &  BigGAN-256~\cite{Brock2019LargeSG} &  \textbf{\toptwo{46.14}} &  \textbf{\toptwo{4.55}} &  \textbf{\topone{0.94}} &  \textbf{\topone{0.15}} &  \textbf{\topone{3.16}} &  \textbf{\topone{0.75}} &  \textbf{\topone{13.10}} \\ 
    & ContraGAN-256~\cite{kang2020contragan} & \placeholder{\topthree{31.97}} & 5.94 & 0.93 & \textbf{\toptwo{0.07}} & \placeholder{\topthree{2.92}} & \placeholder{\topthree{0.62}} & 17.25 \\ 
    & ReACGAN-256~\cite{kang2021rebooting} & \textbf{\topone{89.28}} & \textbf{\topone{4.32}} & \textbf{\topone{0.94}} & 0.05 & 2.66 & \textbf{\toptwo{0.73}} & 16.51 \\ 
    \cmidrule[0.25pt]{2-9}
    & StyleGAN2~\cite{karras2020analyzing} & 25.74 & \placeholder{\topthree{5.59}} & 0.89 & 0.05 & 2.18 & 0.61 & \placeholder{\topthree{14.14}} \\ 
    & StyleGAN3-t~\cite{karras2021alias} & 24.47 & 6.07 & 0.85 & 0.03 & 1.73 & 0.52 & 14.91 \\ 
    \cmidrule[1.0pt]{2-9}
    \parbox[t]{2mm}{\multirow{4}{*}{\rotatebox[origin=c]{90}{\text{AFHQv2}}}}
    & StyleGAN2~\cite{karras2020analyzing} & \textbf{\topone{17.69}} & 1.93 & \textbf{\topone{0.90}} & 0.18 & \textbf{\topone{1.45}} & \placeholder{\topthree{0.83}} & 1.92 \\ 
    &  StyleGAN2 + ADA~\cite{Karras2020TrainingGA} & \placeholder{\topthree{17.61}} & \textbf{\toptwo{1.45}} &  \textbf{\toptwo{0.89}} &  \placeholder{\topthree{0.33}} &  \textbf{\toptwo{1.31}} &  \textbf{\topone{0.86}} &  \textbf{\toptwo{1.42}} \\ 
    & StyleGAN3-t + ADA~\cite{karras2021alias} & 17.31 & \textbf{\topone{1.41}} & 0.83 & \textbf{\topone{0.47}} & 0.97 & \placeholder{\topthree{0.83}} & \textbf{\topone{1.40}} \\ 
    & StyleGAN3-r + ADA~\cite{karras2021alias} & \textbf{\toptwo{17.66}} & \placeholder{\topthree{1.49}} & \placeholder{\topthree{0.84}} & \textbf{\toptwo{0.41}} & \placeholder{\topthree{1.09}} & \textbf{\toptwo{0.85}} & \placeholder{\topthree{1.49}} \\ 
    \cmidrule[1.0pt]{2-9}
    \parbox[t]{2mm}{\multirow{1}{*}{\rotatebox[origin=c]{90}{\text{FQ}}}}
    &  StyleGAN2~\cite{karras2020analyzing} &  3.75 &  0.54 &  0.86 &  0.36 &  1.31 &  0.90 &  - \\
    \cmidrule[1.0pt]{2-9}
    \end{tabular}}
    \label{table:swav}
\end{table*}
\begin{table*}[!htp]
    \caption{Benchmark table evaluated using \textbf{PyTorch-Swin-T}~\cite{liu2021swin}. The resolutions of CIFAR10~\cite{Krizhevsky2009LearningML}, Baby/Papa/Grandpa ImageNet, ImageNet~\cite{Deng2009ImageNetAL}, AFHQv2~\cite{choi2020starganv2, karras2021alias}, and FQ~\cite{karras2019style} datasets are 32, 64, 128, 512, and 1024, respectively. Top-1 and Top-2 performances are indicated in red and blue, respectively.}
    \centering
    \resizebox{0.88\textwidth}{!}{
    \begin{tabular}{clrrrrrrr}
    \cmidrule[1.0pt]{2-9}
    & \textbf{PyTorch-Swin-T}~\cite{liu2021swin} & \text{TS}~$\uparrow$ & \text{FTD}~$\downarrow$ & \text{T-Precision}~$\uparrow$ & \text{T-Recall}~$\uparrow$ & \text{T-Density}~$\uparrow$ & \text{T-Coverage}~$\uparrow$ & \text{IFTD}~$\downarrow$\\
    \cmidrule[1.0pt]{2-9}
    \parbox[t]{2mm}{\multirow{29}{*}{\rotatebox[origin=c]{90}{\text{CIFAR10}}}}
    & DCGAN~\cite{Goodfellow2014GenerativeAN} & 3.16 & 75.08 & 0.33 & 0.10 & 0.36 & 0.17 & 143.96 \\ 
    & LSGAN~\cite{Mao2017LeastSG} & 3.40 & 68.57 & 0.30 & 0.16 & 0.27 & 0.20 & 140.83 \\ 
    & Geometric GAN~\cite{Lim2017GeometricG} & 2.98 & 66.14 & 0.40 & 0.09 & 0.44 & 0.19 & 135.38 \\ 
    \cmidrule[0.25pt]{2-9}
    & ACGAN-Mod~\cite{Odena2017ConditionalIS} & 3.27 & 57.61 & 0.42 & 0.07 & 0.46 & 0.24 & 103.57 \\ 
    & WGAN~\cite{Arjovsky2017WassersteinG} & 2.19 & 158.48 & 0.34 & 0.01 & 0.26 & 0.05 & 201.98 \\ 
    & DRAGAN~\cite{Kodali2018OnCA} & 3.23 & 58.42 & 0.38 & 0.23 & 0.42 & 0.21 & 129.11 \\ 
    & WGAN-GP~\cite{Gulrajani2017ImprovedTO} & 2.87 & 91.45 & 0.32 & 0.13 & 0.30 & 0.15 & 155.69 \\ 
    & PD-GAN~\cite{Miyato2018cGANsWP} & 3.58 & 54.93 & 0.41 & 0.11 & 0.49 & 0.26 & 96.44 \\ 
    & SNGAN~\cite{Miyato2018SpectralNF} & 4.21 & 18.5 & 0.48 & 0.43 & 0.65 & 0.62 & 36.74 \\ 
    & SAGAN~\cite{Zhang2019SelfAttentionGA} & 4.13 & 18.73 & 0.47 & 0.43 & 0.63 & 0.60 & 37.83 \\ 
    & TACGAN~\cite{NIPS2019_8414} & 3.59 & 53.77 & 0.38 & 0.13 & 0.42 & 0.25 & 95.63 \\ 
    & LOGAN~\cite{Wu2019LOGANLO} & 3.77 & 34.66 & 0.41 & 0.41 & 0.49 & 0.39 & 112.82 \\ 
    \cmidrule[0.25pt]{2-9}
    & BigGAN~\cite{Brock2019LargeSG} & 4.72 & 7.69 & \toptwo{\textbf{0.55}} & 0.47 & \placeholder{\topthree{0.92}} & 0.82 & 20.21 \\ 
    & CRGAN~\cite{Zhang2019ConsistencyRF} & 4.95 & 6.91 & 0.53 & \toptwo{\textbf{0.51}} & 0.84 & 0.84 & 19.15 \\ 
    & MHGAN~\cite{kavalerov2021multi} & 4.96 & 8.19 & 0.53 & 0.48 & 0.83 & 0.82 & 21.05 \\ 
    & ICRGAN~\cite{Zhao2020ImprovedCR} & 5.01 & 7.62 & 0.53 & 0.49 & 0.82 & 0.84 & 19.85 \\ 
    & ContraGAN~\cite{kang2020contragan} & 4.66 & 10.39 & 0.54 & 0.46 & 0.85 & 0.76 & 136.36 \\ 
    & BigGAN + DiffAug~\cite{zhao2020differentiable} & 4.48 & \placeholder{\topthree{6.82}} & \topone{\textbf{0.59}} & 0.46 & \topone{\textbf{1.03}} & \placeholder{\topthree{0.86}} & \placeholder{\topthree{18.05}} \\ 
    & BigGAN + LeCam~\cite{lecam2021} & 4.96 & 7.33 & 0.53 & \placeholder{\topthree{0.51}} & 0.84 & 0.83 & 19.79 \\ 
    & ADCGAN~\cite{hou2021cgans} & 4.76 & 7.83 & 0.53 & 0.50 & 0.84 & 0.82 & 20.80 \\ 
    & ReACGAN~\cite{kang2021rebooting} & 4.84 & 8.42 & \toptwo{\textbf{0.55}} & 0.46 & 0.89 & 0.81 & 21.97 \\ 
    \cmidrule[0.25pt]{2-9}
    & StyleGAN2~\cite{karras2020analyzing} & 5.04 & 10.65 & 0.51 & 0.48 & 0.82 & 0.80 & 24.31 \\ 
    &  StyleGAN2 + DiffAug~\cite{karras2020analyzing, zhao2020differentiable} & \toptwo{\textbf{5.16}} &  \toptwo{\textbf{6.07}} &  \toptwo{\textbf{0.55}} &  0.50 &  \toptwo{\textbf{0.96}} & \topone{\textbf{0.88}} & \topone{\textbf{17.55}} \\ 
    & StyleGAN2 + ADA~\cite{Karras2020TrainingGA} & \toptwo{\textbf{5.16}} & \topone{\textbf{6.03}} & 0.54 & \topone{\textbf{0.52}} & 0.91 & \toptwo{\textbf{0.87}} & \toptwo{\textbf{17.78}} \\ 
    & StyleGAN2 + LeCam~\cite{karras2020analyzing, lecam2021} & 4.86 & 12.01 & 0.50 & 0.45 & 0.84 & 0.79 & 26.38 \\ 
    & StyleGAN2 + APA~\cite{deceived2021} & 4.96 & 10.20 & 0.49 & 0.50 & 0.77 & 0.79 & 72.71 \\ 
    & StyleGAN2 + D2D-CE~\cite{kang2021rebooting} & \topone{\textbf{5.23}} & 9.55 & 0.53 & 0.46 & 0.86 & 0.82 & 23.18 \\ 
    & StyleGAN3-r + ADA~\cite{karras2021alias}  & 4.64 & 20.38 & 0.47 & 0.32 & 0.80 & 0.68 & 39.36 \\ 
    \cmidrule[1.0pt]{2-9}

    \parbox[t]{2mm}{\multirow{8}{*}{\rotatebox[origin=c]{90}{\text{Baby-ImageNet}}}}
    & SNGAN~\cite{Miyato2018SpectralNF} & 10.80 & 71.20 & 0.14 & \topone{\textbf{0.57}} & 0.07 & 0.12 & 140.18 \\ 
    & SAGAN~\cite{Zhang2019SelfAttentionGA} & 9.59 & 71.69 & 0.14 & \toptwo{\textbf{0.55}} & 0.07 & 0.12 & 154.44 \\ 
    \cmidrule[0.25pt]{2-9}
    &  BigGAN~\cite{Brock2019LargeSG} &  \topone{\textbf{13.72}} &  \toptwo{\textbf{36.66}} &  \topone{\textbf{0.32}} &  0.39 &  \topone{\textbf{0.29}} &  \topone{\textbf{0.35}} &  \topone{\textbf{90.85}} \\ 
    & ContraGAN~\cite{kang2020contragan} & 12.17 & \placeholder{\topthree{42.04}} & \placeholder{\topthree{0.27}} & 0.44 & \placeholder{\topthree{0.20}} & 0.24 & 229.46 \\ 
    & ReACGAN~\cite{kang2021rebooting} & \toptwo{\textbf{13.71}} & \topone{\textbf{34.98}} & \toptwo{\textbf{0.31}} & 0.40 & \toptwo{\textbf{0.24}} & \placeholder{\topthree{0.29}} & \toptwo{\textbf{100.35}} \\ 
    \cmidrule[0.25pt]{2-9}
    & StyleGAN2~\cite{karras2020analyzing} & \placeholder{\topthree{12.76}} & 50.31 & 0.25 & \toptwo{\textbf{0.55}} & \placeholder{\topthree{0.20}} & \toptwo{\textbf{0.32}} & \placeholder{\topthree{111.18}} \\ 
    & StyleGAN3-t~\cite{karras2021alias} & 6.32 & 172.08 & 0.01 & 0.06 & 0.00 & 0.00 & 322.42 \\ 
    \cmidrule[1.0pt]{2-9}

    \parbox[t]{2mm}{\multirow{8}{*}{\rotatebox[origin=c]{90}{\text{Papa-ImageNet}}}}
    & SNGAN~\cite{Miyato2018SpectralNF} & 8.20 & 78.66 & 0.13 & \toptwo{\textbf{0.51}} & 0.07 & 0.11 & 167.58 \\ 
    & SAGAN~\cite{Zhang2019SelfAttentionGA} & 7.21 & 85.39 & 0.11 & \placeholder{\topthree{0.50}} & 0.06 & 0.10 & 188.46 \\ 
    \cmidrule[0.25pt]{2-9}
    & BigGAN~\cite{Brock2019LargeSG} & \placeholder{\topthree{9.54}} & \placeholder{\topthree{52.72}} & \placeholder{\topthree{0.23}} & 0.38 & \topone{\textbf{0.19}} & \topone{\textbf{0.25}} & \topone{\textbf{128.80}} \\ 
    & ContraGAN~\cite{kang2020contragan} & 9.02 & \toptwo{\textbf{48.02}} & \topone{\textbf{0.25}} & 0.43 & \placeholder{\topthree{0.18}} & \placeholder{\topthree{0.22}} & 247.04 \\ 
    &  ReACGAN~\cite{kang2021rebooting} &  \toptwo{\textbf{9.98}} &  \topone{\textbf{43.34}} &  \topone{\textbf{0.25}} &  0.42 &  \topone{\textbf{0.19}} & \topone{\textbf{0.25}} &  \toptwo{\textbf{133.91}} \\ 
    \cmidrule[0.25pt]{2-9}
    & StyleGAN2~\cite{karras2020analyzing} & \topone{\textbf{10.49}} & 58.18 & 0.17 & \topone{\textbf{0.55}} & 0.12 & \placeholder{\topthree{0.22}} & \placeholder{\topthree{135.01}} \\ 
    & StyleGAN3-t~\cite{karras2021alias} & 5.78 & 150.59 & 0.05 & 0.02 & 0.02 & 0.01 & 323.48 \\ 
    \cmidrule[1.0pt]{2-9}

    \parbox[t]{2mm}{\multirow{8}{*}{\rotatebox[origin=c]{90}{\text{Grandpa-ImageNet}}}}
    & SNGAN~\cite{Miyato2018SpectralNF} & 7.08 & 80.65 & 0.13 & \toptwo{\textbf{0.43}} & 0.08 & 0.13 & 179.30 \\ 
    & SAGAN~\cite{Zhang2019SelfAttentionGA} & 6.65 & 89.89 & 0.11 & \placeholder{\topthree{0.40}} & 0.07 & 0.11 & 206.66 \\ 
    \cmidrule[0.25pt]{2-9}
    & BigGAN~\cite{Brock2019LargeSG} & 8.09 & \placeholder{\topthree{53.61}} & \placeholder{\topthree{0.22}} & 0.34 & \placeholder{\topthree{0.20}} & \toptwo{\textbf{0.27}} & \topone{\textbf{134.14}} \\ 
    & ContraGAN~\cite{kang2020contragan} & \placeholder{\topthree{8.43}} & \toptwo{\textbf{45.50}} & \topone{\textbf{0.27}} & 0.34 & \topone{\textbf{0.24}} & \toptwo{\textbf{0.27}} & 255.97 \\ 
    &  ReACGAN~\cite{kang2021rebooting} &  \topone{\textbf{9.34}} &  \topone{\textbf{43.35}} &  \toptwo{\textbf{0.26}} &  0.32 &  \toptwo{\textbf{0.23}} &  \topone{\textbf{0.29}} &  \placeholder{\topthree{147.42}} \\ 
    \cmidrule[0.25pt]{2-9}
    & StyleGAN2~\cite{karras2020analyzing} & \toptwo{\textbf{9.29}} & 56.41 & 0.16 & \topone{\textbf{0.50}} & 0.12 & 0.23 & \toptwo{\textbf{142.89}} \\ 
    & StyleGAN3-t~\cite{karras2021alias} & 4.30 & 161.33 & 0.10 & 0.00 & 0.05 & 0.02 & 345.20 \\ 
    \cmidrule[1.0pt]{2-9}

    \parbox[t]{2mm}{\multirow{7}{*}{\rotatebox[origin=c]{90}{\text{ImageNet}}}}
    & SNGAN-256~\cite{Miyato2018SpectralNF} & \placeholder{\topthree{17.09}} & 57.72 & 0.09 & \placeholder{\topthree{0.70}} & 0.08 & \placeholder{\topthree{0.13}} & \topone{\textbf{168.27}} \\ 
    \cmidrule[0.25pt]{2-9}

    & BigGAN-256~\cite{Brock2019LargeSG} & \toptwo{\textbf{20.39}} & \toptwo{\textbf{36.78}} & \toptwo{\textbf{0.18}} & 0.69 & \placeholder{\topthree{0.18}} & \toptwo{\textbf{0.26}} & \placeholder{\topthree{181.24}} \\ 
    & ContraGAN-256~\cite{kang2020contragan} & 15.35 & \placeholder{\topthree{48.66}} & \placeholder{\topthree{0.14}} & 0.69 & \toptwo{\textbf{0.22}} & 0.09 & 216.74 \\ 
    &  ReACGAN-256~\cite{kang2021rebooting} & \topone{\textbf{26.33}} &  \topone{\textbf{25.77}} &  \topone{\textbf{0.30}} &  0.47 &  \topone{\textbf{0.26}} &  \topone{\textbf{0.28}} &  \toptwo{\textbf{169.14}} \\ 
    \cmidrule[0.25pt]{2-9}

    & StyleGAN2~\cite{karras2020analyzing} & 13.67 & 64.96 & 0.05 & \topone{\textbf{0.74}} & 0.02 & 0.07 & 204.21 \\ 
    & StyleGAN3-t~\cite{karras2021alias} & 13.57 & 72.34 & 0.03 & \toptwo{\textbf{0.73}} & 0.01 & 0.04 & 218.04 \\ 
    \cmidrule[1.0pt]{2-9}
    \parbox[t]{2mm}{\multirow{4}{*}{\rotatebox[origin=c]{90}{\text{AFHQv2}}}}
    & StyleGAN2~\cite{karras2020analyzing} & \placeholder{\topthree{7.74}} & \placeholder{\topthree{14.86}} & \toptwo{\textbf{0.49}} & 0.35 & \toptwo{\textbf{0.57}} & \toptwo{\textbf{0.56}} & \placeholder{\topthree{14.90}} \\ 
    &  StyleGAN2 + ADA~\cite{Karras2020TrainingGA} &  \topone{\textbf{8.23}} &  \topone{\textbf{12.78}} &  \topone{\textbf{0.53}} &  \placeholder{\topthree{0.40}} &  \topone{\textbf{0.71}} &  \topone{\textbf{0.63}} &  \topone{\textbf{12.99}} \\ 
    & StyleGAN3-t + ADA~\cite{karras2021alias} & 7.47 & \toptwo{\textbf{14.57}} & \placeholder{\topthree{0.40}} & \toptwo{\textbf{0.54}} & \placeholder{\topthree{0.42}} & \placeholder{\topthree{0.55}} & \toptwo{\textbf{14.59}} \\ 
    & StyleGAN3-r + ADA~\cite{karras2021alias} & \toptwo{\textbf{7.75}} & 16.61 & 0.35 & \topone{\textbf{0.56}} & 0.35 & 0.53 & 16.56 \\
    \cmidrule[1.0pt]{2-9}
    \parbox[t]{2mm}{\multirow{1}{*}{\rotatebox[origin=c]{90}{\text{FQ}}}}
    &  StyleGAN2~\cite{karras2020analyzing} &  4.02 &  16.15 &  0.57 &  0.44 &  0.85 &  0.77 &  - \\
    \cmidrule[1.0pt]{2-9}
    \end{tabular}}
    \label{table:swin}
\end{table*}
\clearpage

\begin{table*}[!htp]
    \centering
    \caption{Benckmark table for GANs~\cite{Brock2019LargeSG, esser2021taming, kang2021rebooting, karras2021alias, sauer2022stylegan}, autoregressive models~\cite{esser2021taming, chang2022maskgit, lee2022autoregressive}, and diffusion probabilistic models~\cite{ho2020denoising, song2021scorebased, dhariwal2021diffusion, vahdat2021score, dockhorn2022score}. We evaluate those generative models using \textbf{PyTorch-SwAV}~\cite{Caron2020UnsupervisedLO} and the architecture-friendly resizer: PIL.BILINEAR for the postprocessing step. In the case of ImageNet-128 and ImageNet-256 experiments, we preprocess images using the best performing resizer among PIL.BILINEAR, PIL.BICUBIC, and PIL.LANCZOS based on Fr\'echet Distance. We mark $\dag$ if images are preprocessed using PIL.BILINEAR for evaluation, $\dag\dag$ for the case of PIL.BICUBIC, and $\dag\dag\dag$ for the case of PIL.LANCZOS. Top-1 and Top-2 performances are indicated in red and blue, respectively.}
    \resizebox{0.95\textwidth}{!}{
    \begin{tabular}{clrrrrrrr}
    \cmidrule[0.75pt]{2-9}
    & Model & \text{SS}~$\uparrow$ & \text{FSD}~$\downarrow$ & \text{S-Precision}~$\uparrow$ & \text{S-Recall}~$\uparrow$ & \text{S-Density}~$\uparrow$ & \text{S-Coverage}~$\uparrow$ & Avg. Rank\\
    \cmidrule[0.75pt]{2-9}
    \parbox[t]{2mm}{\multirow{11}{*}{\rotatebox[origin=c]{90}{\text{CIFAR10}}}}
    & ReACGAN + DiffAug (Ours)~\cite{kang2021rebooting} & 14.28 & 0.37 & 0.85 & 0.55 & 1.09 & 0.90 & 8.33 \\ 
    & StyleGAN2-ADA~\cite{kang2020contragan} & 14.57 & 0.36 & 0.84 & 0.59 & 1.08 & 0.92 & 7.50 \\ 
    & StyleGAN2-ADA (Ours)~\cite{kang2020contragan} & \toptwo{\textbf{15.30}} & 0.33 & 0.83 & 0.63 & 1.05 & 0.92 & 6.50 \\ 
    & StyleGAN2 + DiffAug + D2D-CE (Ours)~\cite{kang2021rebooting} & 14.79 & 0.34 & 0.87 & 0.58 & 1.19 & \placeholder{\topthree{0.94}} & 5.50 \\ 
    & DDPM~\cite{ho2020denoising} & 13.47 & 0.49 & \topone{\textbf{0.89}} & 0.62 & \topone{\textbf{1.35}} & 0.93 & 6.17 \\ 
    & DDPM++~\cite{song2021scorebased} & 14.03 & 0.33 & \toptwo{\textbf{0.88}} & 0.67 & \toptwo{\textbf{1.27}} & \topone{\textbf{0.95}} & \toptwo{\textbf{4.00}} \\ 
    & NCSN++~\cite{song2021scorebased} & \placeholder{\topthree{15.03}} & 0.33 & 0.83 & \toptwo{\textbf{0.73}} & 0.97 & 0.92 & 6.17 \\ 
    & LSGM~\cite{vahdat2021score} & 14.78 & 0.45 & \toptwo{\textbf{0.88}} & \placeholder{\topthree{0.69}} & 1.20 & \placeholder{\topthree{0.94}} & 4.50 \\ 
    & LSGM-ODE~\cite{vahdat2021score} & 14.67 & \toptwo{\textbf{0.27}} & 0.85 & \topone{\textbf{0.74}} & 1.06 & \placeholder{\topthree{0.94}} & \placeholder{\topthree{4.33}} \\ 
    &  CLD-SGM~\cite{dockhorn2022score} &  14.27 &  \placeholder{\topthree{0.30}} &  \toptwo{\textbf{0.88}} &  \placeholder{\topthree{0.69}} &  \placeholder{\topthree{1.25}} &  \topone{\textbf{0.95}} &  \topone{\textbf{3.50}} \\ 
    & StyleGAN-XL~\cite{sauer2022stylegan} & \topone{\textbf{16.00}} & \topone{\textbf{0.24}} & 0.84 & 0.45 & 1.06 & 0.91 & 6.50 \\ 
    \cmidrule[0.75pt]{2-9}
    \parbox[t]{2mm}{\multirow{6}{*}{\rotatebox[origin=c]{90}{\text{ImageNet-128}}}}
    & BigGAN~\cite{Brock2019LargeSG}$^{\dag\dag}$ & 128.55 & 4.56 & \topone{\textbf{0.98}} & 0.07 & \topone{\textbf{3.76}} & 0.83 & 3.83  \\ 
    & BigGAN (Ours)\cite{Brock2019LargeSG}$^{\dag}$ & 97.27 & 4.49 & \toptwo{\textbf{0.97}} & 0.12 & \toptwo{\textbf{3.36}} & 0.81 & 4.17  \\ 
    & BigGAN-Deep~\cite{Brock2019LargeSG}$^{\dag\dag}$ & \placeholder{\topthree{230.57}} & \placeholder{\topthree{2.49}} & \toptwo{\textbf{0.97}} & \placeholder{\topthree{0.17}} & \placeholder{\topthree{2.89}} & \placeholder{\topthree{0.91}} & \topone{\textbf{2.83}} \\ 
    & ReACGAN (Ours)~\cite{kang2021rebooting}$^{\dag}$ & 131.14 & 3.18 & 0.94 & 0.10 & 2.40 & 0.83 & 4.17 \\ 
    &  StyleGAN-XL~\cite{sauer2022stylegan}$^{\dag\dag\dag}$ &  \topone{\textbf{311.54}} &  \topone{\textbf{1.04}} &  0.91 &  \toptwo{\textbf{0.39}} &  1.68 &  \toptwo{\textbf{0.94}} &   \topone{\textbf{2.83}}  \\ 
    & ADM-G~\cite{dhariwal2021diffusion}$^{\dag}$ & \toptwo{\textbf{233.15}} & \toptwo{\textbf{1.31}} & 0.92 & \topone{\textbf{0.51}} & 1.66 & \topone{\textbf{0.97}} & \topone{\textbf{2.83}} \\
    \cmidrule[0.75pt]{2-9}
    \parbox[t]{2mm}{\multirow{8}{*}{\rotatebox[origin=c]{90}{\text{ImageNet-256}}}}
    & BigGAN~\cite{Brock2019LargeSG}$^{\dag}$ & 202.85 & 4.02 & \topone{\textbf{0.98}} & 0.10 & \toptwo{\textbf{3.35}} & 0.85 & 5.50  \\ 
    & BigGAN-Deep~\cite{Brock2019LargeSG}$^{\dag}$ & 271.73 & 3.57 & \topone{\textbf{0.98}} & 0.07 & \topone{\textbf{3.43}} & 0.87 & 4.83  \\ 
    & VQGAN~\cite{esser2021taming}$^{\dag}$ & \placeholder{\topthree{323.16}} & 3.03 & 0.91 & 0.25 & 1.62 & 0.84 & 6.00 \\ 
    & StyleGAN-XL~\cite{sauer2022stylegan}$^{\dag\dag}$ & \toptwo{\textbf{383.64}} & \topone{\textbf{1.08}} & 0.90 & \placeholder{\topthree{0.40}} & 1.60 & 0.94 & 4.17 \\ 
    & ADM-G~\cite{dhariwal2021diffusion}$^{\dag}$ & 320.47 & \toptwo{\textbf{1.54}} & 0.92 & \topone{\textbf{0.48}} & 1.59 & \topone{\textbf{0.95}} & \placeholder{\topthree{3.83}} \\ 
    &  ADM-G-U~\cite{dhariwal2021diffusion}$^{\dag}$ &  322.29 &  \placeholder{\topthree{1.78}} &  0.93 &  \toptwo{\textbf{0.42}} &  1.81 &  \topone{\textbf{0.95}} &   \topone{\textbf{3.33}} \\ 
    & MaskGIT~\cite{chang2022maskgit}$^{\dag}$ & 265.34 & 2.48 & \placeholder{\topthree{0.96}} & 0.27 & \placeholder{\topthree{2.42}} & \topone{\textbf{0.95}} & 4.00 \\ 
    & RQ-Transformer~\cite{lee2022autoregressive}$^{\dag}$ & \topone{\textbf{390.48}} & 2.14 & 0.95 & 0.36 & 1.87 & 0.93 & \toptwo{\textbf{3.67}} \\ 
    \cmidrule[0.75pt]{2-9}
    \end{tabular}}
    \label{table:other_generative_models_swav}
\end{table*}


\begin{table*}[!htp]
    \centering
    \vspace{-2mm}
    \caption{Benckmark table for GANs~\cite{Brock2019LargeSG, esser2021taming, kang2021rebooting, karras2021alias, sauer2022stylegan}, autoregressive models~\cite{esser2021taming, chang2022maskgit, lee2022autoregressive}, and diffusion probabilistic models~\cite{ho2020denoising, song2021scorebased, dhariwal2021diffusion, vahdat2021score, dockhorn2022score}. We evaluate those generative models using \textbf{PyTorch-Swin-T}~\cite{liu2021swin} and the architecture-friendly resizer: PIL.BICUBIC for the postprocessing step. In the case of ImageNet-128 and ImageNet-256 experiments, we preprocess images using the best performing resizer among PIL.BILINEAR, PIL.BICUBIC, and PIL.LANCZOS based on Fr\'echet Distance. We mark $\dag$ if images are preprocessed using PIL.BILINEAR for evaluation, $\dag\dag$ for the case of PIL.BICUBIC, and $\dag\dag\dag$ for the case of PIL.LANCZOS. Top-1 and Top-2 performances are indicated in red and blue, respectively.}
    \resizebox{0.95\textwidth}{!}{
    \begin{tabular}{clrrrrrrr}
    \cmidrule[0.75pt]{2-9}
    & Model & \text{TS}~$\uparrow$ & \text{FTD}~$\downarrow$ & \text{T-Precision}~$\uparrow$ & \text{T-Recall}~$\uparrow$ & \text{T-Density}~$\uparrow$ & \text{T-Coverage}~$\uparrow$ & Avg. Rank\\
    \cmidrule[0.75pt]{2-9}
    \parbox[t]{2mm}{\multirow{11}{*}{\rotatebox[origin=c]{90}{\text{CIFAR10}}}}
    & ReACGAN + DiffAug (Ours)~\cite{kang2021rebooting} & 4.91 & 6.14 & 0.55 & 0.48 & 0.88 & 0.83 & 9.33  \\ 
    & StyleGAN2-ADA~\cite{kang2020contragan} & 5.11 & 6.29 & 0.55 & 0.51 & 0.92 & 0.87 & 7.67 \\ 
    & StyleGAN2-ADA (Ours)~\cite{kang2020contragan} & \placeholder{\topthree{5.12}} & 5.43 & 0.57 & 0.50 & 1.00 & 0.89 & 5.83 \\ 
    & StyleGAN2 + DiffAug + D2D-CE (Ours)~\cite{kang2021rebooting} & \topone{\textbf{5.14}} & 6.64 & 0.57 & 0.49 & 0.94 & 0.88 & 7.00 \\ 
    & DDPM~\cite{ho2020denoising} & \topone{\textbf{5.14}} & 7.30 & 0.55 & \toptwo{\textbf{0.55}} & 0.87 & 0.85 & 7.33 \\ 
    & DDPM++~\cite{song2021scorebased} & 5.01 & 4.56 & 0.59 & \toptwo{\textbf{0.55}} & 1.02 & 0.90 & 4.50 \\ 
    & NCSN++~\cite{song2021scorebased} & 4.84 & \placeholder{\topthree{3.54}} & \toptwo{\textbf{0.62}} & 0.54 & \placeholder{\topthree{1.11}} & \toptwo{\textbf{0.92}} & \placeholder{\topthree{4.17}} \\ 
    & LSGM~\cite{vahdat2021score} & 4.39 & 5.93 & \topone{\textbf{0.68}} & 0.48 & \topone{\textbf{1.34}} & \topone{\textbf{0.93}} & 5.00 \\ 
    & LSGM-ODE~\cite{vahdat2021score} & 4.97 & \toptwo{\textbf{3.40}} & 0.59 & \topone{\textbf{0.57}} & 0.97 & \toptwo{\textbf{0.92}} & \toptwo{\textbf{4.00}} \\ 
    & CLD-SGM~\cite{dockhorn2022score} & 5.00 & 3.75 & 0.59 & \toptwo{\textbf{0.55}} & 0.99 & 0.91 & 4.67 \\ 
    &  StyleGAN-XL~\cite{sauer2022stylegan} &  5.05 &  \topone{\textbf{2.99}} &  \toptwo{\textbf{0.62}} &  0.43 &  \toptwo{\textbf{1.13}} &  \toptwo{\textbf{0.92}} & \topone{\textbf{3.83}} \\
    \cmidrule[0.75pt]{2-9}
    \parbox[t]{2mm}{\multirow{6}{*}{\rotatebox[origin=c]{90}{\text{ImageNet-128}}}}
    & BigGAN~\cite{Brock2019LargeSG}$^{\dag}$ & 37.04 & 17.91 & 0.34 & 0.48 & 0.42 & 0.58 & 4.33  \\ 
    & BigGAN (Ours)\cite{Brock2019LargeSG}$^{\dag}$ & 26.90 & 16.72 & 0.32 & \topone{\textbf{0.56}} & 0.38 & 0.57 & 4.67 \\ 
    & BigGAN-Deep~\cite{Brock2019LargeSG}$^{\dag\dag}$ & \toptwo{\textbf{53.71}} & \placeholder{\topthree{9.82}} & \placeholder{\topthree{0.49}} & 0.46 & \topone{\textbf{0.76}} & \placeholder{\topthree{0.73}} & \placeholder{\topthree{3.17}} \\ 
    & ReACGAN (Ours)~\cite{kang2021rebooting}$^{\dag}$ & 29.55 & 15.05 & 0.34 & 0.47 & 0.37 & 0.46 & 5.00 \\ 
    & StyleGAN-XL~\cite{sauer2022stylegan}$^{\dag\dag\dag}$ & \topone{\textbf{60.72}} & \topone{\textbf{4.47}} & \toptwo{\textbf{0.50}} & \placeholder{\topthree{0.50}} & \placeholder{\topthree{0.72}} & \toptwo{\textbf{0.84}} & \toptwo{\textbf{2.00}} \\ 
    &  ADM-G~\cite{dhariwal2021diffusion}$^{\dag}$ &  \placeholder{\topthree{52.88}} &  \toptwo{\textbf{6.03}} &  \topone{\textbf{0.51}} &  \toptwo{\textbf{0.55}} &  \topone{\textbf{0.82}} &  \topone{\textbf{0.89}} &   \topone{\textbf{1.67}} \\ 
    \cmidrule[0.75pt]{2-9}
    \parbox[t]{2mm}{\multirow{8}{*}{\rotatebox[origin=c]{90}{\text{ImageNet-256}}}}
    & BigGAN~\cite{Brock2019LargeSG}$^{\dag}$ & 111.25 & 19.15 & 0.39 & 0.49 & 0.50 & 0.54 & 7.17  \\ 
    & BigGAN-Deep~\cite{Brock2019LargeSG}$^{\dag}$ & 125.19 & 13.93 & 0.43 & 0.43 & 0.55 & 0.58 & 6.17 \\ 
    & VQGAN~\cite{esser2021taming}$^{\dag\dag}$ & \toptwo{\textbf{198.75}} & 11.66 & 0.40 & 0.53 & 0.40 & 0.57 & 5.17 \\ 
    & StyleGAN-XL~\cite{sauer2022stylegan}$^{\dag\dag}$ & 159.50 & \topone{\textbf{5.43}} & 0.48 & 0.52 & \toptwo{\textbf{0.62}} & \toptwo{\textbf{0.77}} & \toptwo{\textbf{3.00}} \\ 
    & ADM-G~\cite{dhariwal2021diffusion}$^{\dag\dag}$ & 131.31 & \placeholder{\topthree{8.37}} & \placeholder{\topthree{0.49}} & \toptwo{\textbf{0.59}} & \placeholder{\topthree{0.61}} & \toptwo{\textbf{0.77}} & \toptwo{\textbf{3.00}} \\ 
    &  ADM-G-U~\cite{dhariwal2021diffusion}$^{\dag\dag}$ &  \placeholder{\topthree{174.74}} &  \toptwo{\textbf{7.81}} &  \topone{\textbf{0.54}} &  \placeholder{\topthree{0.55}} &  \topone{\textbf{0.71}} &  \topone{\textbf{0.81}} &   \topone{\textbf{1.83}} \\ 
    & MaskGIT~\cite{chang2022maskgit}$^{\dag}$ & 108.39 & 13.33 & 0.35 & \topone{\textbf{0.63}} & 0.36 & 0.62 & 6.00 \\ 
    & RQ-Transformer~\cite{lee2022autoregressive}$^{\dag}$ & \topone{\textbf{215.46}} & 9.20 & \toptwo{\textbf{0.52}} & 0.52 & 0.59 & 0.71 & 3.33 \\ 
    \cmidrule[0.75pt]{2-9}
    \end{tabular}}
    \label{table:other_generative_models_swin}
\end{table*}
\clearpage

\begin{figure*}[!ht]
    \includegraphics[width=0.95\linewidth]{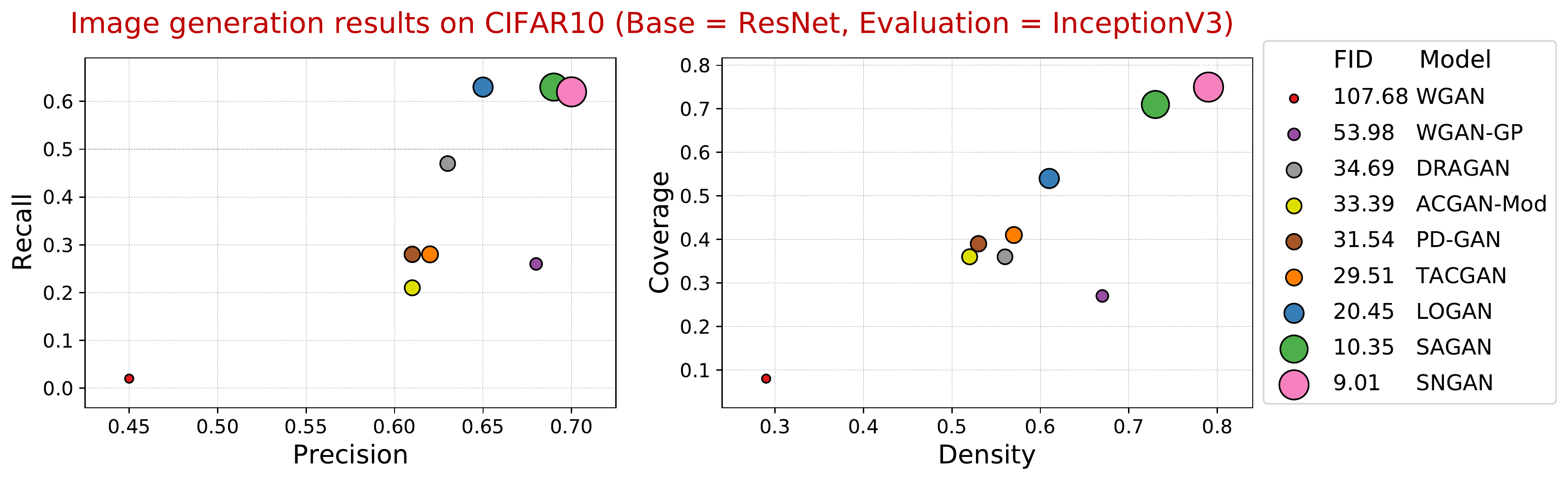}\\    
    \includegraphics[width=0.95\linewidth]{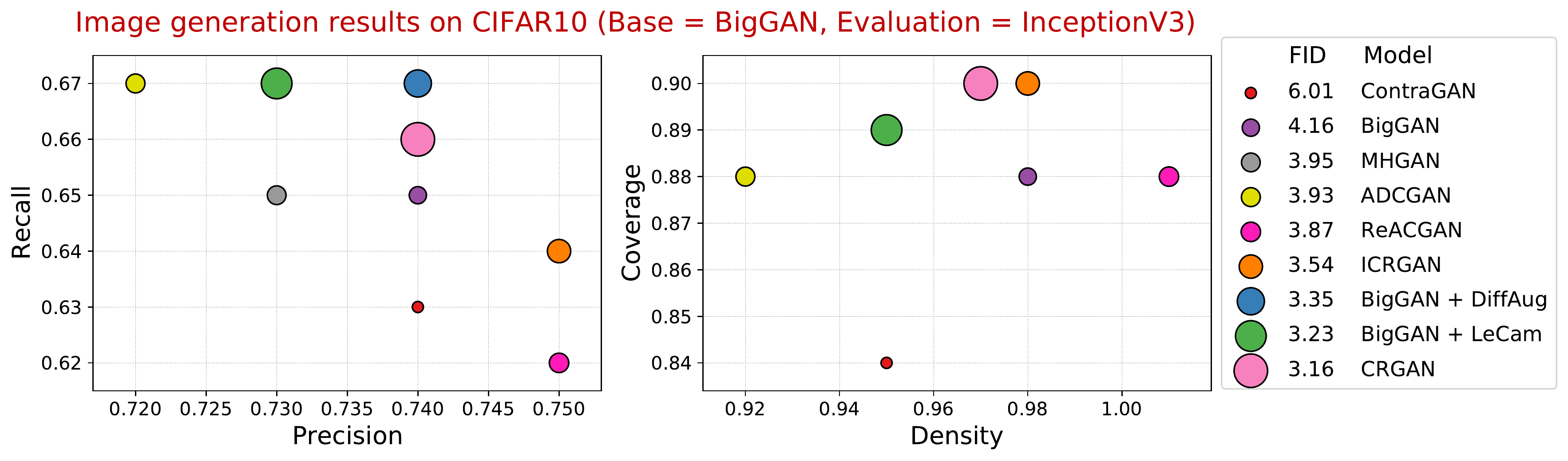}\\
    \includegraphics[width=0.95\linewidth]{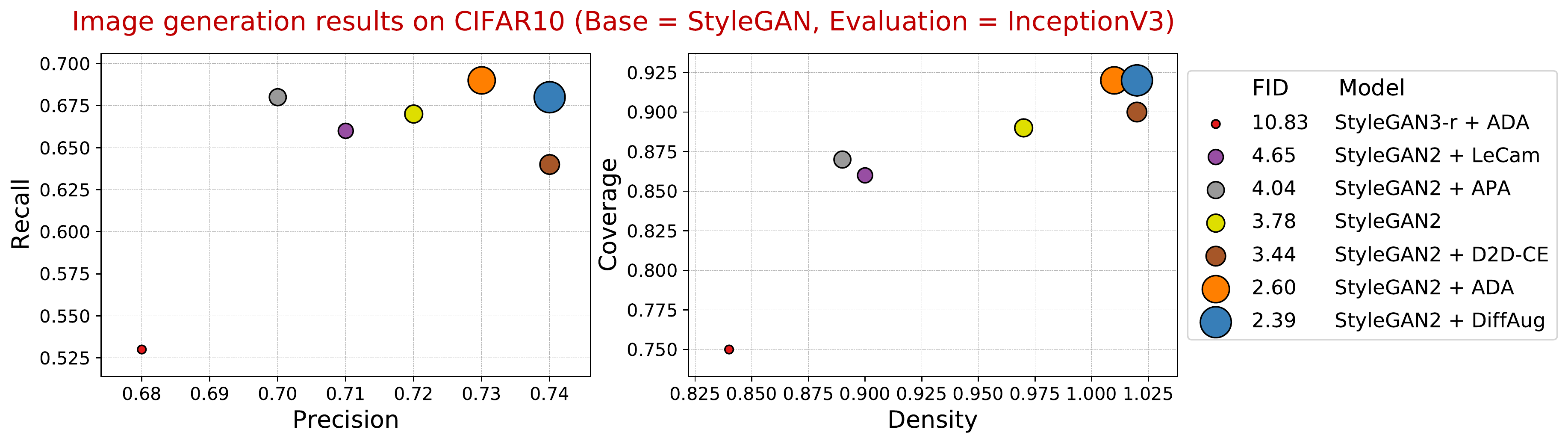}
    \caption{Scatter plots for analyzing GANs from the perspective of fidelity and diversity. We adopt TensorFlow-InceptionV3~\cite{Szegedy2016RethinkingTI} as the evaluation backbone.}
    \label{fig:cifar10_inception_scatter}
    \vspace{-3mm}
\end{figure*} 
\begin{figure*}[!ht]
    \includegraphics[width=0.95\linewidth]{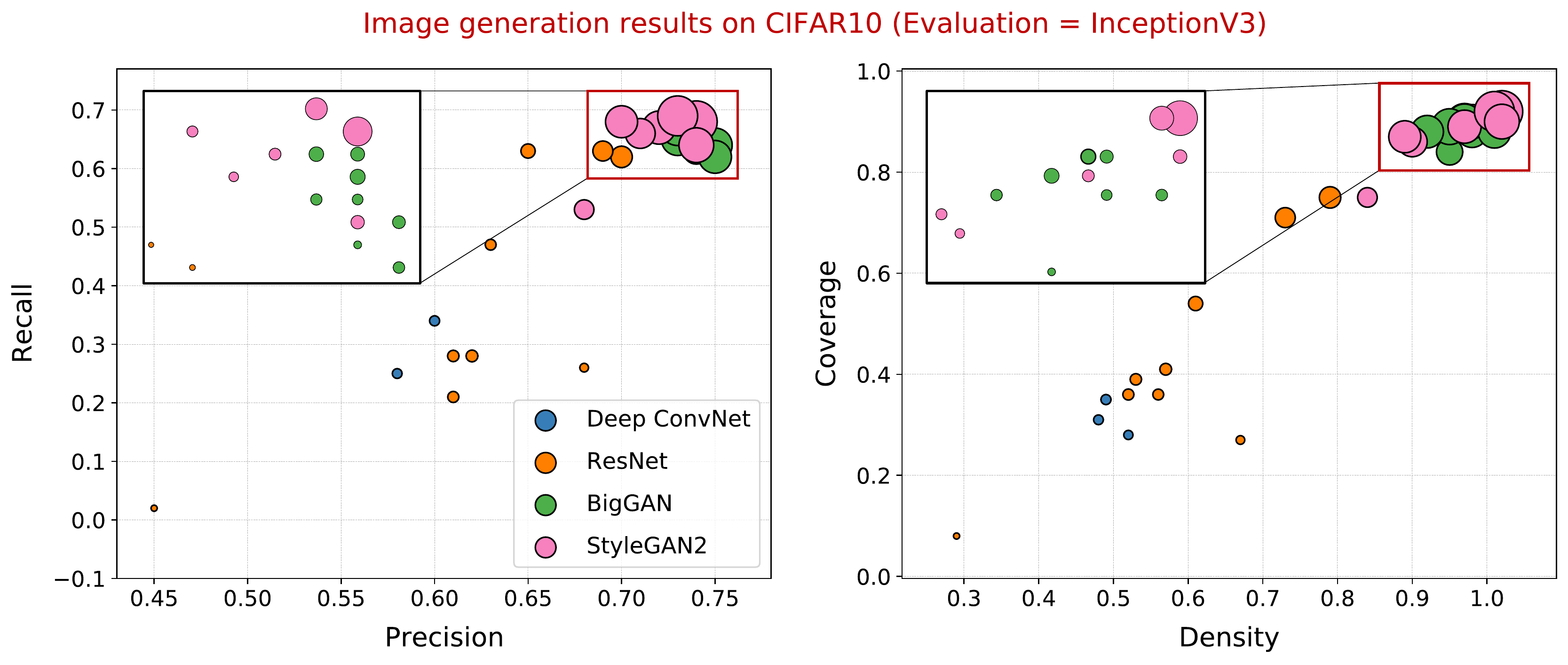}
    \vspace{-3mm}
    \caption{Scatter plots for analyzing the characteristics of distinct GAN architectures. We train GANs on CIFAR10~\cite{Krizhevsky2009LearningML} and evaluate each GAN using TensorFlow-InceptionV3~\cite{Szegedy2016RethinkingTI}.}
    \label{fig:cifar10_inception_whole_scatter}
\end{figure*}
\clearpage

\clearpage

\clearpage

\begin{figure*}[!ht]
    \includegraphics[width=0.99\linewidth]{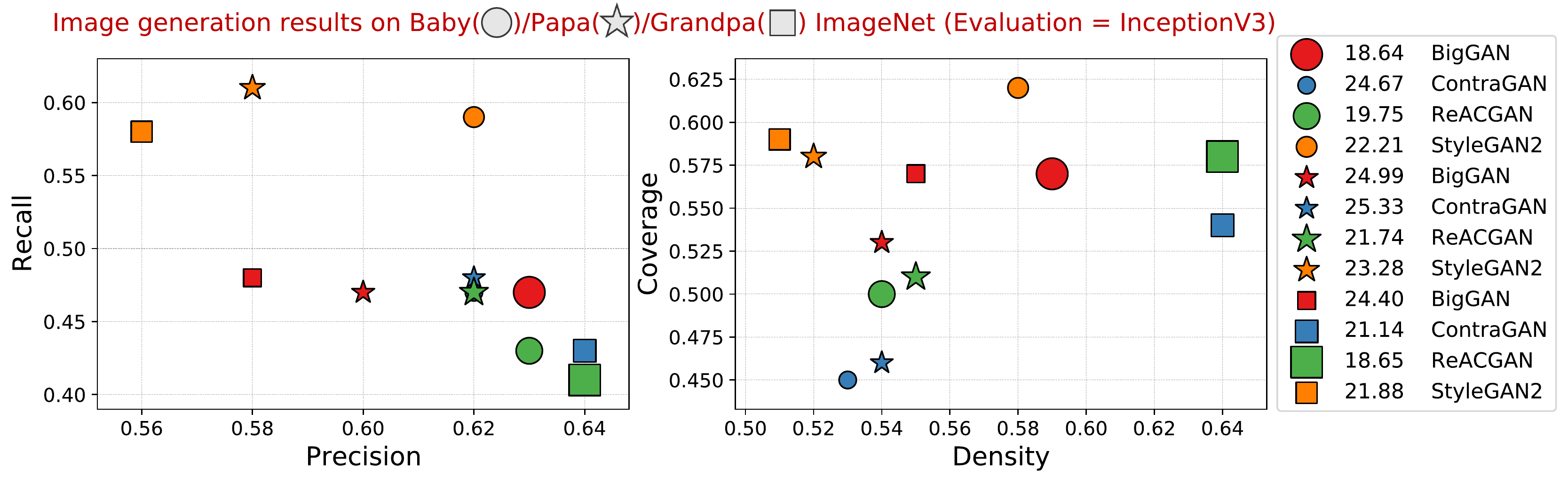}
    \vspace{-3mm}
    \caption{Scatter plots for analyzing the characteristics of distinct GAN architectures. We train GANs on Baby, Papa, and Grandpa ImageNet datasets and evaluate those GANs using TensorFlow-InceptionV3~\cite{Szegedy2016RethinkingTI}.}
    \label{fig:imagenet_series_inception_scatter}
\end{figure*}
\begin{figure*}[hbt!]
    \includegraphics[width=0.95\linewidth]{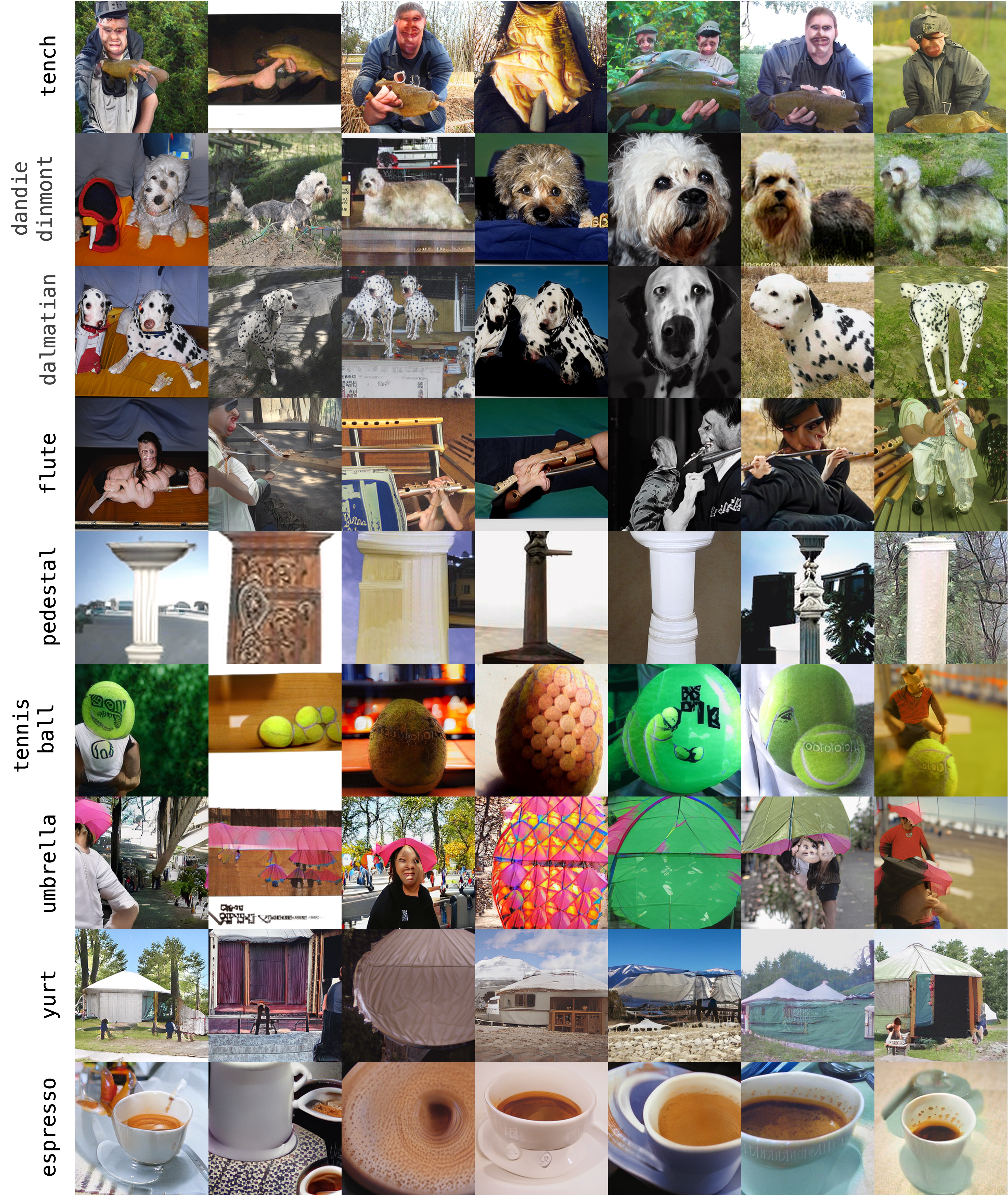}
    \caption{Random selection of generated images on ImageNet-256~\cite{Deng2009ImageNetAL} using StyleGAN-XL~\cite{sauer2022stylegan}~(FID $= 2.32$, FSD $= 1.08$, and FTD $= 5.43$).}
    \label{fig:imagenet_styleganxl}
\end{figure*}
\begin{figure*}[hbt!]
    \includegraphics[width=0.95\linewidth]{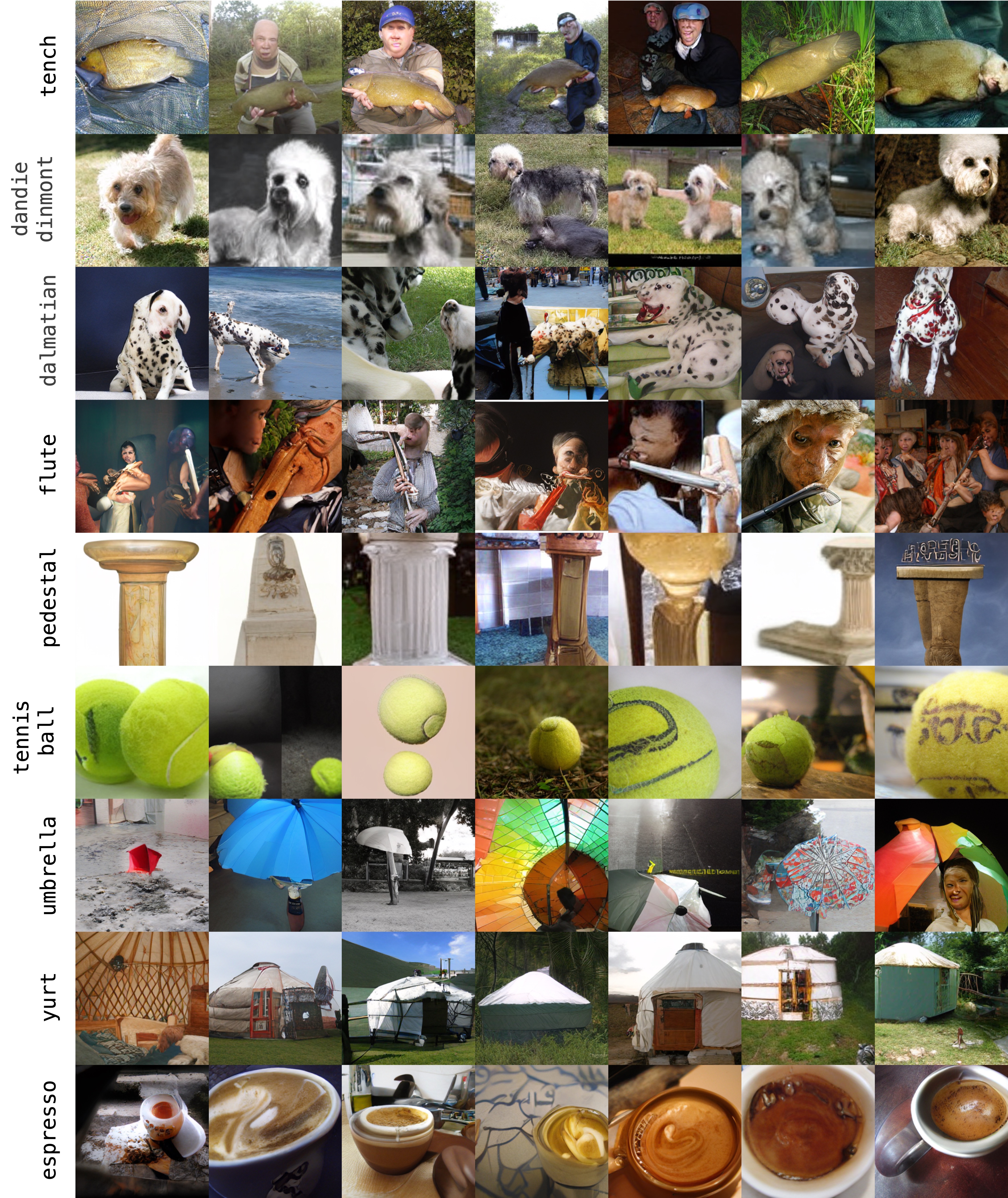}
    \caption{Random selection of generated images on ImageNet-256~\cite{Deng2009ImageNetAL} using RQ-Transformer~\cite{lee2022autoregressive}~(FID $= 3.83$, FSD $= 2.14$, and FTD $= 9.2$).}
    \label{fig:imagenet_rq}
\end{figure*}
\begin{figure*}[hbt!]
    \includegraphics[width=0.95\linewidth]{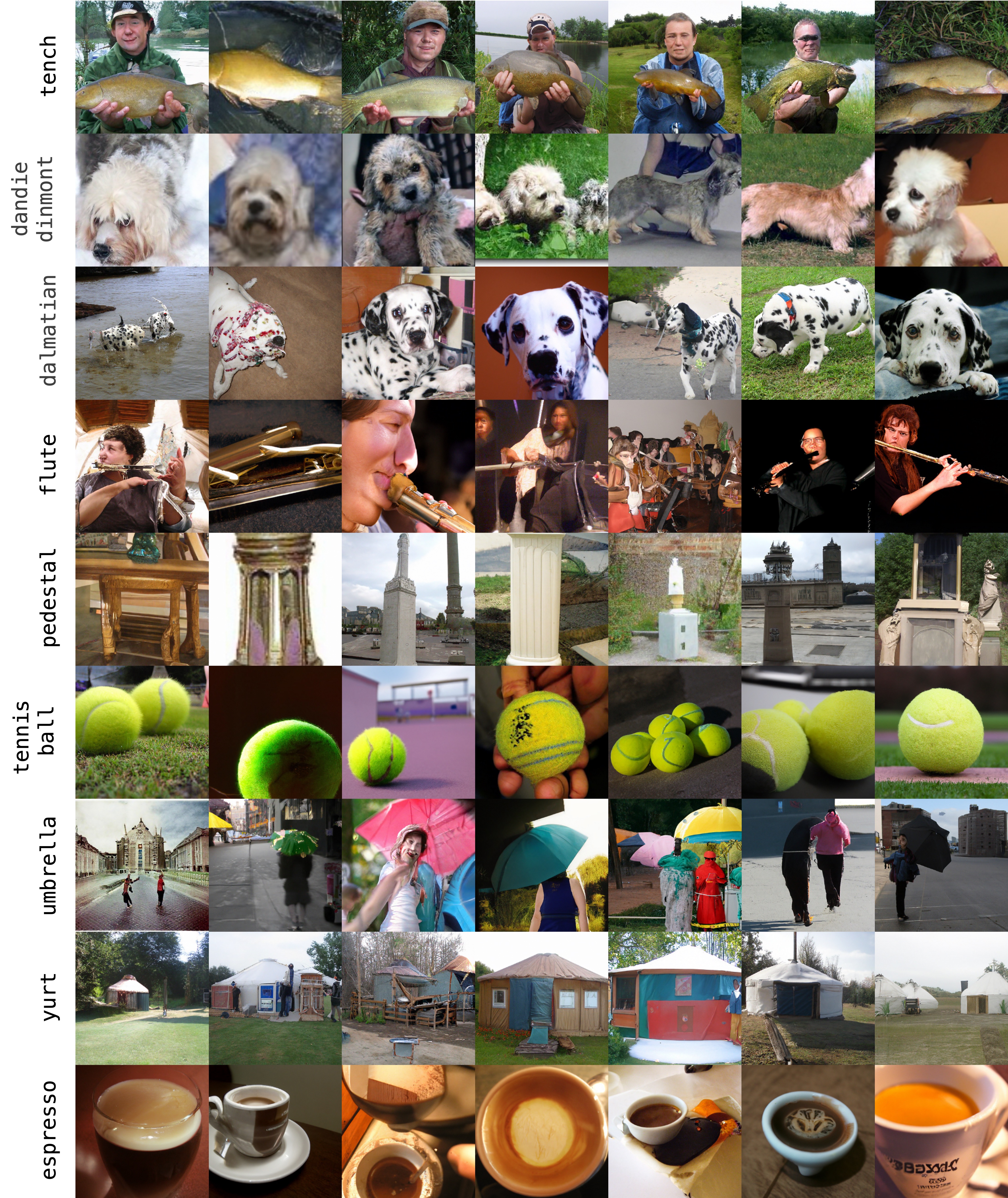}
    \caption{Random selection of generated images on ImageNet-256~\cite{Deng2009ImageNetAL} using ADM-G-U~\cite{dhariwal2021diffusion}~(FID $= 4.01$, FSD $= 1.78$, and FTD $= 7.81$).}
    \label{fig:imagenet_adm}
\end{figure*}

\end{document}